\documentclass[11pt,a4paper]{article}
\ifdefined\pdfsuppresswarningpagegroup\pdfsuppresswarningpagegroup=1\fi

\usepackage[T1]{fontenc}
\usepackage{lmodern}
\usepackage[a4paper,margin=2.5cm]{geometry}
\usepackage{microtype}
\usepackage{amsmath}
\usepackage{amssymb}
\usepackage{graphicx}
\usepackage{xcolor}
\usepackage{array}
\usepackage{caption}
\usepackage{subcaption}
\usepackage{placeins}
\usepackage[ruled,linesnumbered]{algorithm2e}
\usepackage{authblk}
\usepackage[numbers,sort&compress]{natbib}
\usepackage{url}
\usepackage[colorlinks=true,linkcolor=blue!60!black,citecolor=blue!60!black,urlcolor=blue!60!black,breaklinks=true]{hyperref}

\hypersetup{
  pdftitle={Assessing feature scorer results on high-dimensional datasets with t-SNE},
  pdfauthor={Bruno Iochins Grisci, Mario Inostroza-Ponta, Márcio Dorn},
  pdfsubject={Published in Neurocomputing 652 (2025) 130561, https://doi.org/10.1016/j.neucom.2025.130561},
  pdfkeywords={Dimensionality reduction, Data visualization, Feature scoring, Machine learning interpretability}
}

\makeatletter
\renewcommand{\@seccntformat}[1]{\@ifundefined{#1@cntformat}{\csname the#1\endcsname.\quad}{\csname #1@cntformat\endcsname}}
\makeatother

\makeatletter
\newcommand{\bottomnotice}[1]{%
  \begingroup
  \renewcommand{\@makefntext}[1]{\noindent##1}%
  \ifdefined\H@@footnotetext\H@@footnotetext{#1}\else\@footnotetext{#1}\fi
  \endgroup}
\makeatother

\title{Assessing feature scorer results on high-dimensional datasets with t-SNE}

\author[a]{Bruno Iochins Grisci}
\author[b]{Mario Inostroza-Ponta}
\author[a,c,d]{Márcio Dorn\thanks{Correspondence to: Federal University of Rio Grande do Sul, Center for Biotechnology, Structural Bioinformatics and Computational Biology Lab - SBCB, Av.~Bento Gonçalves 9500, 91501-970, Porto Alegre, RS, Brazil.\newline
\textit{E-mail addresses:} \href{mailto:bigrisci@inf.ufrgs.br}{bigrisci@inf.ufrgs.br} (B.I. Grisci), \href{mailto:mario.inostroza@usach.cl}{mario.inostroza@usach.cl} (M. Inostroza-Ponta), \href{mailto:mdorn@inf.ufrgs.br}{mdorn@inf.ufrgs.br} (M. Dorn).\newline
\textit{URLs:} \url{https://brunogrisci.github.io/} (B.I. Grisci), \url{https://informatica.usach.cl/academico/mario-inostroza-ponta/} (M. Inostroza-Ponta), \url{https://sbcb.inf.ufrgs.br/} (M. Dorn).\newline
\textit{ORCID iDs:} \href{https://orcid.org/0000-0003-4083-5881}{0000-0003-4083-5881} (B.I. Grisci), \href{https://orcid.org/0000-0003-1295-8972}{0000-0003-1295-8972} (M. Inostroza-Ponta), \href{https://orcid.org/0000-0001-8534-3480}{0000-0001-8534-3480} (M. Dorn).}}

\affil[a]{Institute of Informatics, Federal University of Rio Grande do Sul, Av.~Bento Gonçalves 9500, Porto Alegre, 91501-970, RS, Brazil}
\affil[b]{Computer Engineering Department, University of Santiago de Chile, Av.~Libertador Bernardo O'Higgins, Santiago, 9170022, RM, Chile}
\affil[c]{Center for Biotechnology, Federal University of Rio Grande do Sul, Av.~Bento Gonçalves, Porto Alegre, 91501-970, RS, Brazil}
\affil[d]{National Institute of Science and Technology - Forensic Science, Porto Alegre, RS, Brazil}

\date{}

\begin{document}

\maketitle

\bottomnotice{\fcolorbox{black!55}{black!4}{%
\begin{minipage}{\dimexpr\linewidth-2\fboxsep-2\fboxrule\relax}
\small
\textbf{How to cite:} B.I. Grisci, M. Inostroza-Ponta, M. Dorn, Assessing feature scorer results on high-dimensional datasets with t-SNE, \emph{Neurocomputing} 652 (2025) 130561. \url{https://doi.org/10.1016/j.neucom.2025.130561}

\medskip
\begin{minipage}[c]{0.15\linewidth}
\href{https://creativecommons.org/licenses/by-nc-nd/4.0/}{\includegraphics[width=\linewidth,alt={Creative Commons Attribution-NonCommercial-NoDerivatives 4.0 International license badge}]{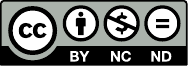}}
\end{minipage}\hfill
\begin{minipage}[c]{0.82\linewidth}
\textbf{License:} \copyright~2025. This manuscript version is made available under the CC-BY-NC-ND 4.0 license \url{https://creativecommons.org/licenses/by-nc-nd/4.0/}
\end{minipage}
\end{minipage}}}

\begin{abstract}
While vast literature on high-dimensional data visualization is available, there are not many works regarding the visualization of feature scorers and their results.
Feature scorers are algorithms that assign numerical importance to each feature of multi-dimensional datasets.
These importance scores can be used in several applications, such as feature selection, knowledge discovery, and machine learning interpretability.
There are several feature scorers to choose from, and often no single metric or ground truth is available to guarantee the quality of their results.
In this scenario, visualization can become valuable to support the decision of which method to choose and how good its results are.
For this goal, this work presents ``weighted t-SNE.''
It modifies the relationship between data points in the embedded 2D space to reflect the importance of each dimension of the original datasets as assessed by a feature scorer.
This research discusses how to implement weighted t-SNE, proposes the silhouette coefficient as a numerical evaluation of the results, and shows several examples of its use in practice.
Synthetic and real-world tabular datasets are used in the experiments together with nine feature scorers, ranging from Mutual Information to neural networks.
Each feature scorer produces unique visualizations, and weighted t-SNE can be used to compare and choose the one that better suits a given dataset and task.
Weighted t-SNE can also visually show the importance of features learned by machine learning models and help us see how they are organizing the data, increasing their interpretability.
\end{abstract}
\nopagebreak[4]

\section{Introduction}\label{sec1}

High-dimensional data refers to datasets whose dimensionality is comparable to or even higher than their sample size.
It is often present in tables containing information regarding the sciences or business \cite{sohns2021attribute}.
Some examples are data of genes expression from cancer patients \cite{feltes2019cumida, feltes2021benchmarking}, SNPs in forensic science \cite{avila2019forensic}, COVID-19 patients' hemogram exams \cite{avila2020hemogram, formica2020complete, dorn2021comparison}, astrophysics \cite{lyon2016fifty}, and sales from e-commerce \cite{sakar2019real}.
One of the key aspects of those large tabular datasets is the presence of dozens to millions of columns, each representing a \emph{feature}, \emph{input}, or \emph{dimension} of the data being studied \cite{grisci2021relevance}.

Data analyses such as clustering, classification, regression, or outlier detection rely first on the transformation, selection, and removal of features \cite{sohns2021attribute}.
Each feature represents one attribute of a data sample or instance.
Dimensionality reduction and feature selection allow large datasets to become manageable by keeping only the relevant information, making the data more easily understandable, and freeing storage space.
Feature scorers are a large and diverse group of algorithms that play a crucial role in this process.
They use distinct strategies, metrics, and criteria to determine the importance score of each feature.
Once obtained, these scores are numerical values that can be used to filter the data or even for knowledge discovery.
Because several machine learning models can output feature importance as well, these scores can even help with machine learning interpretability \cite{grisci2021relevance, molnar2022interpretable, shrikumar2017learning}.
It is important to note that the feature importance is always concerning a specific task and data, and it should not be assumed that this value will be the same if the context for the feature changes.

Since there are several feature scorers and no single metric guarantees the quality of the results, visualization becomes a valuable tool to help in the method selection and show how good the results are.
An inadequate set of scores can harm the subsequent steps of data analysis.
Moreover, the visualization of a low-dimensional embedding of the high-dimensional data using methods like the t-Distributed Stochastic Neighbor Embedding (t-SNE) \cite{maaten2008visualizing} is already an established aspect of exploratory data analysis \cite{sohns2021attribute}.
Visual cluster analysis often employs these techniques to project the high-dimensional data into 2D scatterplots, assuming that the projection faithfully matches the actual clusters \cite{xia2021revisiting}.

In this scenario, visualizing the clusters affected by the feature importance scores in a 2D scatterplot would be a valuable tool for comparing feature scorers.
The algorithms with better results (a more significant and representative set of scores) would lead to better clusters in the projection.
However, due to the challenges of visualizing high-dimensional tabular data and the algorithms being overlooked as a pre-processing step, they are usually applied as a black-box at the beginning of an analysis \cite{may2011guiding}.
In the vast literature on high-dimensional data visualization, few works are interested in displaying the effects of feature scoring \cite{grisci2021relevance}.
Most of the research interest is in the evaluation of feature selection \cite{mukherjee2017t, li2016extracting, xi2018feature} or the use of visualization to allow the user to interact with the selection process \cite{dennig2019fdive, artur2019novel, krause2014infuse}.

\begin{figure}[tbp]
\centering
\begin{subfigure}[t]{0.45\linewidth}
  \centering
  \fbox{\includegraphics[width=.9\linewidth, height=.9\linewidth, alt={t-SNE scatterplot of the synthetic regression dataset without feature importance; samples with negative (blue) and positive (red) target values are mixed}]{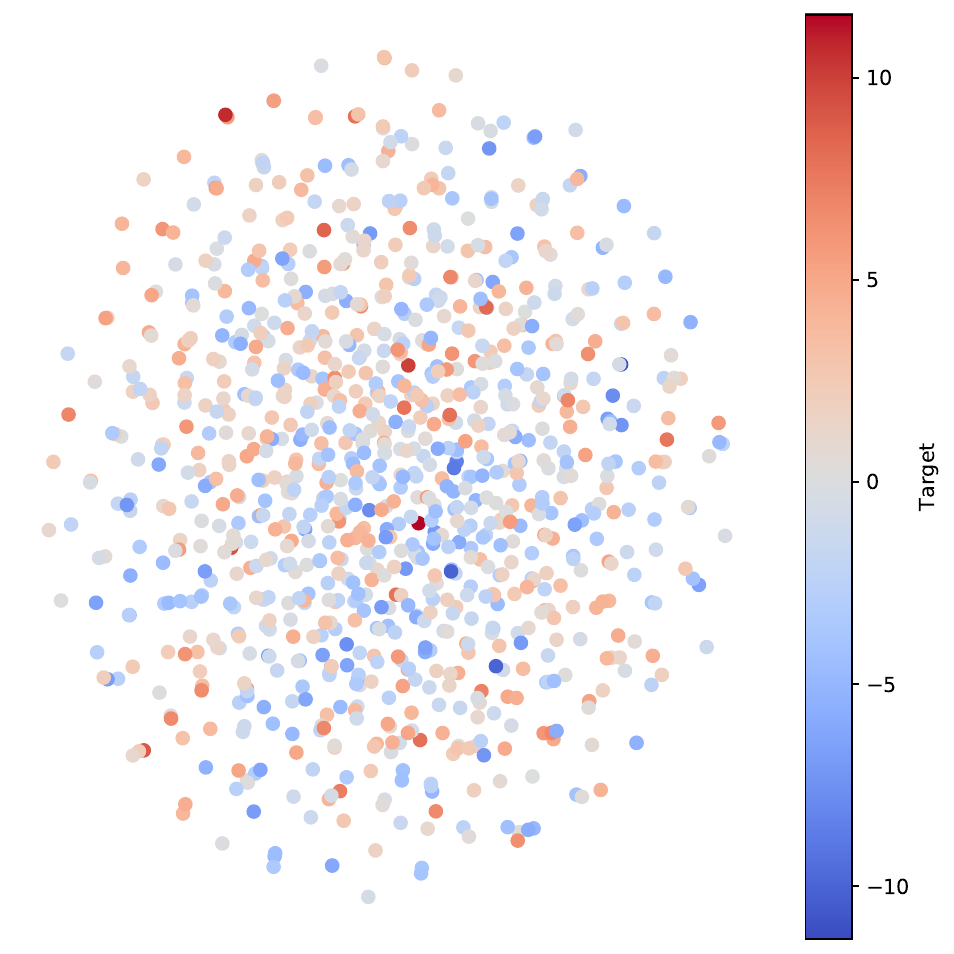}}
  \caption{No feature importance}
  \label{fig:wtsneregressionnoselection}
\end{subfigure}
\begin{subfigure}[t]{0.45\linewidth}
  \centering
  \fbox{\includegraphics[width=.9\linewidth, height=.9\linewidth, alt={Weighted t-SNE scatterplot of the synthetic regression dataset using the feature importance learned by a neural network; samples are sorted from blue (negative) to red (positive) target values}]{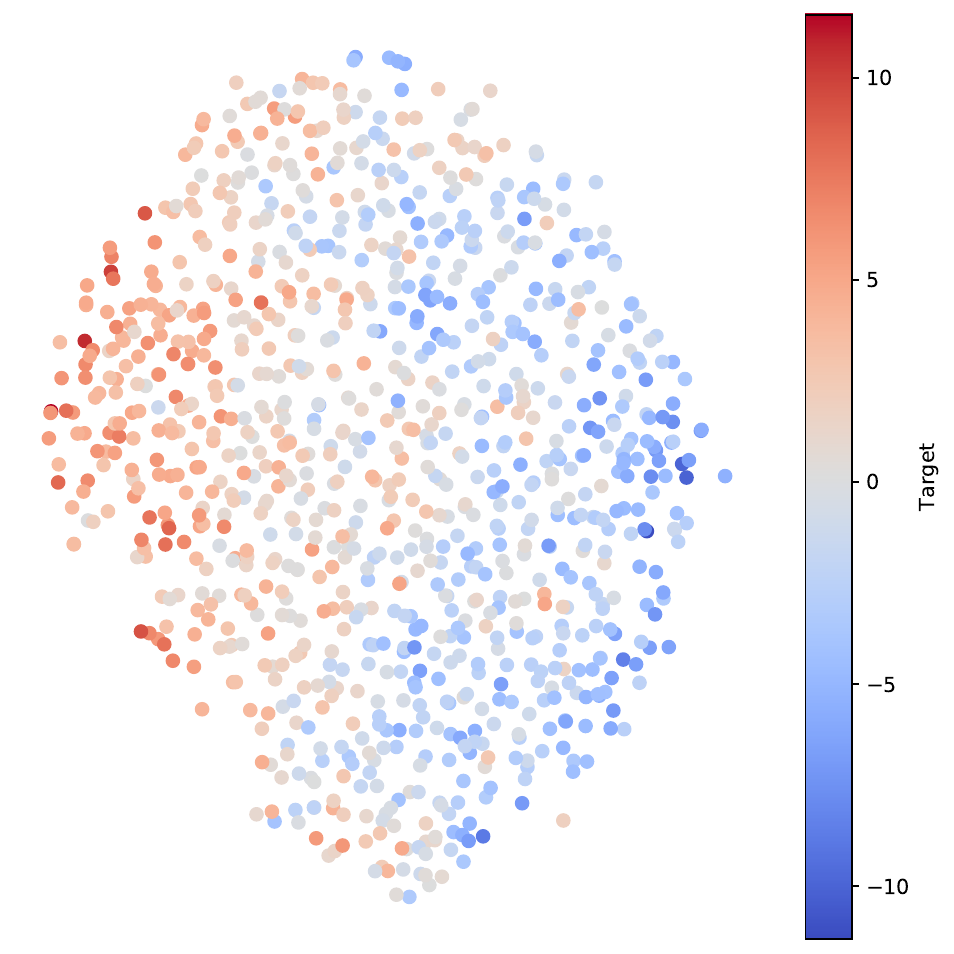}}
  \caption{Feature importance as learned by a neural network}
  \label{fig:wtsneregressionrelagg}
\end{subfigure}
\caption{Comparison between the visualization using weighted t-SNE of a synthetic regression dataset before and after considering the feature importance as learned by a neural network. Each figure is the visualization of the 2D embedding of a regression dataset (negative target values in blue, positive values in red) composed mostly of irrelevant features. On the left is the regular visualization, in which the samples are mixed. On the right, it is possible to visualize how the feature importance learned by a neural network disentangles the samples. This experiment is further detailed in Section~\ref{sec:wtsneexperiments}. Interactive versions of all the plots in this publication are available at \url{https://sbcblab.github.io/wtsne/}.}
\label{fig:teaser}
\end{figure}

To tackle this problem, we propose an extension of t-SNE to visualize the results from feature scoring algorithms: \emph{weighted t-SNE}.
It modifies the relationship between data points to reflect the importance of each dimension.
The desired outcome is that relevant features have more influence on the position of the points in the final projection.
Weighted t-SNE can be used as an inspection tool to compare and choose feature scorers in different datasets.
Additionally, combined with machine learning, it could better represent the patterns learned by the models.
Fig.~\ref{fig:teaser} shows an example of this application.
In Fig.~\ref{fig:wtsneregressionnoselection}, a regular projection is made with t-SNE of a synthetic dataset.
After training a neural network to perform regression on this dataset, the importance score of each feature was extracted from the model and used to create the projection shown in Fig.~\ref{fig:wtsneregressionrelagg}.
It is possible to see how the feature importance learned by the neural network sorted the samples by comparing the two scatterplots.
This work focuses on continuous or numerical features, which can also be ordinal or categorical.
The requirement for running t-SNE on the data is that it must be meaningful.
For instance, a suitable distance measure for the high-dimensional space must exist.
This requirement is often not reached if the feature types are mixed.

A preliminary version of weighted t-SNE appeared in the research by \citet{grisci2021relevance}.
It focused in \emph{relevance aggregation}, a method for neural network interpretability.
In that study, authors used weighted t-SNE to visualize the patterns learned by the networks trained on tabular data.
Despite those results showing the potential of this kind of visualization, \citet{grisci2021relevance} only used weighted t-SNE to validate the main method, relevance aggregation.
\citet{junior2022interpretability} and \citet{zhao2023evaluating} later used weighted t-SNE to visualize neural network results for absenteeism prediction and to evaluate data transformation techniques on software defect prediction models, respectively.
\citet{lu2022variable} and \citet{ma2023dimension} also explored the idea of scaling features for t-SNE visualizations, albeit for distinct purposes.
The present research presents a thorough description, critical analysis, and discussion of an updated weighted t-SNE, together with a large set of experiments on several datasets and feature scorers.

The remainder of this paper has the following organization.
Section~\ref{sec:background} presents and discusses feature scorers, related visualization methods, and the silhouette coefficient metric.
Section~\ref{sec:wtsne} describes the proposed weighted t-SNE.
Section~\ref{sec:wtsneexperiments} details the experiments and results, as well as a discussion on machine learning interpretability.
Finally, Section~\ref{sec:conclusion} concludes this work.

\section*{Statement of significance}
\addcontentsline{toc}{section}{Statement of significance}

Understanding the effectiveness of feature scorers is a fundamental challenge in machine learning and data science.
While prediction accuracy is commonly used to evaluate feature scoring and selection, it does not fully capture how feature importance influences data representation.
Visualization techniques such as t-SNE provide a powerful tool for exploring high-dimensional data, yet they do not inherently account for feature relevance.
This paper introduces weighted t-SNE, a novel extension that integrates feature importance scores into the dimensionality reduction process, enabling direct inspection of how different scorers structure data.

\paragraph{Problem:} Selecting the best feature scorer is often difficult due to the lack of ground truth for feature importance.
Existing evaluation methods typically rely on classification or regression performance, but these metrics do not reveal how well a scorer identifies relevant features.
Additionally, feature scoring is distinct from feature selection.
While feature selection reduces dimensionality by removing features, feature scoring assigns relative importance to all features, and this process has received less attention from visualization tools.
Without proper inspection methods, irrelevant or misleading feature scores may negatively impact downstream tasks, such as model interpretability, classification, or regression.
For high-dimensional data, the manual inspection of thousands of numerical values is infeasible.
Moreover, two models with identical predictive accuracy may assign different importance scores to features, making numerical comparison difficult, a challenge that weighted t-SNE helps resolve by providing an intuitive visual representation.

\paragraph{What is already known:} Traditional t-SNE or similar tools like UMAP are widely used for exploratory data analysis, often assuming that clusters in the visualization correspond to meaningful patterns in the data.
However, it does not incorporate feature importance, making it unsuitable for evaluating feature scorers.
Visualization techniques for feature selection exist, such as RadViz, Self-Organizing Maps (SOM), and SmartStripes, but they focus on selecting a subset of features rather than analyzing the distribution of importance scores across all dimensions.
Machine learning interpretability tools such as SHAP and LIME provide insights into model decisions but do not offer direct visualization of feature scorers across an entire dataset.

\paragraph{What this paper adds:} Weighted t-SNE is a visualization tool specifically designed for analyzing feature scorers, bridging the gap between feature scoring and visual interpretability.
It allows users to visually assess how different scoring methods affect data representation.
It can be used in labeled datasets, providing insight into how feature importance influences different types of machine learning models.
Beyond classification accuracy, weighted t-SNE has multiple practical applications: It helps revealing outliers in high-dimensional datasets where irrelevant features might otherwise obscure them.
It visually tracks how a model transforms feature representations internally, making it easier to analyze learned feature importance compared to raw numerical scores.
Weighted t-SNE is a valuable tool for machine learning and data science pipelines.
By allowing visualization of feature scorers, it helps users select the most suitable scorer for a given dataset, ultimately improving downstream tasks such as feature selection, classification, or regression.
By incorporating feature importance into dimensionality reduction, weighted t-SNE provides a new perspective on feature scoring, offering insights that neither accuracy-based evaluation nor traditional visualization methods provide.

\section{Related work}\label{sec:background}

\subsection{Feature scoring}\label{subsec:featurescoring}

Feature scorers are algorithms or methods that, given a dataset, assign to each of its features (dimensions) a score measuring its importance or relevance based on specific criteria or metrics.
For example, given a tabular dataset with two classes and $n$ features, a standard feature scorer would return $n$ scores corresponding to each dimension of the dataset in a way that those features with larger scores are better at distinguishing the two classes.
These importance scores can be used in several applications, such as dimensionality reduction, feature selection, knowledge discovery, outlier detection, and machine learning interpretability \cite{lazar2012survey, ang2016supervised}.

Because there is a large overlap in methods and application, it is important to distinguish between feature scoring and feature selection as two related but distinct tasks:

\begin{itemize}
\item \textbf{Feature scoring:} This task focuses on assigning numerical importance scores to features based on their relevance to a specific task (e.g., classification, regression, statistical significance, domain relevance).
The scores provide a quantitative measure of how useful or relevant each feature is, but they do not directly reduce the dimensionality of the dataset.
Feature scoring allows the comparison of features and enables feature ranking and feature selection.
\item \textbf{Feature selection:} This task involves using the importance scores from feature scoring (or other selection criteria) to select a subset of features for downstream tasks.
Typically, a threshold is applied to the scores, and features below the threshold are removed from the dataset.
The goal of feature selection is to reduce dimensionality, improve computational efficiency, and enhance the interpretability of the data.
Usually, after feature selection the features are used as equally important.
\end{itemize}

For example, if the features are ranked by their scores, a threshold can be set to remove all features below it from the dataset.
After the reduction, the remaining features will ideally be more informative for training predictors, can be more easily analyzed by humans, or take less processing time and memory space for algorithms.
Moreover, knowing which features are more important and which are less according to given criteria can be exploited for analyzing and interpreting the data \cite{grisci2019neuroevolution, grisci2021relevance}.
Further differences between feature scoring and feature selection are discussed in the experiments of Section~\ref{sec:wtsneexperiments}.

A feature on its own may seem irrelevant, but when combined with other features, it could become highly relevant \cite{gheyas2010feature, ang2016supervised}.
Ideally, the features with high scores should be strongly relevant, but sometimes they can be weakly relevant if non-redundant features are helpful in improving the predictions.
Meanwhile, the irrelevant, redundant, or noisy features receive lower scores \cite{ang2016supervised}.
Feature scorers are considered \emph{univariate} if each feature is analyzed independently, and \emph{multivariate} if the interactions and correlations between features are taken into account.
These methods are usually divided in the five groups presented below \cite{ang2016supervised, lazar2012survey} with examples that will be used in the experiments of Section~\ref{sec:wtsneexperiments}.

\begin{description}
\item[Filter:] these methods take into account the inherent characteristics of the data in conjunction with evaluation criteria (information, distance, consistency, dependency, etc.), and are not limited to being classifiers.
Most filters treat the problem as a ranking problem and are univariate.
They do not rely on specific learning algorithms, providing more general solutions that different classifiers can use.
Filters are faster and more computationally efficient than the other groups of selectors, but they may ignore relationships between different features or the effects that they have when combined \cite{lazar2012survey, ang2016supervised}.
An example is the Kruskal--Wallis one-way analysis of variance \cite{kruskal1952use}, a non-parametric statistical test for discovering if samples originate from the same distribution.
The Kruskal--Wallis test can be used as a simple filter feature scorer.
In this case, every feature is individually tested to check if it belongs to the same distribution for the different groups (or classes), and their scores come from the statistical significance (the more certain of a feature belonging to different distributions for each group, the larger its importance) \cite{grisci2019neuroevolution, grisci2021relevance}.
Another example is the Minimum Redundancy Maximum Relevance (mRMR) \cite{peng2005feature}, a filter algorithm that iteratively selects the features that maximize mutual information to the target class and minimize the redundancy regarding all the features previously selected.
Mutual Information is a measure of how much information one random variable has in relation to another variable \cite{vergara2014review}.
It is possible to quantify a feature's relevance based on how much information it holds with respect to the target class.
As a final example, ReliefF scores each feature according to how different they are from nearby instances \cite{kononenko1997overcoming}.
For all samples in a dataset, it looks for the k-neighbor samples of each class and then weights how much each feature differs between samples, thus being an efficient algorithm that considers feature-correlation \cite{robnik2003theoretical}.
\item[Wrapper:] the scoring is made using some optimization algorithm, often a metaheuristic, and then wrapping a classifier (or regressor) around the selected features \cite{ang2016supervised}.
The accuracy is the criteria of evaluation.
The set of most discriminating features is found by minimizing the classification error, which often results in better accuracy than filters.
However, wrappers are highly dependent on the learning algorithm being used as a classifier, so the solutions do not generalize.
There is no guarantee that the quality of performance of the selected features will be transferable for other classifiers.
Wrappers are more likely to suffer from overfitting and to present huge computational costs since the training of the classifier needs to be performed for each new subset being evaluated \cite{lazar2012survey, ang2016supervised}.
\item[Embedded:] the scoring process is integrated within the learning algorithm and is conducted simultaneously with the classification (or regression).
This approach is more efficient than wrappers because it avoids the repetition of training a classifier and is less prone to overfitting while achieving similar performance.
Despite this, the computational complexity in high-dimensional data remains a challenge. \cite{ang2016supervised}.
Examples of a simple embedded algorithm are the least absolute shrinkage and selection operator (Lasso) \cite{fonti2017feature} and decision trees, but even support-vector machines (SVM) and neural networks can be treated as embedded scorers.
In these cases, some metric is used to evaluate the impact of each feature inside the model.
This idea is further discussed in Section~\ref{subsec:machinelearning}.
\item[Ensemble:] multiple classifiers or regressors are trained and their predictions are combined to achieve a better result.
A popular example is the random forest \cite{breiman2001random}, which is an ensemble of decision trees.
\item[Hybrid:] a combination of different methods (which may or may not belong to the same group), different scorer algorithms, or different criteria, in an attempt to capitalize on their unique strengths.
The most common combination is that of wrappers and filters \cite{ang2016supervised}.
\end{description}

Unlike filters, wrapper and embedded methods rely on the accuracy of a classifier during evaluation, while also utilizing strategies to search the feature space to complete the selection process \cite{lazar2012survey}.
Hybrid methods usually have the best results, as they combine the strengths of the other approaches, decrease the computational costs narrowing the search space, and reduce overfitting \cite{ang2016supervised}.
Feature scorers are usually global in that the features' importance is computed over the entire data space \cite{kriegel2012subspace}.
If the importance must be computed locally (different subspaces of the data have different feature scores), an alternative method is subspace clustering \cite{kriegel2012subspace, parsons2004subspace}.
More details on how particular feature scorers work, including mathematical formulation, are present in the \hyperref[app:supp]{Supplementary material}.

Feature scoring can be applied to several domains, some of them listed in Section~\ref{sec1}.
One of the most critical and prevalent applications of feature scorers is in gene selection.
Due to its relevance, it will be described in more detail in this section and used in some experiments in Section~\ref{sec:wtsneexperiments}.
Feature selection is applied to genomic data (a popular example being gene expression from microarray experiments) to discover subsets of genes capable of separating samples from different populations \cite{bolon2014review, feltes2019cumida, grisci2018microarray, feltes2018perspectives}.
Genomic data can be effectively used for reliable cancer diagnosis, prognosis, or clinical treatment, but it often contains irrelevant, redundant, or noisy values \cite{boulesteix2008evaluating, ang2016supervised}.
The discovery of genes capable of differentiating samples from different target annotations (the samples' classes) is an essential aspect in the analysis of microarray data \cite{lazar2012survey}.
These informative genes are used in the identification of diseases or as potential drugs targets \cite{lazar2012survey}.
In contrast to regular feature selection, the elimination of redundant features can lose informative genes with highly correlated expressions in gene selection.

Most gene selection studies are focused on filters due to their efficiency and generability \cite{ang2016supervised}.
Nevertheless, many challenges persist, such as the presence of technical defects in the experiments \cite{ang2016supervised}, class imbalance \cite{powers2005short, lazar2012survey}, data bias \cite{grisci2023use}, stability \cite{barbieri2024analysis}, and the difference between the several analyses standards.
More importantly, retrieving domain-specific information from the data is not an easy task, and determining the relevance or redundancy of a feature is difficult, leading to unexpected biases and mistakes in conclusions.
Despite a large number of methods available, there is still much room for innovations and improvements \cite{ang2016supervised, feltes2018perspectives, grisci2019neuroevolution, barbieri2024analysis}.
Gene selection is an open problem with many challenges and new alternatives rising, with several methods for differentially expressed genes discovery being only slightly different among them \cite{lazar2012survey, SHIM2022100331}.
There is also the need for visualization to make sense of clusters of samples in these large high-dimensional datasets \cite{kobak2019art, diaz2021review, mramor2007visualization, grisci2021relevance}.
In this context, creating analytics tools for the comparison and inspection of feature scorers is fundamental.

\subsection{Visualization methods}\label{subsec:otherviz}

Several algorithms are used to visualize multi-dimensional data in the 2D or 3D spaces.
One of the most common approaches is to compute the Principal Component Analysis (PCA) of the data and display the first components \cite{pearson1901lines}.
The PCA extracts the principal components of the data through the covariance between dimensions so that each component is a linear transformation of the original dimensions that maximizes the variance information.
Unlike t-SNE, PCA is deterministic and tends to preserve the distance of the points in the lower dimensions, keeping the global structure of the data but losing its local structure.
PCA also cannot faithfully represent non-linearity in the data, which would defeat the purpose of using it to visualize feature selectors that are non-linear.
As discussed by \citet{kobak2019art}, PCA can be used to initialize the positions of the points for the t-SNE.

A more recent option to t-SNE is the Uniform Manifold Approximation and Projection (UMAP) \cite{mcinnes2018umap}.
Previous experiments conducted by \citet{xia2021revisiting} showed that UMAP and t-SNE are the best dimensionality reduction methods for cluster identification and membership identification.
UMAP uses a sampling-based approach to optimize repulsive forces between data points \cite{kobak2019art}.
While UMAP is said to be faster than regular t-SNE, it was observed that it is slower than optimized implementations of t-SNE and does not present better results for transcriptomics data \cite{kobak2019art}.

A distinct approach for non-linear multi-dimensional 2D visualization is RadViz \cite{hoffman1997dna}.
This algorithm places each feature as equally spaced positions around the perimeter of a circle to serve as ``anchor points''.
The data samples are represented as points inside the circle, their positions determined by the values of their features.
Each data point is connected to the anchor points as if by a spring whose stiffness is proportional to the value of the corresponding feature (scaled between zero and one).
The final data point is in the equilibrium position for all its springs, and points close to a variable anchor have a higher value than for the other variables.
RadViz has been used to visualize gene expression data \cite{mramor2007visualization} and has also been expanded into attribute-RadViz \cite{artur2019novel} as a tool for user-guided feature selection.
One disadvantage of RadViz is that its complexity grows and clarity decreases as the number of dimensions to be represented increases, which is an impediment in the use-cases with thousands of features present in Section~\ref{sec:wtsneexperiments}.

Other methods of user-guided feature selection are the FDive \cite{dennig2019fdive}, which uses Self-Organizing Maps (SOM) \cite{kohonen1982self} to determine feature relevance, and the SmartStripes \cite{may2011guiding}, which uses a color data table to convey information about the features.
Once again, although the topics are closely related, the focus of the present work is the visualization of feature scoring, not feature selection.
Another approach that resembles the results of SmartStripes but was developed to show the importance scores of features is the table heatmap \cite{grisci2021relevance}.
In this approach, the original dataset is displayed, and each cell is colored with hue value-intensity proportional to the feature importance.
However, table heatmap is unfeasible for large datasets.

\subsection{t-distributed stochastic neighbor embedding}\label{subsec:tsne}

The t-Distributed Stochastic Neighbor Embedding (t-SNE) \cite{maaten2008visualizing} is one of the most popular algorithms for visualizing high-dimensional datasets.
It is a non-linear dimensionality reduction technique that allows the embedding of high dimensions in a much smaller space, usually 2D, to be used in plots.
The core idea of t-SNE is to use methods from machine learning to optimize the embedding in a way that preserves the local neighborhood of points.
Because of that, the local structure of points clustered together is usually kept in the embedding space.
However, the global structure of the dataset can be lost.
It is important to note that in t-SNE the low-dimensional embedding space does not necessarily represent any meaningful dimension of the original space, which is different from other methods, some of them, like PCA, described in Section~\ref{subsec:otherviz}.

The algorithm is now presented in more detail, based on the definitions of \citet{maaten2008visualizing} and \citet{kobak2019art}.
Several practical improvements were proposed in the past years that made t-SNE more efficient regarding memory consumption and computing time \cite{kobak2019art, polivcar2019embedding, van2014accelerating, linderman2019fast, macosko2015highly}.
Considering that we desire to visualize a $N$-dimensional dataset $\mathbf{X} \in \mathbb{R}^N$, t-SNE will create a lower dimensional embedding $\mathbf{Y} \in \mathbb{R}^n$, in which $n \ll N$.
The desired result is that two points $\chi_i, \chi_j \in \mathbf{X}$ that are close to each other should be represented by lower-dimensional points $\gamma_i, \gamma_j \in \mathbf{Y}$ that are also close to each other.
The similarities between the original and the embedding space points can be modeled as probability densities.
Similarities in $\mathbf{X}$ can be computed by a Gaussian distribution as shown in Eq.~\eqref{eq:tsne1}.
These are conditional probabilities that can be symmetrized to obtain joint probabilities $p_{ij}$ as in Eq.~\eqref{eq:tsne2}.

\begin{equation}
p_{j \rvert i} = \frac{\exp(-\|\chi_i - \chi_j\|^{2} /2 \sigma^2_i)}{\sum_{k \neq i} \exp(-\|\chi_i - \chi_k\|^2 / 2 \sigma^2_i)}
\label{eq:tsne1}
\end{equation}

\begin{equation}
p_{ij} = \frac{p_{j \rvert i} + p_{i \rvert j}}{2}
\label{eq:tsne2}
\end{equation}

The similarities in the embedding space $\mathbf{Y}$ are computed similarly.
However, instead of the Gaussian distribution, the Student's t-distribution is used, as shown in Eq.~\eqref{eq:tsne3}.
This change allows some distances to be less faithfully preserved in the embedding thanks to the fatter tails of the t-distribution.
This is needed because there is less available space in the embedding (because there are fewer dimensions), so without this freedom from the fatter tails, the points would end up crowded together.

\begin{equation}
q_{ij} = \frac{(1 + \|\gamma_i - \gamma_j\|^2)^{-1}}{\sum_{k \neq l} (1+ \|\gamma_k - \gamma_l\|^2)^{-1}}
\label{eq:tsne3}
\end{equation}

Now that the distributions $\mathbf{P}$ and $\mathbf{Q}$ are defined from $p_{ij}$ and $q_{ij}$, the goal of t-SNE is to make $\mathbf{P}$ and $\mathbf{Q}$ as similar to each other as possible.
The Kullback--Leibler (KL) divergence is an alternative to measure the similarity between two probability distributions, shown in Eq.~\eqref{eq:tsne4}.
The divergence can be seen as a cost $KL(P \| Q)$ to be minimized.
Because it is differentiable, it can be optimized with gradient descent.

\begin{equation}
KL(P \| Q) = \sum_{ij} p_{ij} \log \frac{p_{ij}}{q_{ij}}
\label{eq:tsne4}
\end{equation}

The Gaussian kernels centered over each data point in $\mathbf{X}$ have a bandwidth $\sigma_i$ (Eq.~\eqref{eq:tsne1}).
As the data density varies, each data point has its optimal value for $\sigma_i$.
For instance, a larger value of $\sigma_i$ will be adequate in sparser regions.
These values are defined as the perplexity value, and the Gaussian kernels fit the nearest neighbors within one standard deviation of the probability density.
The perplexity is a tunable parameter, as the user should determine how many close neighbors each point has.
When it was first proposed, it was stated that the performance of t-SNE is robust to changes in the perplexity between the typical values of $5$ and $50$ \cite{maaten2008visualizing}.
However, the choice of perplexity and number of iterations during optimization can significantly impact the result of t-SNE \cite{wattenberg2016how}.
Another point raised is that there is no guarantee that distances between well-separated clusters are meaningful without fine-tuning the perplexity.
A rule-of-thumb that resulted in good empirical results is to set the perplexity to $max(30, m/100)$, $m$ being the number of samples in the dataset \cite{kobak2019art}.
This work employs this strategy to avoid choosing an arbitrary perplexity.
One issue of SNE-like methods is that information on scales beyond the chosen perplexity tends to be excluded.
One possible solution is using generalized similarities, computed as averages of softmin ratios with varying bandwidths, encompassing all neighborhood sizes.
The inclusion of different scales causes a slight increase in computational complexity, but these novel multi-scale similarities do not require additional parameters \cite{lee2015multi}.

Some works have already used the t-SNE as a feature extractor for several applications \cite{mukherjee2017t, li2016extracting, xi2018feature}.
In those applications, the features of the low-dimensional embeddings obtained from t-SNE are used instead of the original features from the high-dimensional data \cite{mukherjee2017t}.
The t-SNE, among other methods, can also be straightforwardly used to visualize feature selection results.
This requires the creation of two plots, one with the complete set of features and another with only the subset of selected features \cite{chen2018flight}.
These are not the goals of the present work, which uses t-SNE to visualize the impact of feature importance scores and not to create the scores themselves or to only show a subset of features with no regard to their importance.
The difference will be further discussed in Section~\ref{sec:wtsneexperiments}.

\subsection{Silhouette coefficient}\label{subsec:silhouette}

The silhouette coefficient is a metric used to indicate the separation between clusters of points \cite{rousseeuw1987silhouettes}.
The silhouette measures how close each point in a cluster is to the points in the nearest clusters.
Its range is $[-1.0, 1.0]$.
A value close to $1$ indicates that the point is far from other clusters and is well matched to its assigned cluster, a value of $0$ suggests the point is near the decision boundary between clusters, and a value close to $-1$ indicates that the point is closer to some other cluster than the one it is assigned and may have been wrongly assigned to its cluster.
The silhouette coefficient is the average silhouette over all samples.

The silhouette coefficient in a visualization of a dataset can be used as an approximation on how well distinct classes are separated in the final plot \cite{artur2019novel}.
In this case, the silhouette coefficient is computed using the actual class labels of each data point.
For instance, if the points in the low-dimensional embedding space have a silhouette coefficient close to zero or less, this embedding failed to divide the classes.
Of course, for this analysis to make sense, it always needs to consider the silhouette coefficient of the original high-dimensional space, as the embedding is not expected to segregate classes that do not have clear boundaries in the original space.

Considering that there are two or more clusters and that $\phi \in C_i$ is a data point assigned to the cluster $C_i$, $a(\phi)$ is, as defined in Eq.~\eqref{eq:silhouette1}, the mean distance between $\phi$ and every other data point in $C_i$.
$\lvert C_i \rvert$ is the number of points in $C_i$ and $d(\phi, \psi)$ is the distance $d(\cdot)$ between the points $\phi, \psi \in C_i$.
The value $a(\phi)$ measures how well $\phi$ belongs to $C_i$ (the smaller the value the better).

\begin{equation}
a(\phi) = \frac{1}{\lvert C_i \rvert - 1} \sum_{\psi \in C_i, \phi \neq \psi} d(\phi, \psi)
\label{eq:silhouette1}
\end{equation}

Analogously, the value $b(\phi)$, defined in Eq.~\eqref{eq:silhouette2}, is the smallest mean distance between the point $\phi$ and all points in a cluster $C_j \neq C_i$.
The value of $b(\phi)$ measures the dissimilarity of $\phi$ to all points from other clusters.

\begin{equation}
b(\phi) = \min_{j \neq i} \frac{1}{\lvert C_j \rvert} \sum_{\chi \in C_j} d(\phi, \chi)
\label{eq:silhouette2}
\end{equation}

The silhouette of the point $\phi \in C_i$, called $s(\phi)$, is computed combining the values $a(\phi)$ and $b(\phi)$ as in Eq.~\eqref{eq:silhouette3}.
The $s(\phi)$ will be close to $1$ when $a(\phi) \ll b(\phi)$, meaning that $\phi$ has a low dissimilarity to its own cluster (small $a(\phi)$) and a high dissimilarity to the other clusters (large $b(\phi)$).

\begin{equation}
s(\phi) =
\begin{cases}
  \frac{b(\phi) - a(\phi)}{\max\{ a(\phi), b(\phi) \}} & \text{if $\lvert C_i \rvert > 1$} \\
  0 & \text{if $\lvert C_i \rvert = 1$}
\end{cases}
\label{eq:silhouette3}
\end{equation}

Finally, the silhouette coefficient $SC$ is defined as the mean $s(\phi)$ over the entire dataset.
This is shown in Eq.~\eqref{eq:silhouette6}, in which $C$ is the set of all clusters combined.

\begin{equation}
SC = \frac{1}{\lvert C \rvert} \sum_{\phi \in C} s(\phi)
\label{eq:silhouette6}
\end{equation}

\subsection{Trustworthiness}\label{subsec:trustworthiness}

The trustworthiness $T$ is used to measure how much the local patterns in a projection (for instance, the embedded 2D space from t-SNE) mirror the actual patterns in the high dimensional data \cite{espadoto2019toward}.
It is the proportion of points that are close in both the high dimensional and embedded space, computed according to Eq.~\eqref{eq:trustworthiness} \cite{espadoto2019toward, van2009learning}.

\begin{equation}
T(k) = 1 - \frac{2}{mk (2m - 3k - 1)} \sum^m_{i=1} \sum_{j \in \mathcal{N}_{i}^{k}} \max(0, (r(i, j) - k))
\label{eq:trustworthiness}
\end{equation}

In which for every sample $\phi_i$, $\mathcal{N}_{i}^{k}$ are its $k$ nearest neighbors in the embedded space, and every sample $\phi_j$ is its $r(i,j)$th nearest neighbor in the high dimensional space.
Nearest neighbors in the embedded space that are not neighbors in the high dimensional space are penalized proportionally to their rank in the high dimensional space \cite{van2009learning}.
The range of values for $T(k)$ is in $[0, 1]$, with $1$ being the best.
Following the work of \citet{espadoto2019toward} and \citet{10.2312:cgvc.20151234}, the parameter $k$ is set to be $k=7$.
The trustworthiness is used to assess the quality of the 2D projections generated by the visualization methods.

\section{Weighted t-SNE}\label{sec:wtsne}

The proposed weighted t-SNE visualization is a method for inspecting the results from feature scoring algorithms.
From a practical standpoint, we identified some desired properties that weighted t-SNE should have to be helpful for the analysis and comparison of feature scorers.
It should:
(i) account for non-linearity in the data;
(ii) be model-agnostic (work for any scorer);
(iii) use the importance score of each feature to produce a corresponding embedding;
and (iv) consider all features in the original dataset if they all have an importance score larger than zero.
Regarding the first item, many datasets contain complex structures in high-dimensional space, and the visualization must be able to reveal it in the projection space \cite{sohns2021attribute}.
Moreover, non-linear and local methods like t-SNE are preferred in cluster and membership identification \cite{xia2021revisiting}.
Considering all features with non-zero scores is important to distinguish between different scorers' behavior.
For instance, to distinguish a scorer that concentrates all importance on a few features from a scorer that spreads the relevance among more features, even those that, in practice, are irrelevant.
Moreover, this avoids the issue of having to define a hard threshold on what is a large enough score for a feature to be considered relevant.

Weighted t-SNE does not change the t-SNE algorithm itself but changes the relationship between data points (their distance) to reflect the importance of each dimension (feature).
It leverages that feature scorers offer some feature relevance value, indicating the feature's contribution to classes or clusters identification.
These values are the \emph{weights} that give \emph{weighted} t-SNE its name.
All weights are scaled so that the maximum importance value is one and the minimum value (for an irrelevant feature) is zero.

The projection regarding the importance of the features is accomplished by incorporating the weights in the distance calculation.
The weighted Euclidean distance $d_w(\cdot)$ \cite{horan1969multidimensional, carroll1970analysis} (Eq.~\eqref{eq:wed2}) is used in Eq.~\eqref{eq:tsne1} from Section~\ref{subsec:tsne} to compute the distance between two $N$-dimensional points $\phi$ and $\psi$, using the $N$ weights in $\omega$.

\begin{equation}
d_w(\phi,\psi,\omega) = \sum_{i=1}^N \sqrt{(\omega[i](\psi[i] - \phi[i]))^2}
\label{eq:wed2}
\end{equation}

Considering a $N$-dimensional dataset $\mathbf{X} \in \mathbb{R}^N$ with $m$ points, the difference of each dimension $i$ between the points $\phi$ and $\psi$ is multiplied by weight $\omega[i]$ (the importance score of that feature).
As shown in Eq.~\eqref{eq:wed3}, this is equivalent to multiplying each point by $\omega$ or to perform the Hadamard product (element-wise product) between a matrix composed of $m$ columns repeating the $N$ (the number of features) values of $\omega$ and $\mathbf{X}$ to obtain a new scaled dataset $\mathbf{X}' \in \mathbb{R}^N$ and use it to create the visualization.

\begin{equation}
\omega[i](\psi[i] - \phi[i]) = (\psi[i] \cdot \omega[i]) - (\phi[i] \cdot \omega[i])
\label{eq:wed3}
\end{equation}

Because the weights range from zero (irrelevant) to one (relevant), scaling the dimensions by their weights allows features of higher importance to have a greater impact on the distance between points.
The desired outcome is that relevant features influence the point position more in the final visualization.

Because the feature scaling can be performed before the embedding optimization, the additional computing cost of weighted t-SNE is negligible.
With the current improvements in the implementation of t-SNE, the code runs in a matter of seconds or minutes.
It should not significantly impact the total time of the data analysis pipeline.

It is worth noting that the method faces the same challenges as regular t-SNE regarding hyperparameter tuning \cite{wattenberg2016how}.
It should be possible to adopt this strategy to create variants of other visualization methods, for instance, a ``weighted UMAP''.
Such variants would inherit the advantages and drawbacks of their original algorithms.
This research focuses on t-SNE due to the advantages discussed in Sections~\ref{subsec:otherviz} and \ref{subsec:tsne} and because narrowing on a single variant allows for a more diverse set of experiments.

The main goal of weighted t-SNE is to provide an inspection tool to compare and choose feature scorers in different datasets.
Additionally, it can be one more item in the toolset of interpretable machine learning (Section~\ref{subsec:machinelearning}).
Weighted t-SNE was not designed for the individual visualization of specific features nor to score features themselves.
Instead, it provides a global visualization of the impact scorers have on the entire dataset and how it relates to data clusters.
As such, the analysis of the results of a feature scorer using weighted t-SNE only makes sense in a comparative manner, meaning that the visualization of the dataset after the scoring needs to be contrasted to a visualization of the dataset before the scoring.
The single 2D projection of weighted t-SNE for a feature scorer is meaningless if the 2D projection of the original data is not present, allowing the user to perceive or measure the difference between the two.

Weighted t-SNE focuses on feature scoring rather than feature selection, meaning it visualizes feature importance scores rather than selecting a subset of features.
While feature selection and scoring are related, they serve different purposes.
Feature selection typically receives more attention in visualization tools, whereas feature scoring lacks intuitive inspection methods.
Weighted t-SNE fills this gap but does not replace feature selection techniques.

Weighted t-SNE can be used with other visualization tools, specifically those designed to visualize individual features' importance, for a richer evaluation.
An example of this application was demonstrated using a preliminary version of weighted t-SNE and the table heatmaps to analyze neural networks on multiple datasets in a previous study \cite{grisci2021relevance}.
The same research also explored using weighted t-SNE to visualize different scores for different clusters of the same dataset.
It highlighted the learning of distinct importance scores for each data class by neural networks.

Recognizing that weighted t-SNE is an ``indirect way'' to evaluate feature scorers relying on a visual approximation of the data separation after scoring is essential.
However, evaluating feature selection by classification or regression metrics (training a model on the top $n$ features) is also indirect.
Even though feature scoring and feature selection are typical parts of classification or regression pipelines (mainly during preprocessing), these are not the same tasks, nor is the only purpose of feature selection to improve classification accuracy.
Thus, using an independent classifier to evaluate the selection of features, while recommended as additional validation, is not a perfect measure.
Evaluating feature selectors based on classification metrics would apply to feature selection algorithms (the top $n$ features would be used to train a classifier) but not necessarily to feature scorers, for which all features receive a score.
Although deeply connected, feature scoring and selection are different problems, as discussed in the next section.
The results from feature scorers can be converted into feature selection.
However, in the process, it is necessary to define a cutting threshold external to the algorithm and to discard features with scores under that threshold.
The scores themselves are not considered in this context.
Weighted t-SNE was designed with this issue in mind, thus visualizing all the scores.
Some of the experiments in the next section highlight these differences.
Another point is that not all feature scorers are based on classification or regression models, such as the Kruskal--Wallis Filter or the mRMR.
While an improvement in classification or regression metrics given a specific selection of features can indirectly measure the selection quality, it will not always be the case.
Machine learning models are known to learn biases \cite{roberts2021common} and shortcuts \cite{geirhos2020shortcut} from the data that lead to better metrics without necessarily discovering the rules or representations desired by the user.
In this scenario, a scorer that follows a different strategy (for instance, a statistical or information-based algorithm) may find a better set of relevant features from a knowledge discovery standpoint but that does not maximize the prediction accuracy of a model.
Because all these methods, including weighted t-SNE, approximate the measurement of the scoring quality, it is recommended that they be used together to combine their strengths and mitigate their drawbacks.

\section{Experiments and discussion}\label{sec:wtsneexperiments}

Six datasets were used for the experiments with weighted t-SNE and feature scorers.
They are fully described in Table~\ref{tab:tsnedata}.
Two datasets are synthetic classification tasks, so the number of relevant and irrelevant features can be controlled.
The first dataset used in the study was modeled after the exclusive-OR (XOR), using two out of a total of $n$ inputs \cite{tan2009automated, grisci2021relevance}.
The XOR function in this problem is determined by two specific binary features in the input data, which are known to the user but not to the machine learning models.
The other $n - 2$ features, where $n = 50$, are randomly generated binary values with no effect on the output.
Samples are assigned either to class 0 or class 1 according to the following values of their two relevant features: $0 \oplus 0 = 0$, $1 \oplus 1 = 0$, $0 \oplus 1 = 1$, and $1 \oplus 0 = 1$.
The other 48 features are zero or one and do not impact the sample's class.
Despite its apparent simplicity, this problem is made more difficult by its non-linearity and the presence of many irrelevant features.

Another synthetic dataset with two classes and 100 samples created with \emph{scikit-learn} \cite{guyon2003design, scikit-learn}\footnote{\url{https://scikit-learn.org/stable/modules/generated/sklearn.datasets.make_classification.html}.} was also used.
This dataset has 5000 input features, with only $50$ being informative for class separation (unknown to the algorithms), and the rest being irrelevant.
Two other datasets are genes expression from cancer microarray experiments due to their importance, as discussed in Section~\ref{subsec:featurescoring}.
They are from liver and prostate cancer microarray experiments obtained from the CuMiDa database \cite{feltes2019cumida}\footnote{\url{https://sbcb.inf.ufrgs.br/cumida}.} under the accession codes GSE22405 and GSE6919\_U95B.
These datasets are composed of two classes, one with samples from healthy tissue and the other from tumorous tissue.
The choice of gene expression datasets followed the recommendations of \citet{grisci2023use} regarding the data quality and up-to-dateness.

Two additional datasets are used to illustrate the points made in Section~\ref{subsec:machinelearning}.
The first is a synthetic regression dataset (Fig.~\ref{fig:teaser}) with 100 features and 1000 samples with continuous target values ranging between $-12$ and $12$ created with \emph{scikit-learn} \cite{celeux2012regularization, scikit-learn}.\footnote{\url{https://scikit-learn.org/stable/modules/generated/sklearn.datasets.make_sparse_uncorrelated.html}.}
It was only tested with the neural network because some of the other algorithms are not directly suited for regression.
The second is an RNA-seq dataset of $23,822$ cells from adult mouse cortex \cite{tasic2018shared}, divided by the hierarchy of cell types (23 classes).
This larger dataset was used for the machine learning experiments.
Except for the XOR dataset, all the data was standardized.
It is a best practice only to apply feature selection or scoring in balanced datasets.
All datasets used in our experiments have balanced classes to avoid issues.
There is no general way to describe the effects of class imbalance in all feature scorers because each algorithm has distinct strategies, advantages, and drawbacks.
However, the work of \citet{kamalov2022feature} further discusses the problem.

\begin{table}[tbp]
\centering
\caption{Description of the datasets used in the weighted t-SNE experiments. Relevant is the number of features which are relevant to the task. All datasets are numerical and tabular.}
\label{tab:tsnedata}
\begin{tabular}{lllllll}
\hline
\textbf{Dataset} & \textbf{Samples} & \textbf{Features} & \textbf{Relevant} & \textbf{Classes} & \textbf{Origin} & \textbf{Reference} \\ \hline
XOR          & 500        & 50         & 2       & 2  & Synthetic  & \cite{tan2009automated, grisci2021relevance} \\
Synth        & 100        & 5000       & 50      & 2  & Synthetic  & \cite{guyon2003design, scikit-learn} \\
Liver        & 48         & 22\,284    & Unknown & 2  & Microarray & \cite{feltes2019cumida} \\
Prostate     & 115        & 12\,647    & Unknown & 2  & Microarray & \cite{feltes2019cumida} \\
Regression   & 1000       & 100        & 4       & -- & Synthetic  & \cite{celeux2012regularization, scikit-learn} \\
Mouse cortex & 23\,822    & 45\,769    & Unknown & 23 & RNA-seq    & \cite{tasic2018shared, kobak2019art} \\ \hline
\end{tabular}
\end{table}

The weighted t-SNE was implemented using Python 3, the \emph{openTSNE} library,\footnote{\url{https://opentsne.readthedocs.io/en/latest/}.} and \emph{NumPy} \cite{harris2020array}.
The perplexity was computed as described in Section~\ref{subsec:tsne}, and the positions were initialized using PCA, following the recommendations from \citet{kobak2019art}.
The appropriate learning rate is selected according to $max(200, n / 12)$ \cite{belkina2019automated}.
All experiments were conducted with $500$ iterations.
The silhouette coefficient was computed using the implementation from \emph{scikit-learn} \cite{scikit-learn}.\footnote{\url{https://scikit-learn.org/stable/modules/generated/sklearn.metrics.silhouette_score.html}.}

Nine feature scorers were chosen to generate the visualizations: Kruskal--Wallis Filter, Mutual Information, mRMR, ReliefF, Lasso, Decision Tree, Random Forest, Linear SVM, and Neural Network.
These algorithms represent the several scoring strategies discussed in Section~\ref{subsec:featurescoring}, as summarized in Table~\ref{tab:fsprop}.
All their outputs are feature importance weights between zero (irrelevant) and one (relevant), so the method presented in Section~\ref{sec:wtsne} can be applied.
The machine learning-based methods are further described and discussed in Section~\ref{subsec:machinelearning}.
The neural network was trained using \emph{Keras} with the \emph{TensorFlow} backend.
The experiment with the XOR dataset had two hidden layers of 20 neurons each, ReLU activation and L1 regularization with a factor of $0.001$.
The network trained for the regression task had two hidden layers with 128 and 64 neurons, and the network for the mouse cortex dataset had four hidden layers with 100, 200, 100, and 100 neurons.
For the other experiments, the neural network had four hidden layers with 32 neurons each and L1 regularization with a factor of $0.01$.
Because the goal of the experiments is only to show the functionality of weighted t-SNE, the other scorers were trained using the recommended hyperparameters of their implementations from \emph{scikit-learn} \cite{scikit-learn}.
Further details about each feature scorer and how feature importance is computed are present in the \hyperref[app:supp]{Supplementary material}.
Weighted t-SNE can even be used to compare different sets of hyperparameters for the same feature scorer.
In addition to the scorers, the results for ``no scoring'' are also shown, corresponding to a standard t-SNE visualization of the original high-dimensional data.

\begin{table}[tbp]
\centering
\caption{Properties of the feature scorers used in the experiments. The type refers to how the scorer works and the correlation to how interactions between features are treated.}
\label{tab:fsprop}
\begin{tabular}{llll}
\hline
\textbf{Scorer}    & \textbf{Type} & \textbf{Correlation} & \textbf{Reference} \\ \hline
Kruskal--Wallis    & Filter        & Univariate           & \cite{kruskal1952use} \\
Mutual Information & Filter        & Univariate           & \cite{vergara2014review} \\
mRMR               & Filter        & Multivariate         & \cite{peng2005feature} \\
ReliefF            & Filter        & Multivariate         & \cite{kononenko1997overcoming} \\
Lasso              & Embedded      & Multivariate         & \cite{fonti2017feature} \\
Decision Tree      & Embedded      & Multivariate         & \cite{stone1984classification} \\
Random Forest      & Ensemble      & Multivariate         & \cite{breiman2001random} \\
Linear SVM         & Embedded      & Multivariate         & \cite{cortes1995support} \\
Neural Network     & Embedded      & Multivariate         & \cite{grisci2021relevance} \\ \hline
\end{tabular}
\end{table}

\begin{figure}[tbp]
\centering
\begin{subfigure}[t]{.23\textwidth}
  \centering
  \fbox{\includegraphics[width=.9\linewidth, height=.9\linewidth, alt={t-SNE scatterplot of the XOR dataset without feature scoring, with samples colored by class}]{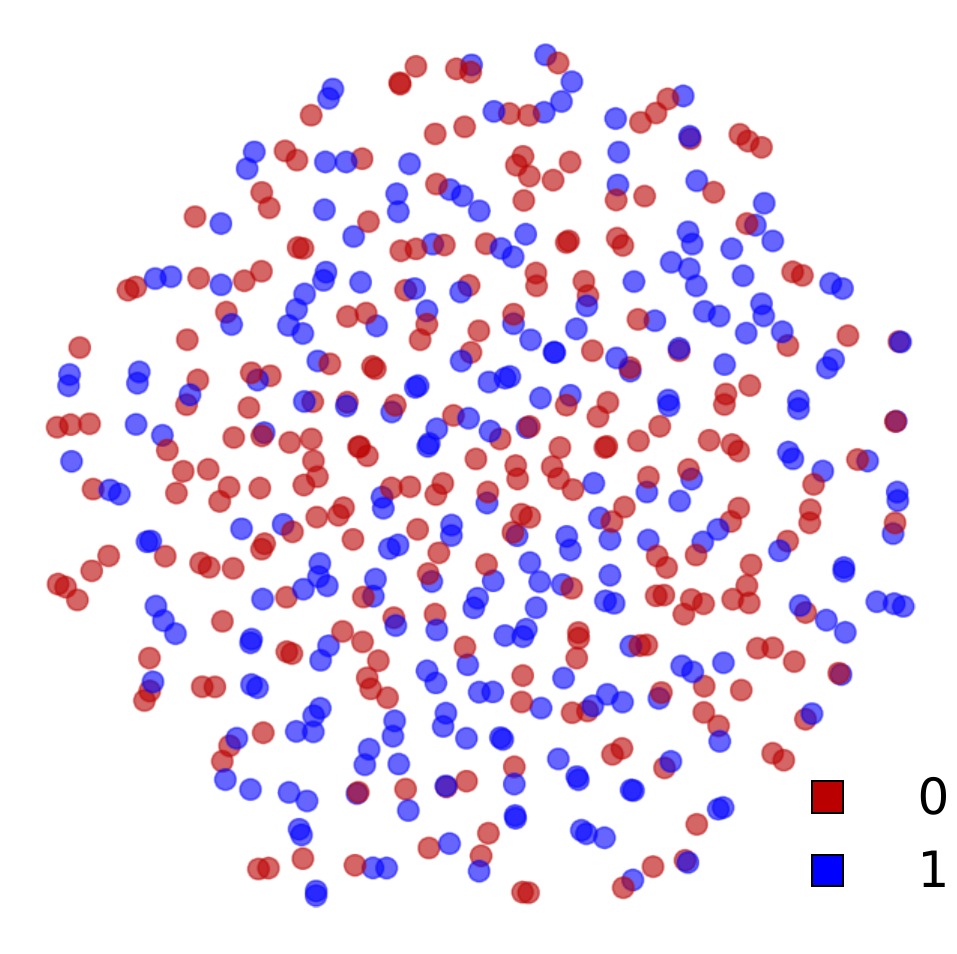}}
  \caption{No scoring}
  \label{fig:wtsnexornoselection}
\end{subfigure}
\begin{subfigure}[t]{.23\textwidth}
  \centering
  \fbox{\includegraphics[width=.9\linewidth, height=.9\linewidth, alt={Weighted t-SNE scatterplot of the XOR dataset using the Kruskal-Wallis feature scores, with samples colored by class}]{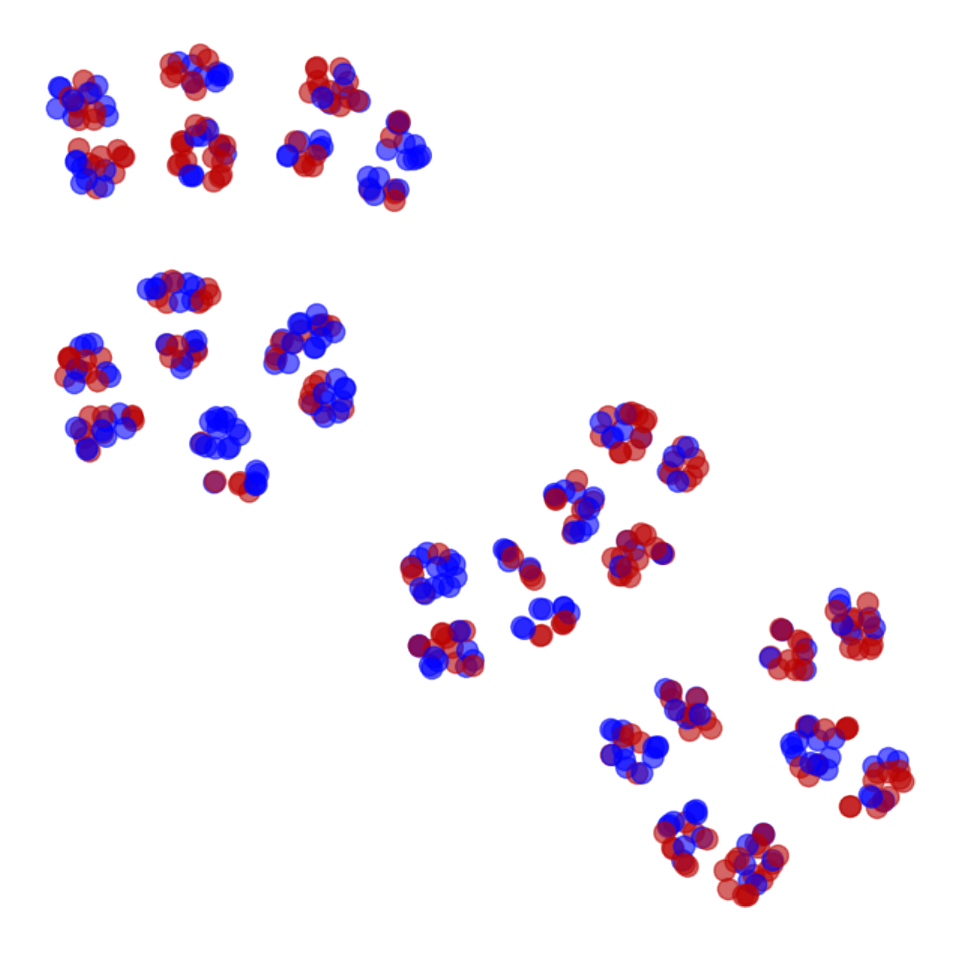}}
  \caption{Kruskal--Wallis}
  \label{fig:wtsnexorkw}
\end{subfigure}
\begin{subfigure}[t]{.23\textwidth}
  \centering
  \fbox{\includegraphics[width=.9\linewidth, height=.9\linewidth, alt={Weighted t-SNE scatterplot of the XOR dataset using the Mutual Information feature scores, with samples colored by class}]{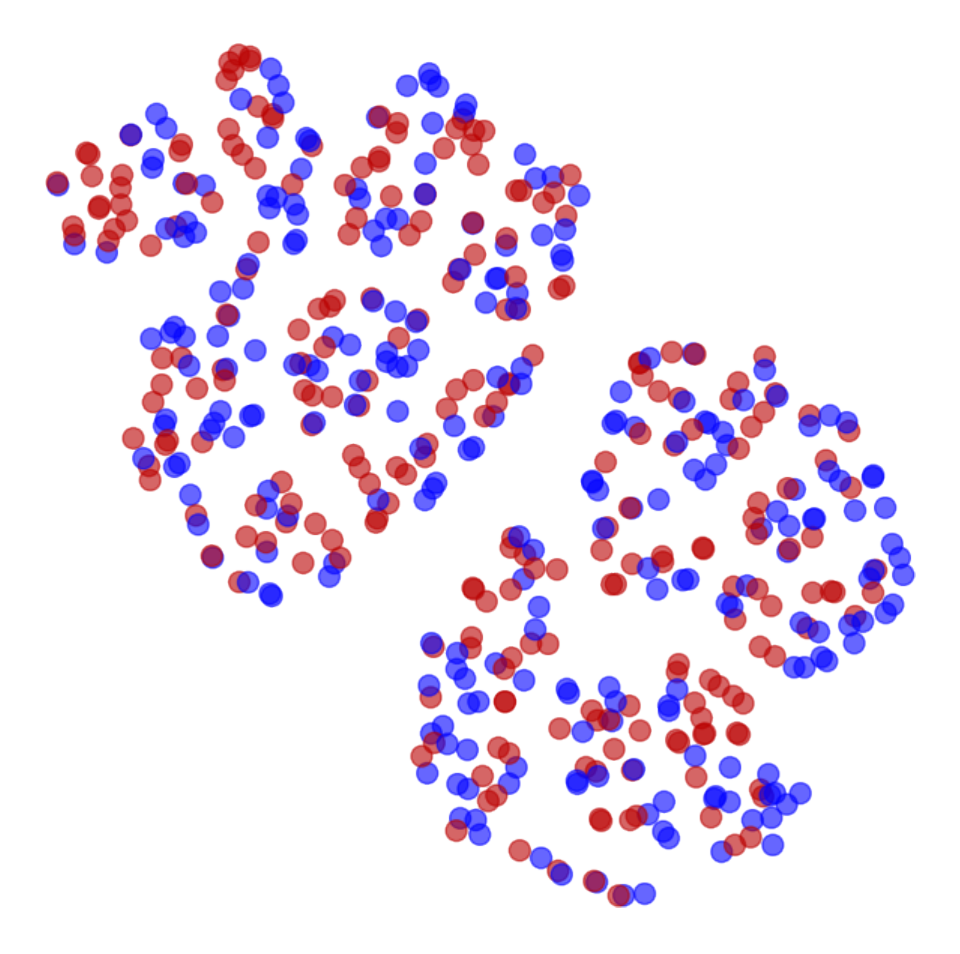}}
  \caption{Mutual Information}
  \label{fig:wtsnexormi}
\end{subfigure}
\begin{subfigure}[t]{.23\textwidth}
  \centering
  \fbox{\includegraphics[width=.9\linewidth, height=.9\linewidth, alt={Weighted t-SNE scatterplot of the XOR dataset using the mRMR feature scores, with samples colored by class}]{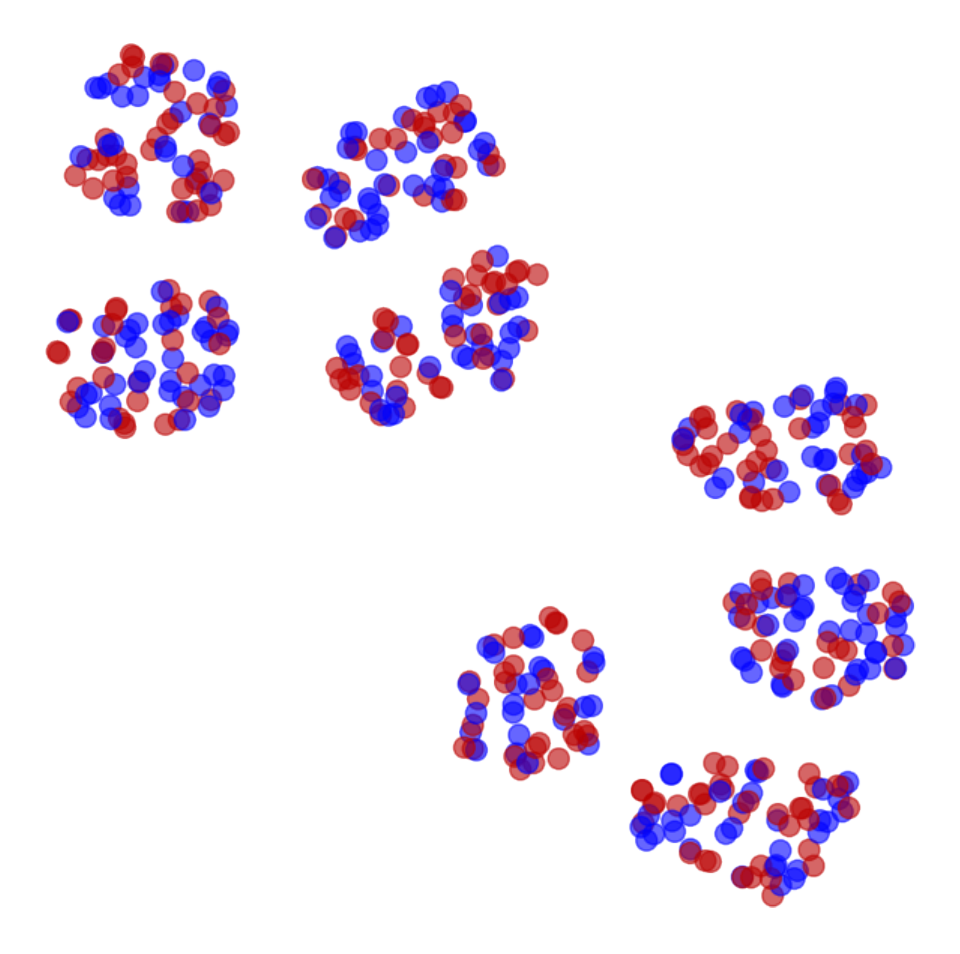}}
  \caption{mRMR}
  \label{fig:wtsnexormrmr}
\end{subfigure}

\begin{subfigure}[t]{.23\textwidth}
  \centering
  \fbox{\includegraphics[width=.9\linewidth, height=.9\linewidth, alt={Weighted t-SNE scatterplot of the XOR dataset using the ReliefF feature scores, with samples colored by class}]{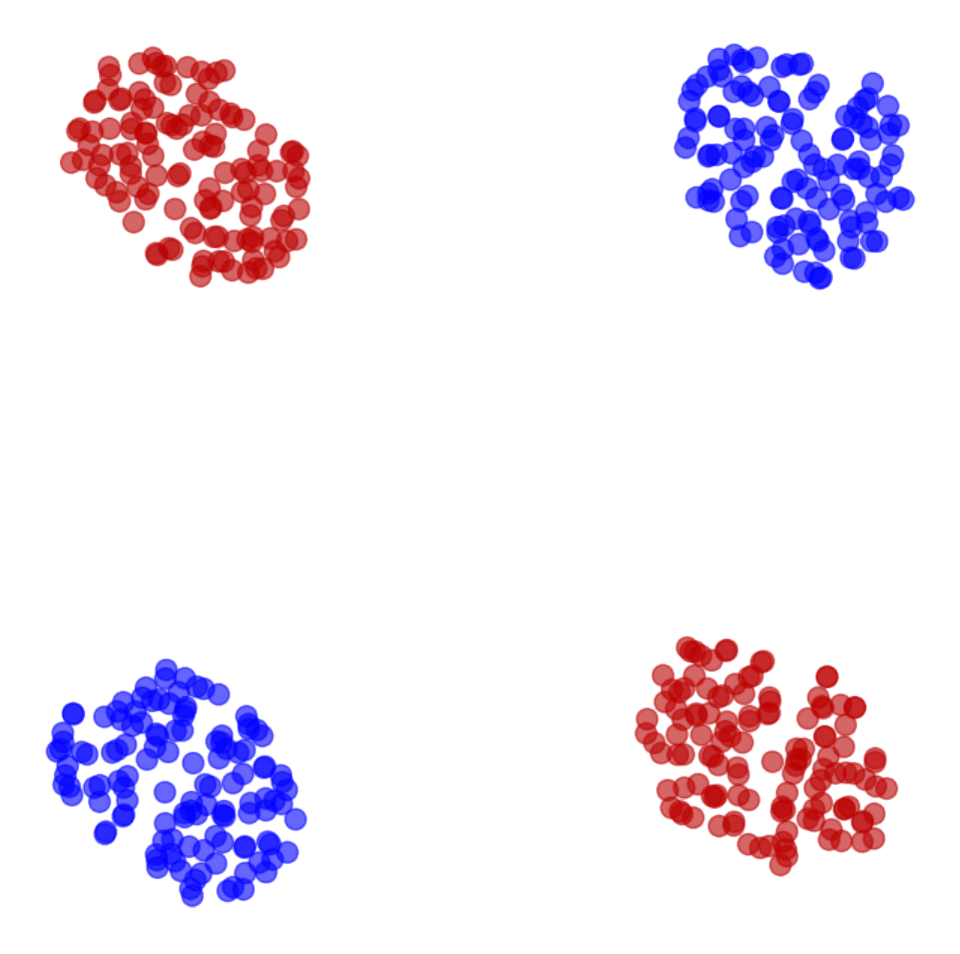}}
  \caption{ReliefF}
  \label{fig:wtsnexorrelieff}
\end{subfigure}
\begin{subfigure}[t]{.23\textwidth}
  \centering
  \fbox{\includegraphics[width=.9\linewidth, height=.9\linewidth, alt={Weighted t-SNE scatterplot of the XOR dataset using the Lasso feature scores, with samples colored by class}]{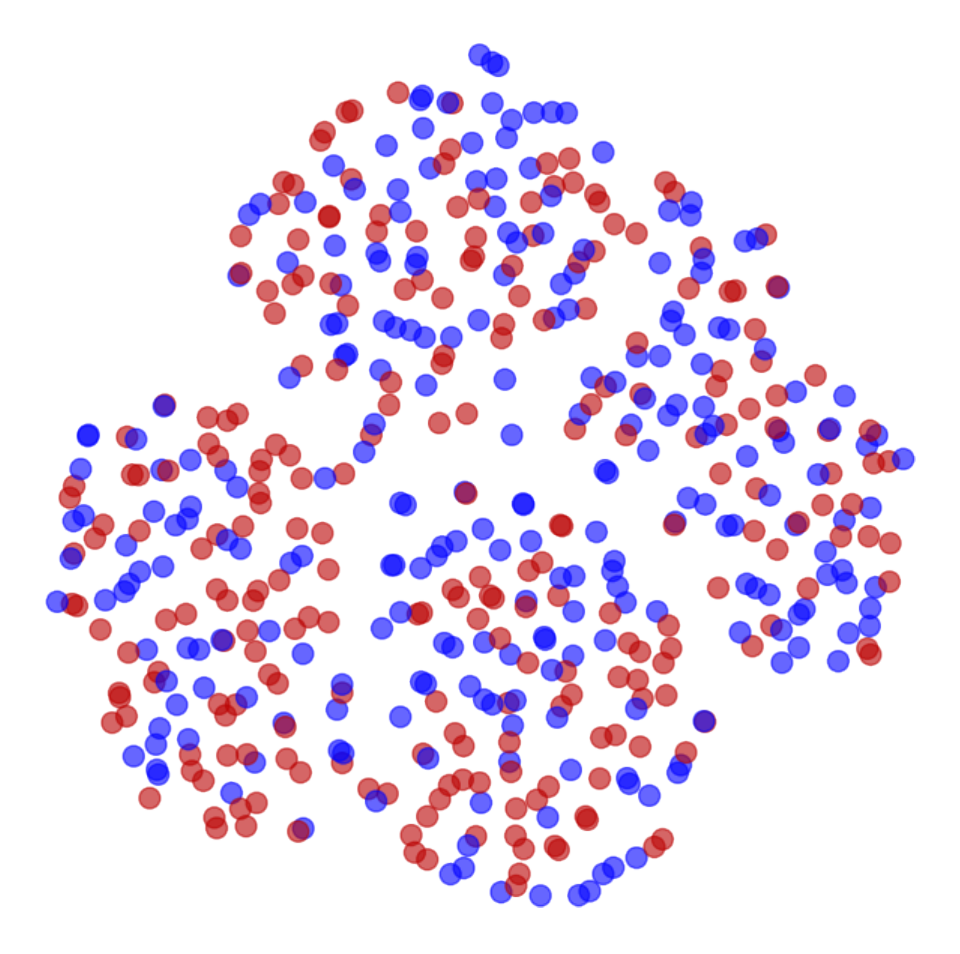}}
  \caption{Lasso}
  \label{fig:wtsnexorlasso}
\end{subfigure}
\begin{subfigure}[t]{.23\textwidth}
  \centering
  \fbox{\includegraphics[width=.9\linewidth, height=.9\linewidth, alt={Weighted t-SNE scatterplot of the XOR dataset using the Decision Tree feature scores, with samples colored by class}]{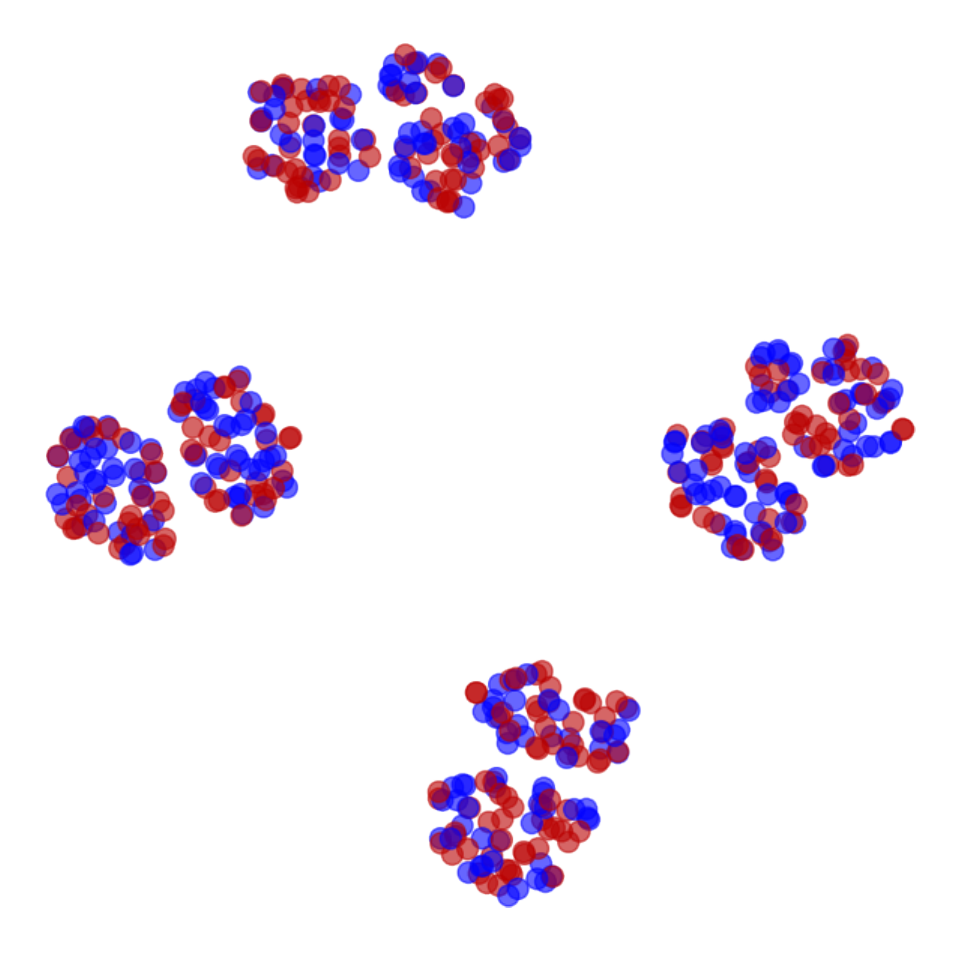}}
  \caption{Decision Tree}
  \label{fig:wtsnexordt}
\end{subfigure}
\begin{subfigure}[t]{.23\textwidth}
  \centering
  \fbox{\includegraphics[width=.9\linewidth, height=.9\linewidth, alt={Weighted t-SNE scatterplot of the XOR dataset using the Random Forest feature scores, with samples colored by class}]{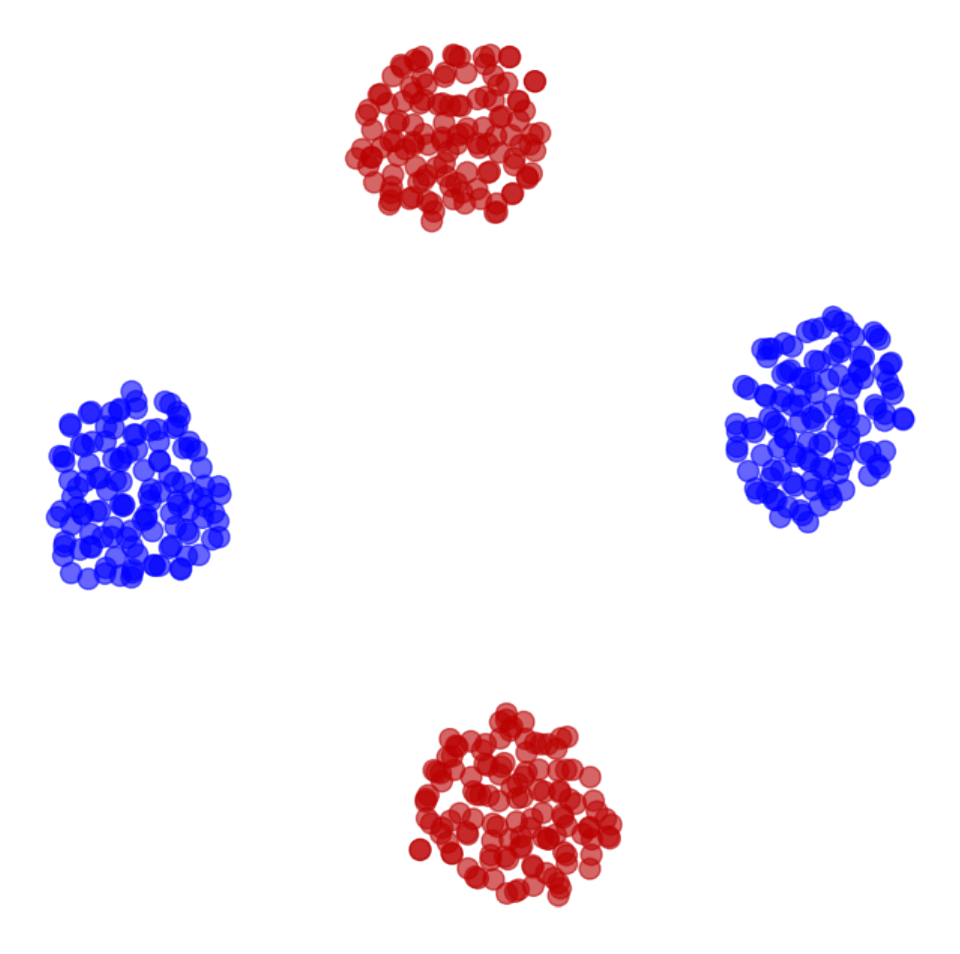}}
  \caption{Random Forest}
  \label{fig:wtsnexorrf}
\end{subfigure}

\begin{subfigure}[t]{.23\textwidth}
  \centering
  \fbox{\includegraphics[width=.9\linewidth, height=.9\linewidth, alt={Weighted t-SNE scatterplot of the XOR dataset using the Linear SVM feature scores, with samples colored by class}]{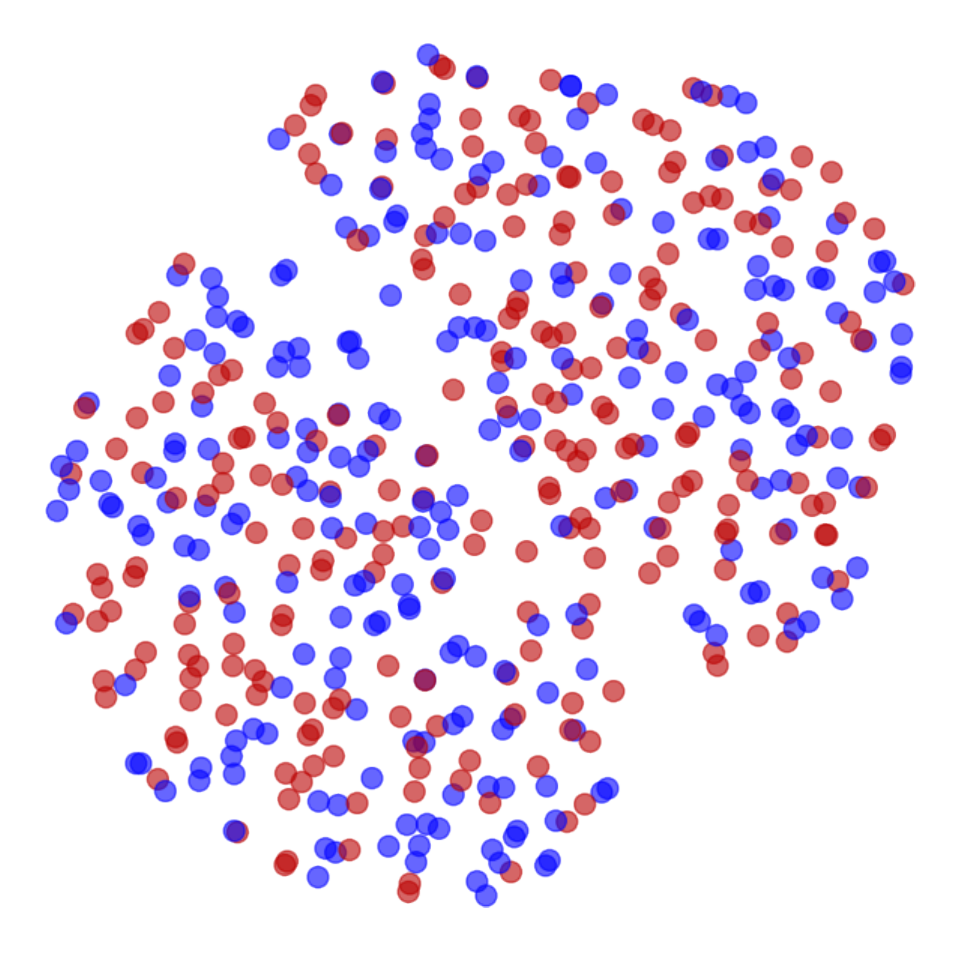}}
  \caption{Linear SVM}
  \label{fig:wtsnexorlinearsvm}
\end{subfigure}
\begin{subfigure}[t]{.23\textwidth}
  \centering
  \fbox{\includegraphics[width=.9\linewidth, height=.9\linewidth, alt={Weighted t-SNE scatterplot of the XOR dataset using the neural network (relevance aggregation) feature scores, with samples colored by class}]{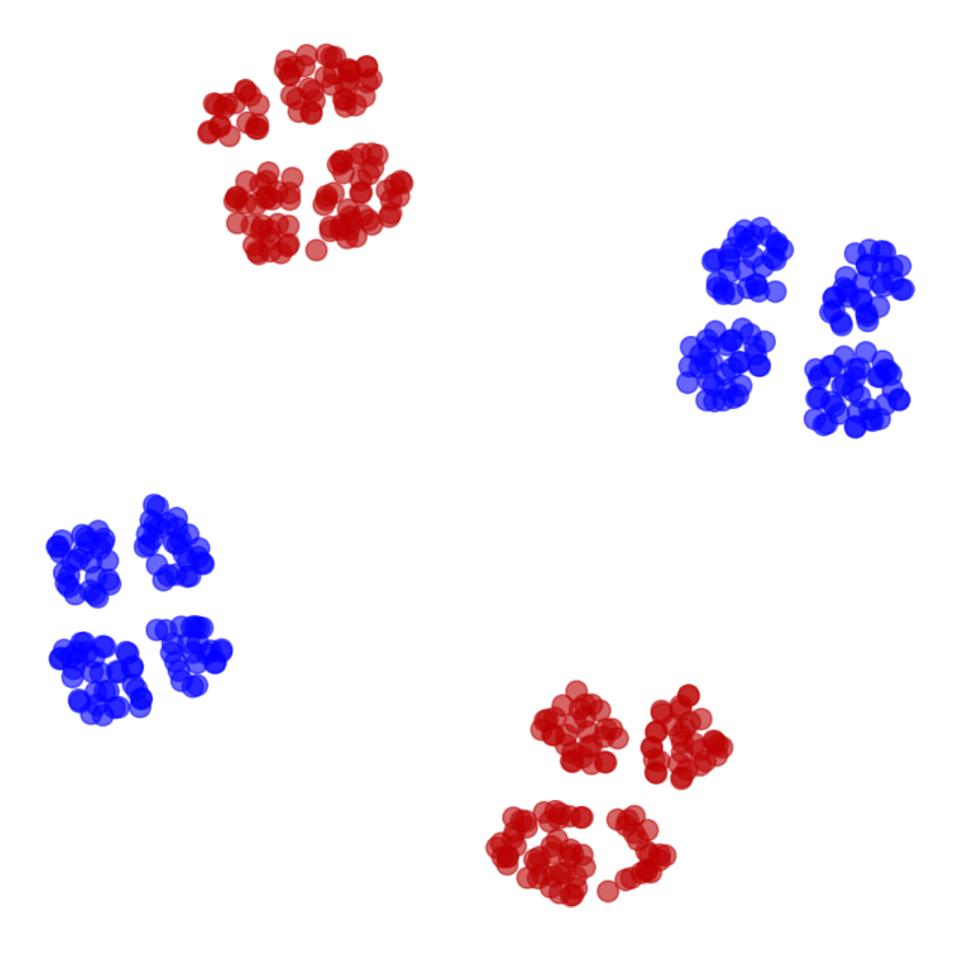}}
  \caption{Neural Network}
  \label{fig:wtsnexorrelagg}
\end{subfigure}
\caption{Weighted t-SNE visualization for the XOR dataset. Each figure is the visualization of the 2D embedding with weighted t-SNE for each of the nine feature scorers, plus the original data in ``no scoring''. Each red or blue point represents one sample from one of the two classes. The color legend is shown in Fig.~\ref{fig:wtsnexornoselection} and omitted in the other figures.}
\label{fig:wtsnexor}
\end{figure}

\begin{figure}[tbp]
\centering
\begin{subfigure}[t]{.23\textwidth}
  \centering
  \fbox{\includegraphics[width=.9\linewidth, height=.9\linewidth, alt={t-SNE scatterplot of the synthetic Synth dataset without feature scoring, with samples colored by class}]{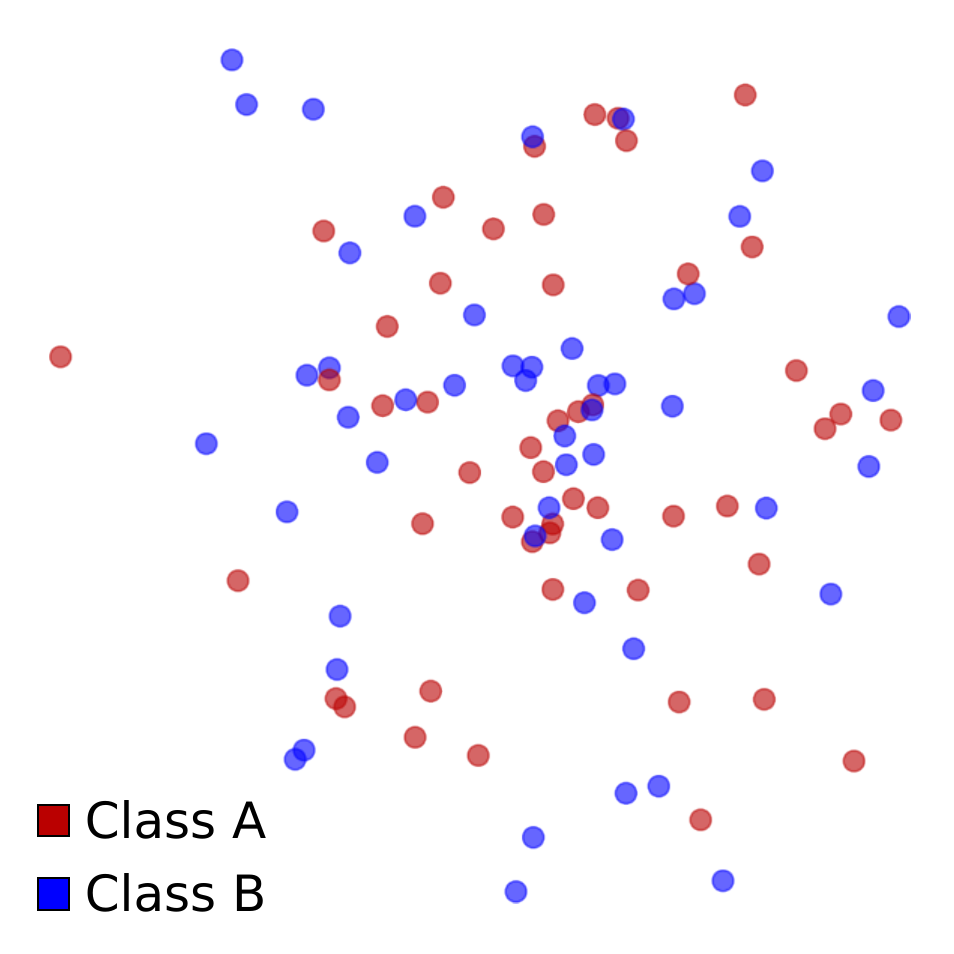}}
  \caption{No scoring}
  \label{fig:wtsnesynthnoselection}
\end{subfigure}
\begin{subfigure}[t]{.23\textwidth}
  \centering
  \fbox{\includegraphics[width=.9\linewidth, height=.9\linewidth, alt={Weighted t-SNE scatterplot of the synthetic Synth dataset using the Kruskal-Wallis feature scores, with samples colored by class}]{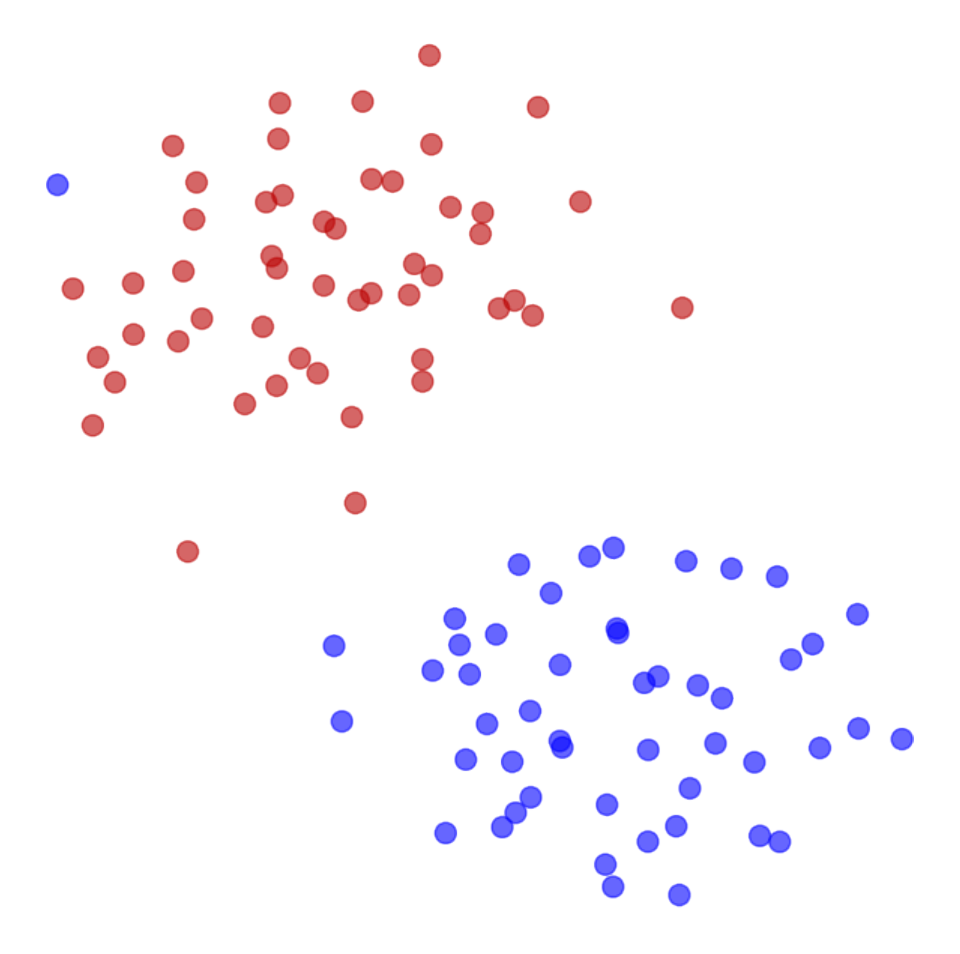}}
  \caption{Kruskal--Wallis}
  \label{fig:wtsnesynthkw}
\end{subfigure}
\begin{subfigure}[t]{.23\textwidth}
  \centering
  \fbox{\includegraphics[width=.9\linewidth, height=.9\linewidth, alt={Weighted t-SNE scatterplot of the synthetic Synth dataset using the Mutual Information feature scores, with samples colored by class}]{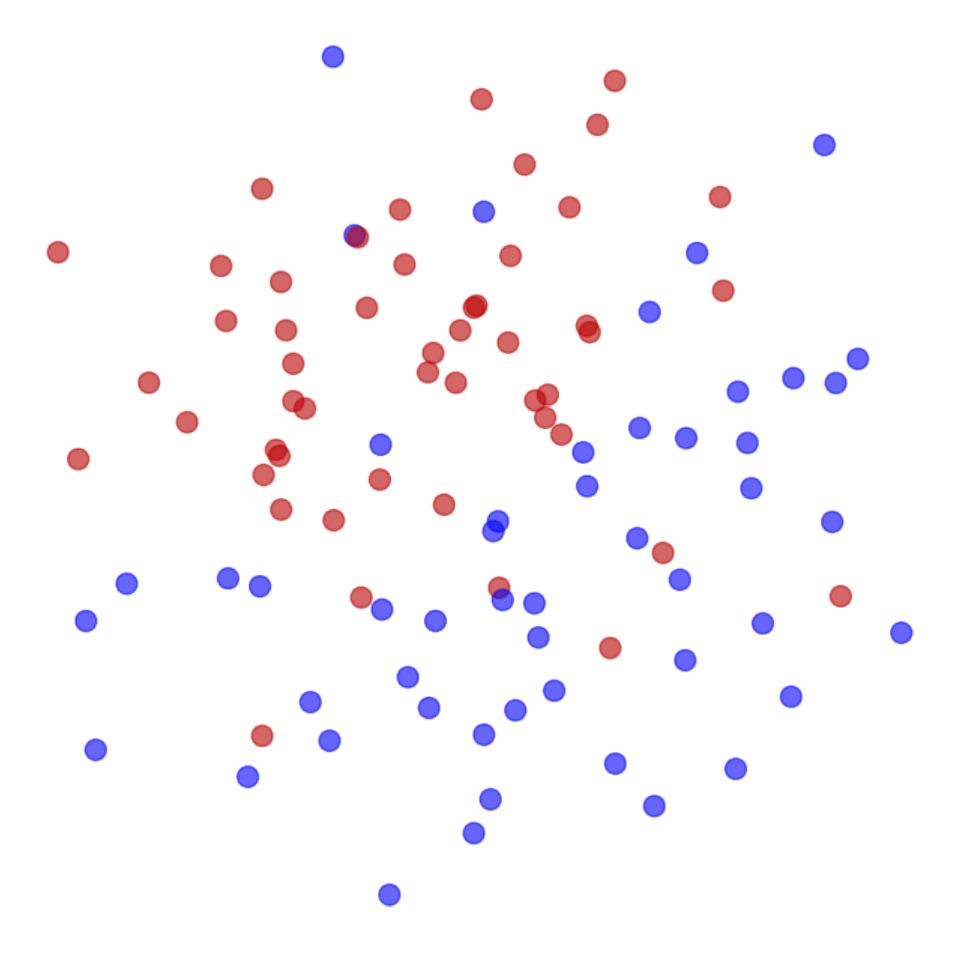}}
  \caption{Mutual Information}
  \label{fig:wtsnesynthmi}
\end{subfigure}
\begin{subfigure}[t]{.23\textwidth}
  \centering
  \fbox{\includegraphics[width=.9\linewidth, height=.9\linewidth, alt={Weighted t-SNE scatterplot of the synthetic Synth dataset using the mRMR feature scores, with samples colored by class}]{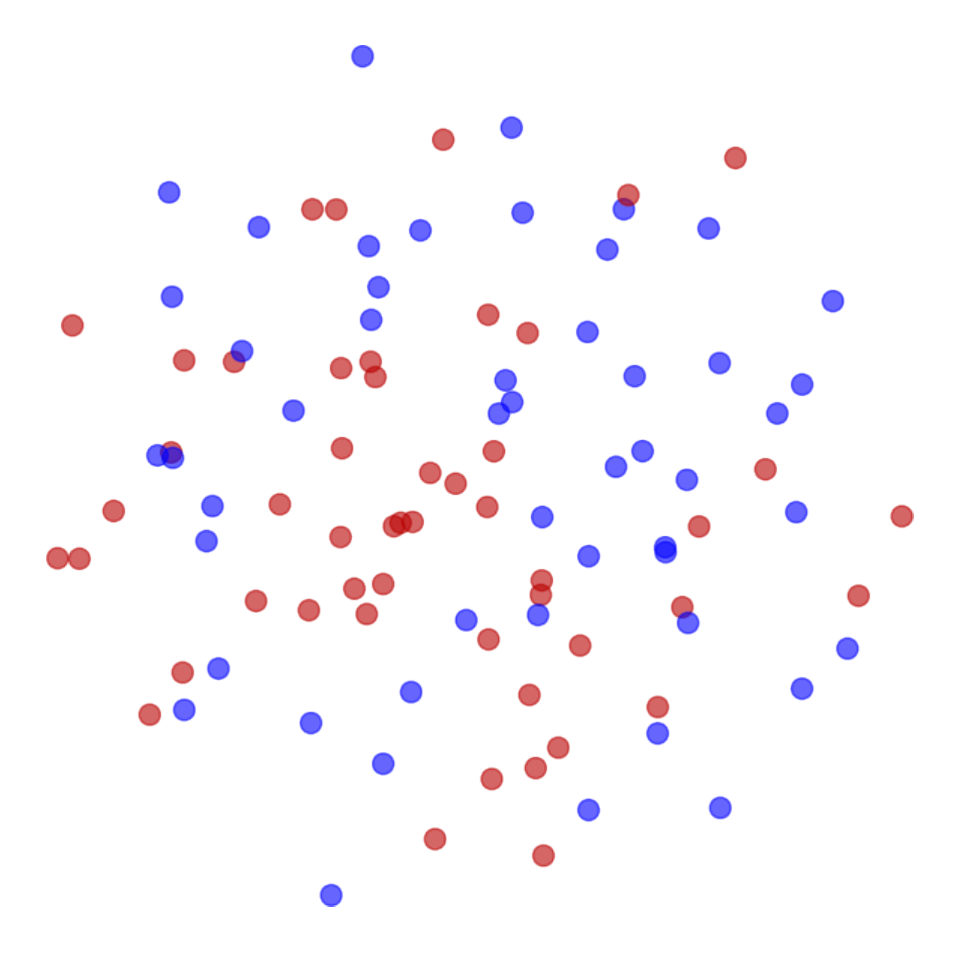}}
  \caption{mRMR}
  \label{fig:wtsnesynthmrmr}
\end{subfigure}

\begin{subfigure}[t]{.23\textwidth}
  \centering
  \fbox{\includegraphics[width=.9\linewidth, height=.9\linewidth, alt={Weighted t-SNE scatterplot of the synthetic Synth dataset using the ReliefF feature scores, with samples colored by class}]{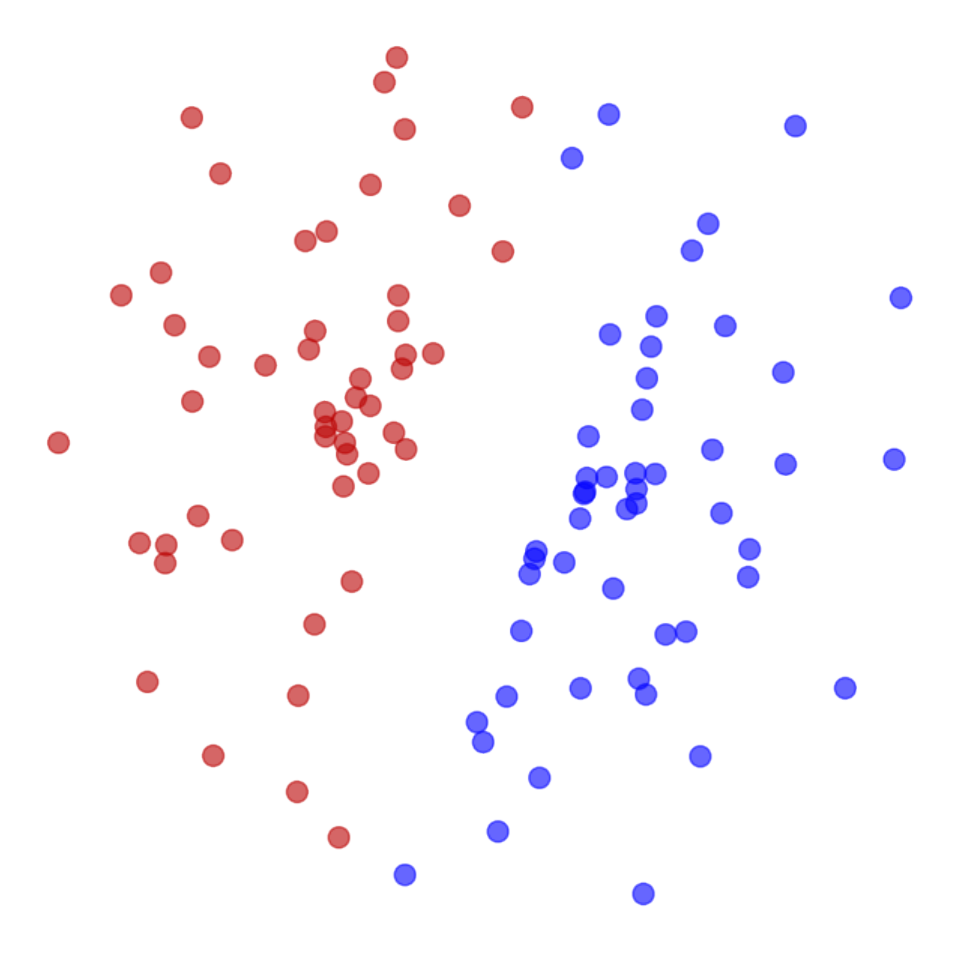}}
  \caption{ReliefF}
  \label{fig:wtsnesynthrelieff}
\end{subfigure}
\begin{subfigure}[t]{.23\textwidth}
  \centering
  \fbox{\includegraphics[width=.9\linewidth, height=.9\linewidth, alt={Weighted t-SNE scatterplot of the synthetic Synth dataset using the Lasso feature scores, with samples colored by class}]{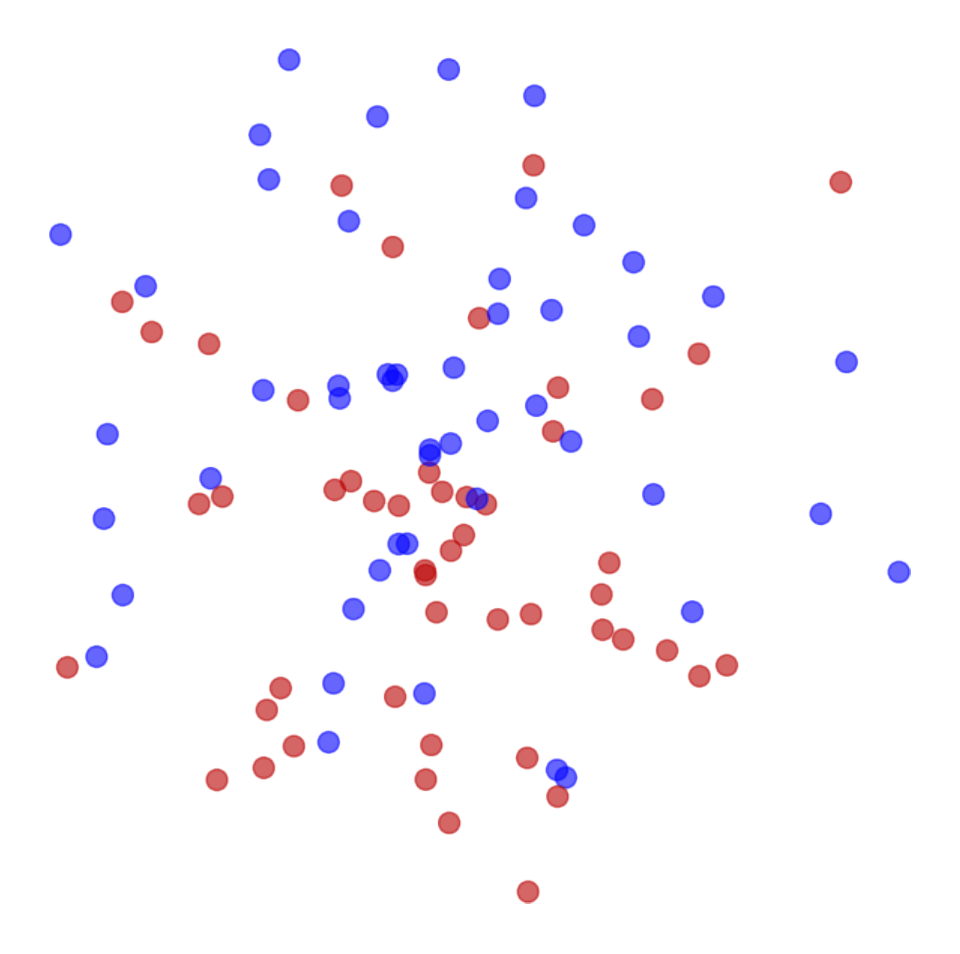}}
  \caption{Lasso}
  \label{fig:wtsnesynthlasso}
\end{subfigure}
\begin{subfigure}[t]{.23\textwidth}
  \centering
  \fbox{\includegraphics[width=.9\linewidth, height=.9\linewidth, alt={Weighted t-SNE scatterplot of the synthetic Synth dataset using the Decision Tree feature scores, with samples colored by class}]{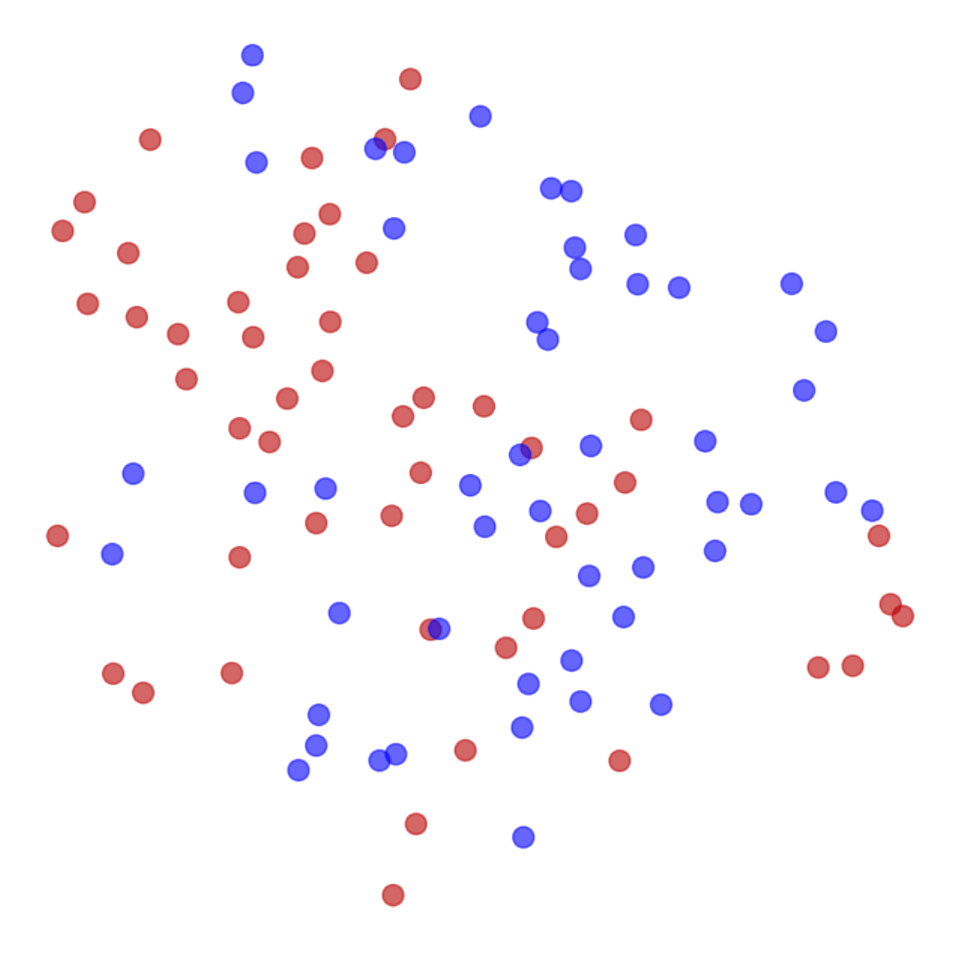}}
  \caption{Decision Tree}
  \label{fig:wtsnesynthdt}
\end{subfigure}
\begin{subfigure}[t]{.23\textwidth}
  \centering
  \fbox{\includegraphics[width=.9\linewidth, height=.9\linewidth, alt={Weighted t-SNE scatterplot of the synthetic Synth dataset using the Random Forest feature scores, with samples colored by class}]{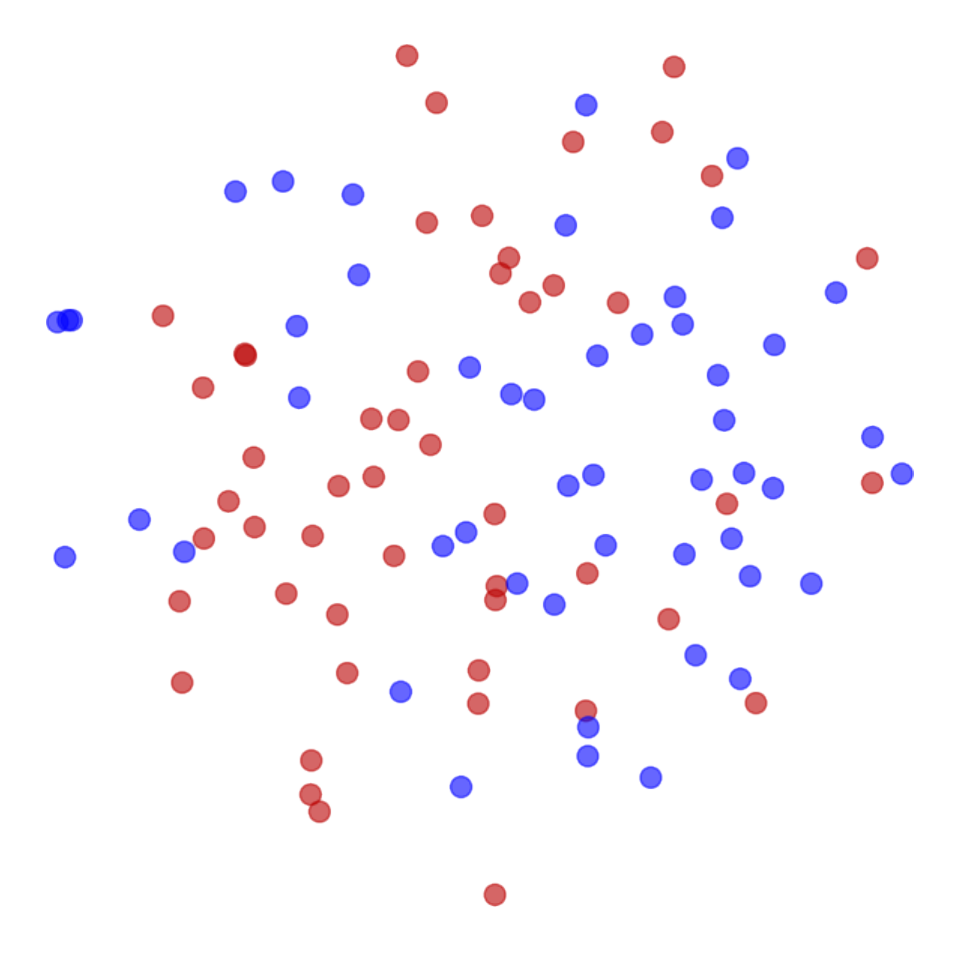}}
  \caption{Random Forest}
  \label{fig:wtsnesynthrf}
\end{subfigure}

\begin{subfigure}[t]{.23\textwidth}
  \centering
  \fbox{\includegraphics[width=.9\linewidth, height=.9\linewidth, alt={Weighted t-SNE scatterplot of the synthetic Synth dataset using the Linear SVM feature scores, with samples colored by class}]{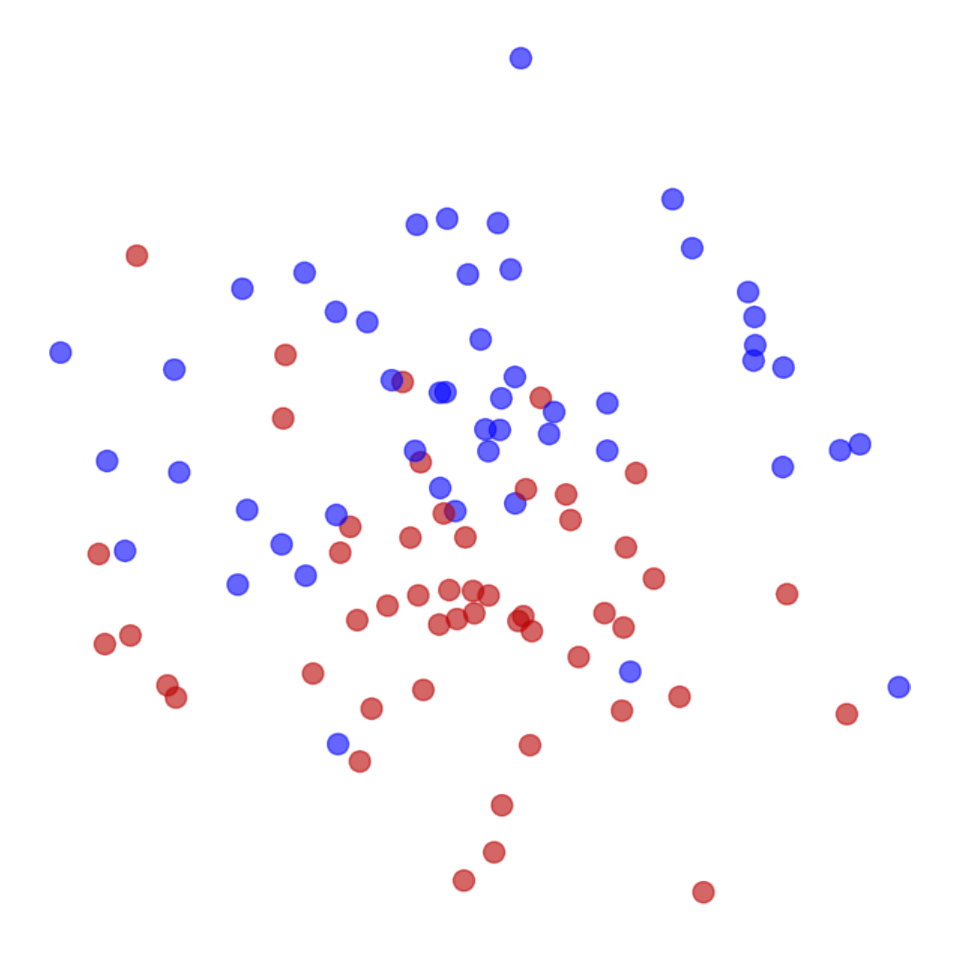}}
  \caption{Linear SVM}
  \label{fig:wtsnesynthlinearsvm}
\end{subfigure}
\begin{subfigure}[t]{.23\textwidth}
  \centering
  \fbox{\includegraphics[width=.9\linewidth, height=.9\linewidth, alt={Weighted t-SNE scatterplot of the synthetic Synth dataset using the neural network (relevance aggregation) feature scores, with samples colored by class}]{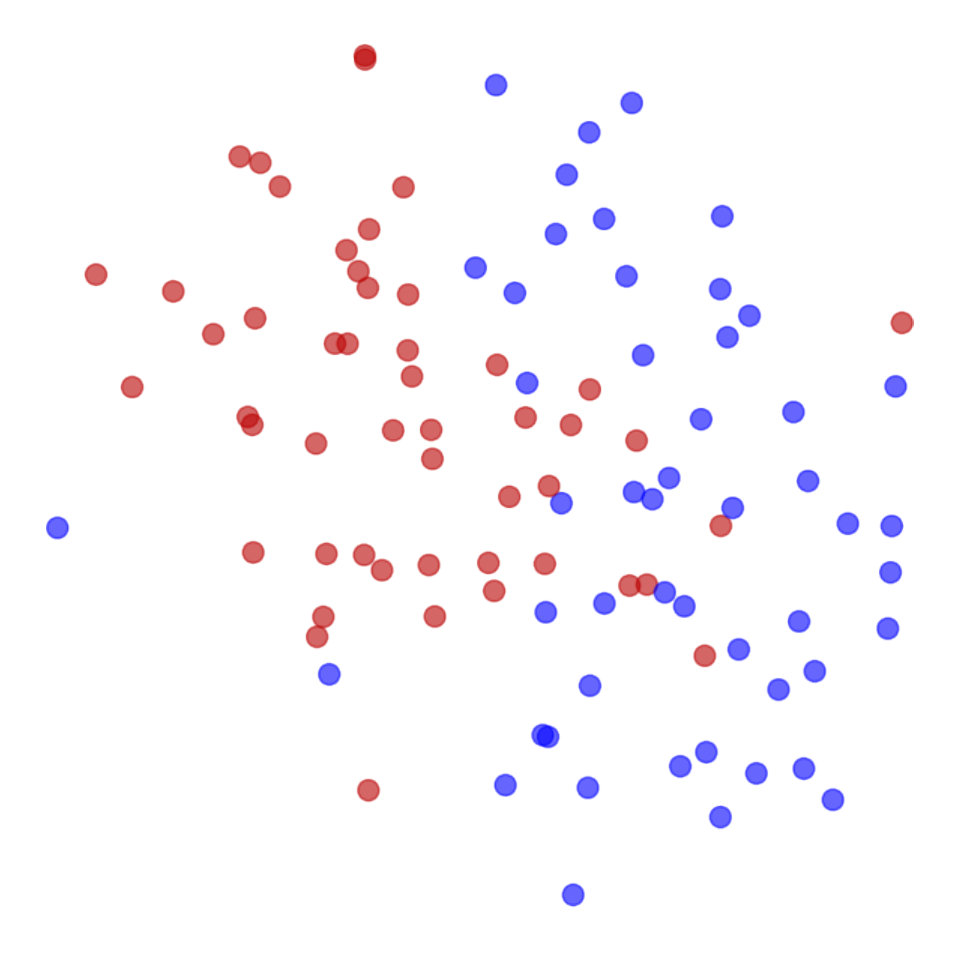}}
  \caption{Neural Network}
  \label{fig:wtsnesynthrelagg}
\end{subfigure}
\caption{Weighted t-SNE visualization for the synthetic 2-classes dataset. Each red or blue point represents one sample from one of the two classes. The color legend is shown in Fig.~\ref{fig:wtsnesynthnoselection} and omitted in the other figures.}
\label{fig:wtsnesynth}
\end{figure}

\begin{figure}[tbp]
\centering
\begin{subfigure}[t]{.23\textwidth}
  \centering
  \fbox{\includegraphics[width=.9\linewidth, height=.9\linewidth, alt={t-SNE scatterplot of the liver cancer dataset without feature scoring, with samples colored by class}]{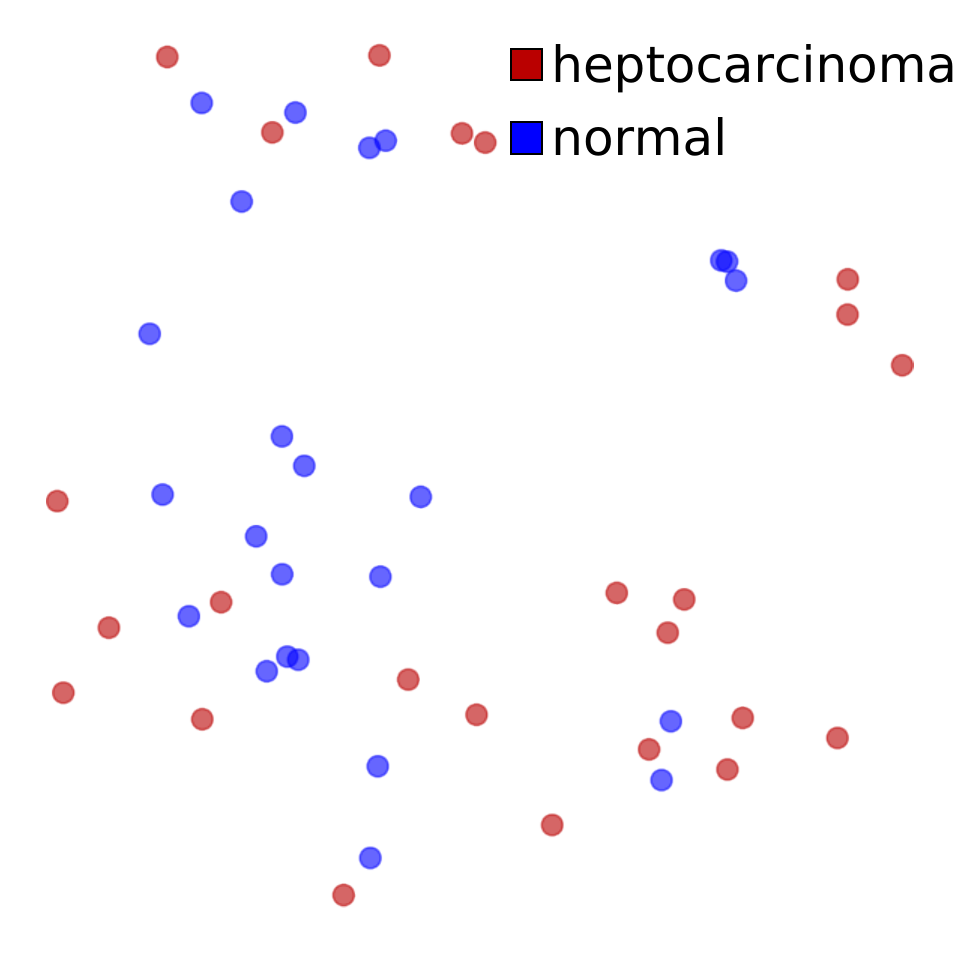}}
  \caption{No scoring}
  \label{fig:wtsnelivernoselection}
\end{subfigure}
\begin{subfigure}[t]{.23\textwidth}
  \centering
  \fbox{\includegraphics[width=.9\linewidth, height=.9\linewidth, alt={Weighted t-SNE scatterplot of the liver cancer dataset using the Kruskal-Wallis feature scores, with samples colored by class}]{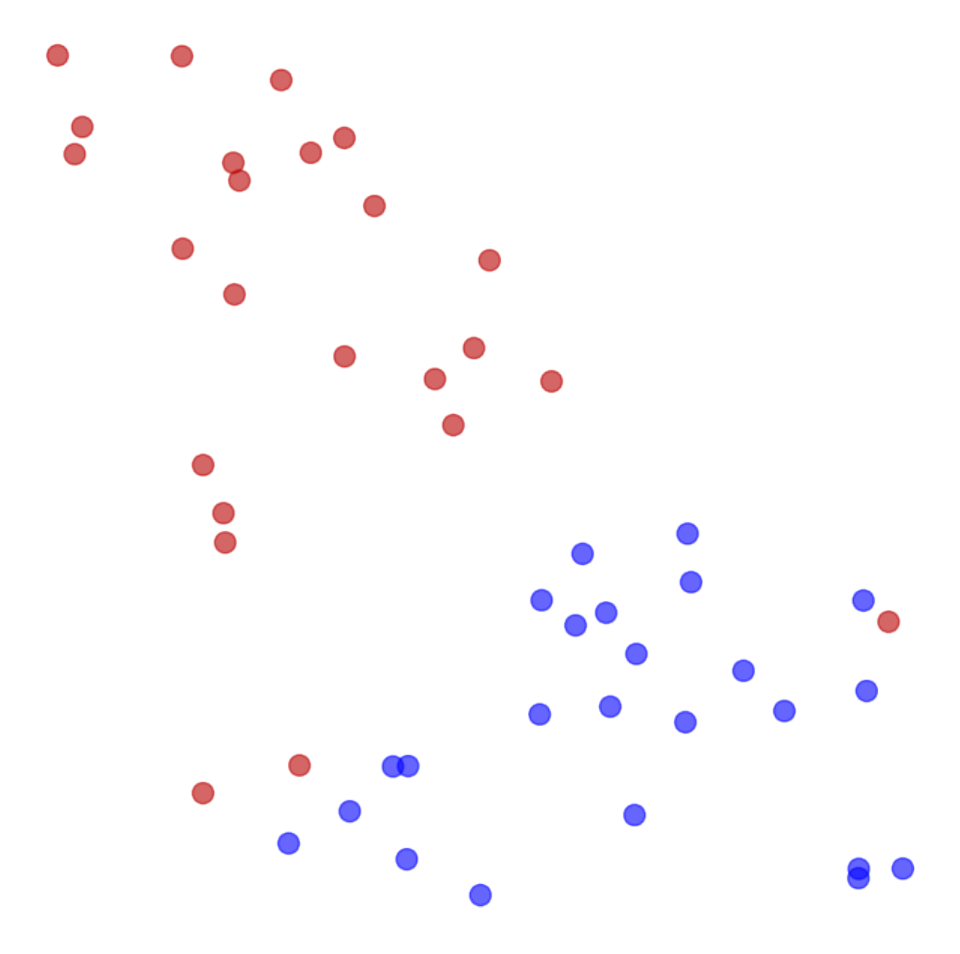}}
  \caption{Kruskal--Wallis}
  \label{fig:wtsneliverkw}
\end{subfigure}
\begin{subfigure}[t]{.23\textwidth}
  \centering
  \fbox{\includegraphics[width=.9\linewidth, height=.9\linewidth, alt={Weighted t-SNE scatterplot of the liver cancer dataset using the Mutual Information feature scores, with samples colored by class}]{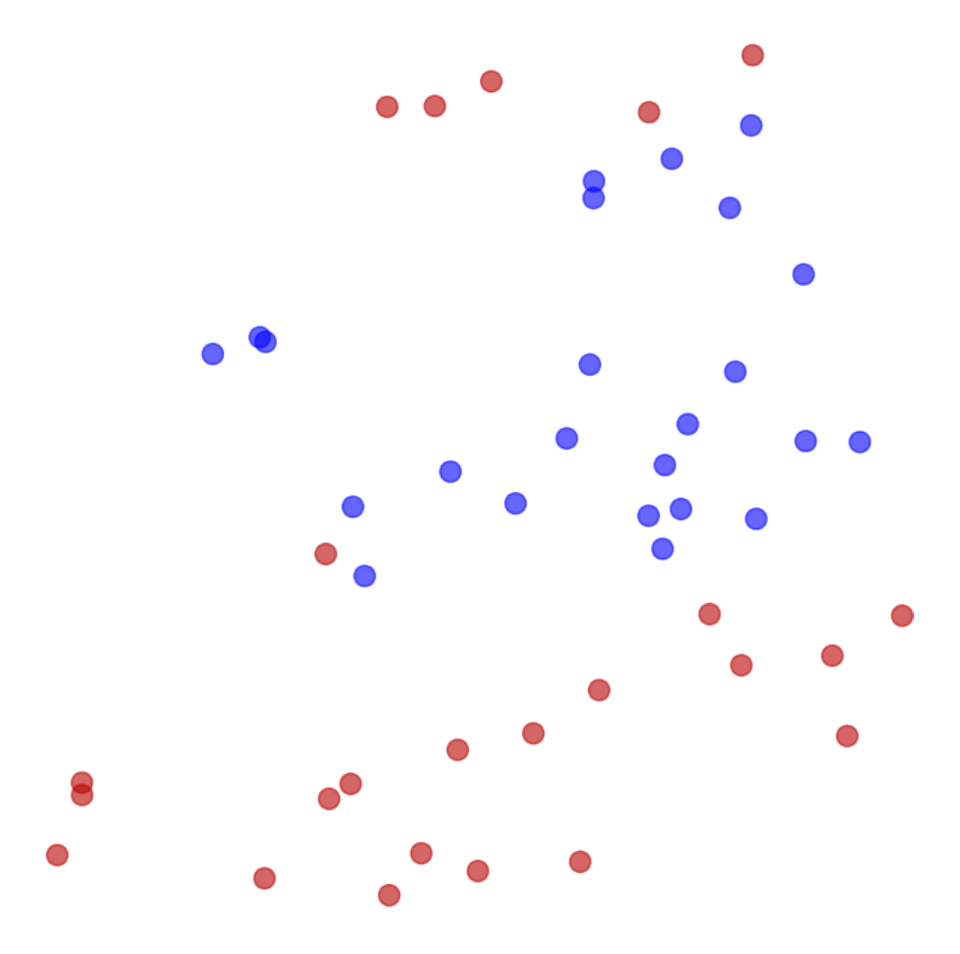}}
  \caption{Mutual Information}
  \label{fig:wtsnelivermi}
\end{subfigure}
\begin{subfigure}[t]{.23\textwidth}
  \centering
  \fbox{\includegraphics[width=.9\linewidth, height=.9\linewidth, alt={Weighted t-SNE scatterplot of the liver cancer dataset using the mRMR feature scores, with samples colored by class}]{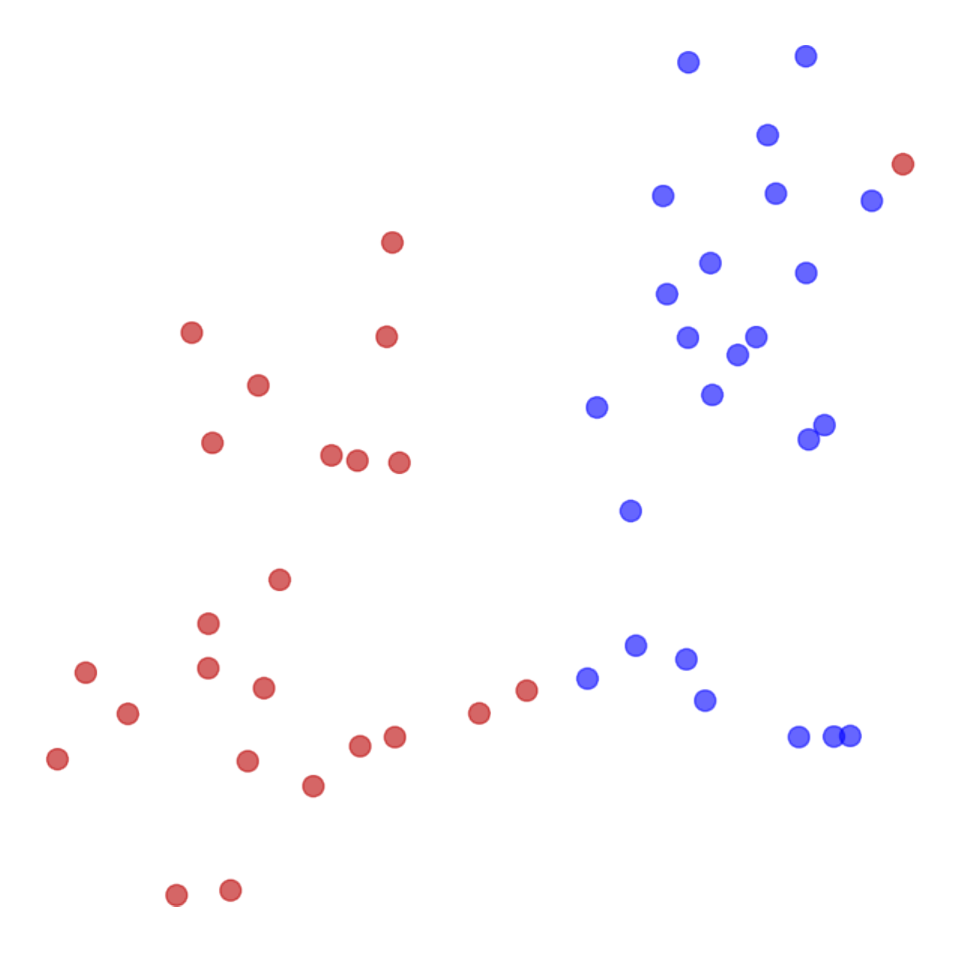}}
  \caption{mRMR}
  \label{fig:wtsnelivermrmr}
\end{subfigure}

\begin{subfigure}[t]{.23\textwidth}
  \centering
  \fbox{\includegraphics[width=.9\linewidth, height=.9\linewidth, alt={Weighted t-SNE scatterplot of the liver cancer dataset using the ReliefF feature scores, with samples colored by class}]{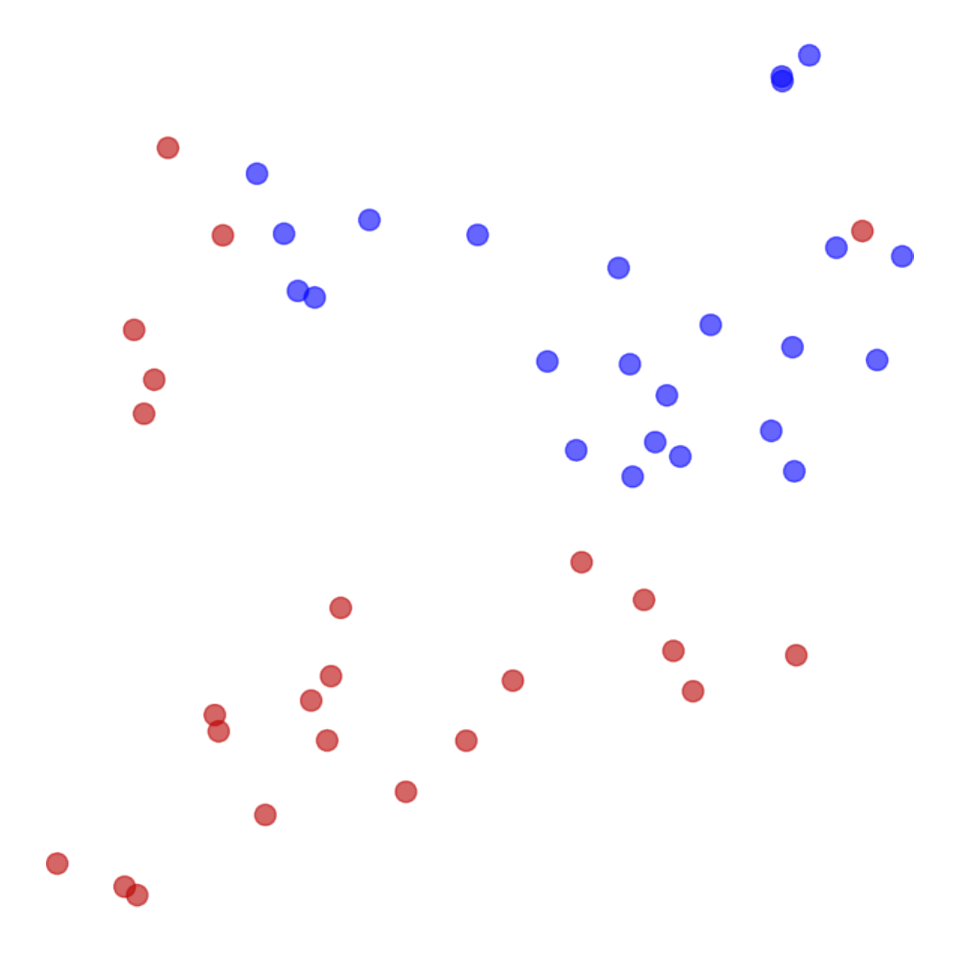}}
  \caption{ReliefF}
  \label{fig:wtsneliverreliefffs}
\end{subfigure}
\begin{subfigure}[t]{.23\textwidth}
  \centering
  \fbox{\includegraphics[width=.9\linewidth, height=.9\linewidth, alt={Weighted t-SNE scatterplot of the liver cancer dataset using the Lasso feature scores, with samples colored by class}]{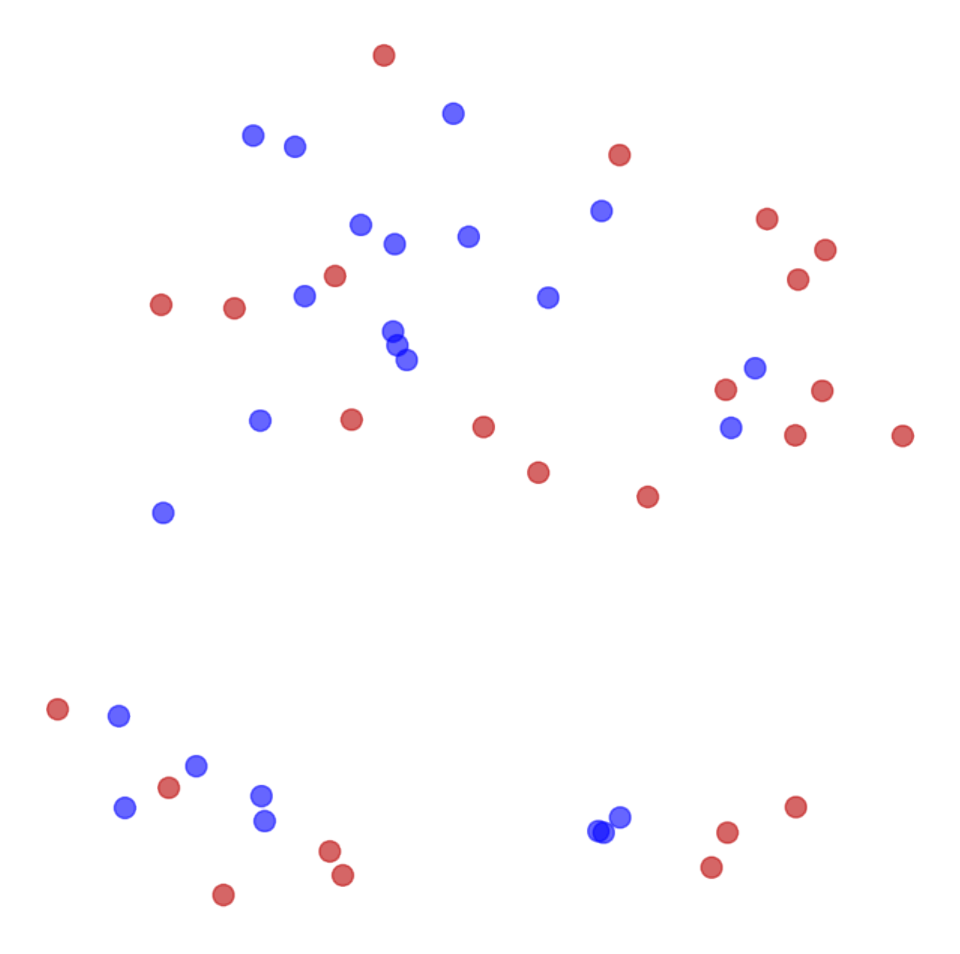}}
  \caption{Lasso}
  \label{fig:wtsneliverlasso}
\end{subfigure}
\begin{subfigure}[t]{.23\textwidth}
  \centering
  \fbox{\includegraphics[width=.9\linewidth, height=.9\linewidth, alt={Weighted t-SNE scatterplot of the liver cancer dataset using the Decision Tree feature scores, with samples colored by class}]{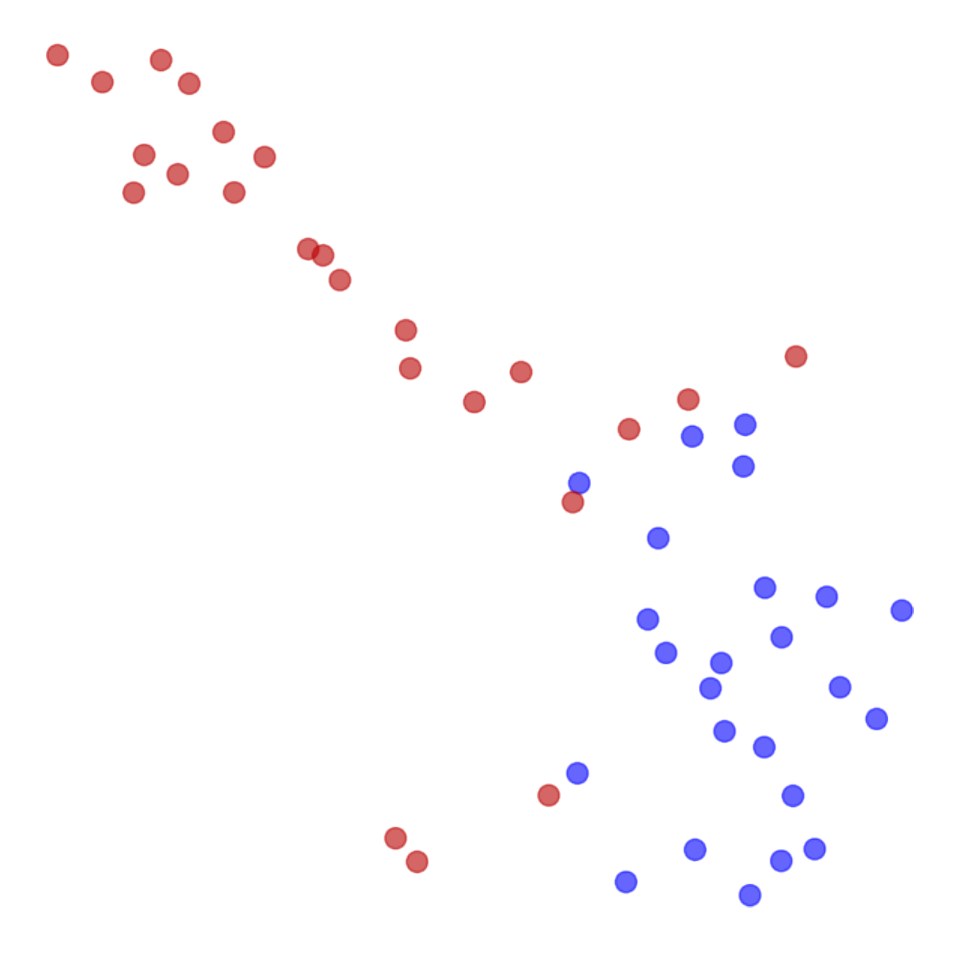}}
  \caption{Decision Tree}
  \label{fig:wtsneliverdt}
\end{subfigure}
\begin{subfigure}[t]{.23\textwidth}
  \centering
  \fbox{\includegraphics[width=.9\linewidth, height=.9\linewidth, alt={Weighted t-SNE scatterplot of the liver cancer dataset using the Random Forest feature scores, with samples colored by class}]{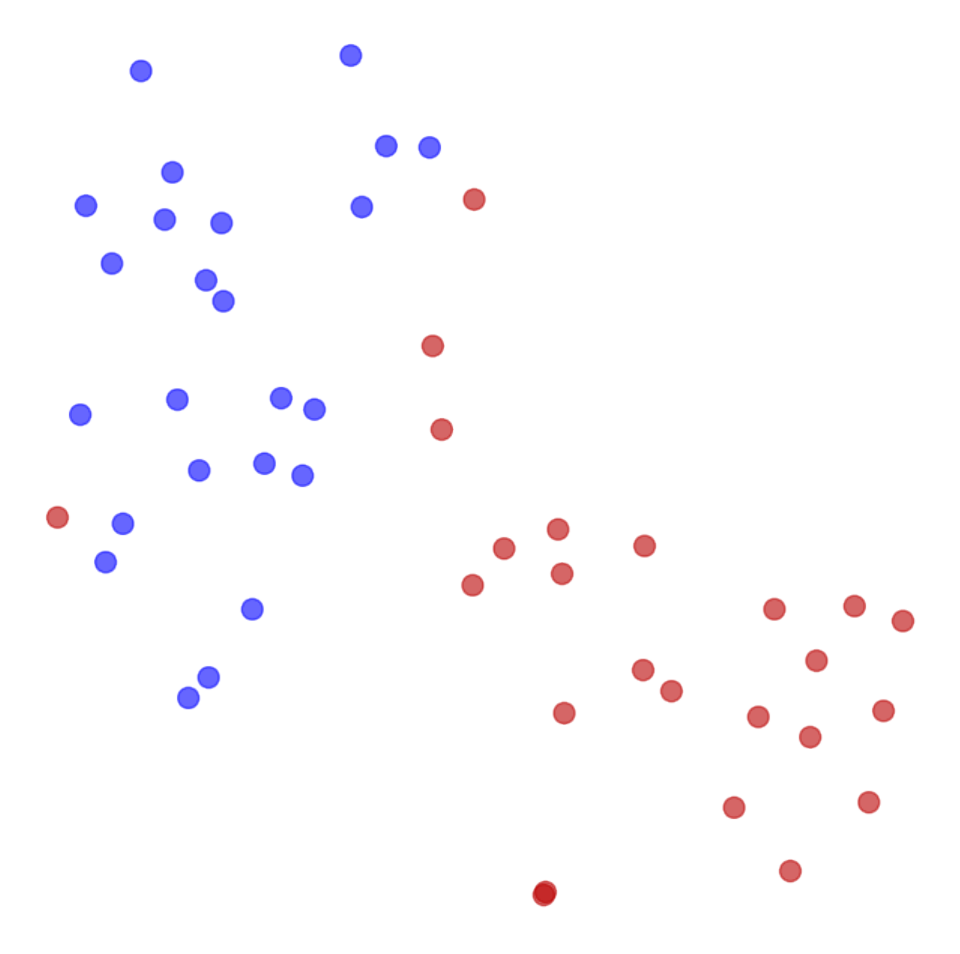}}
  \caption{Random Forest}
  \label{fig:wtsneliverrf}
\end{subfigure}

\begin{subfigure}[t]{.23\textwidth}
  \centering
  \fbox{\includegraphics[width=.9\linewidth, height=.9\linewidth, alt={Weighted t-SNE scatterplot of the liver cancer dataset using the Linear SVM feature scores, with samples colored by class}]{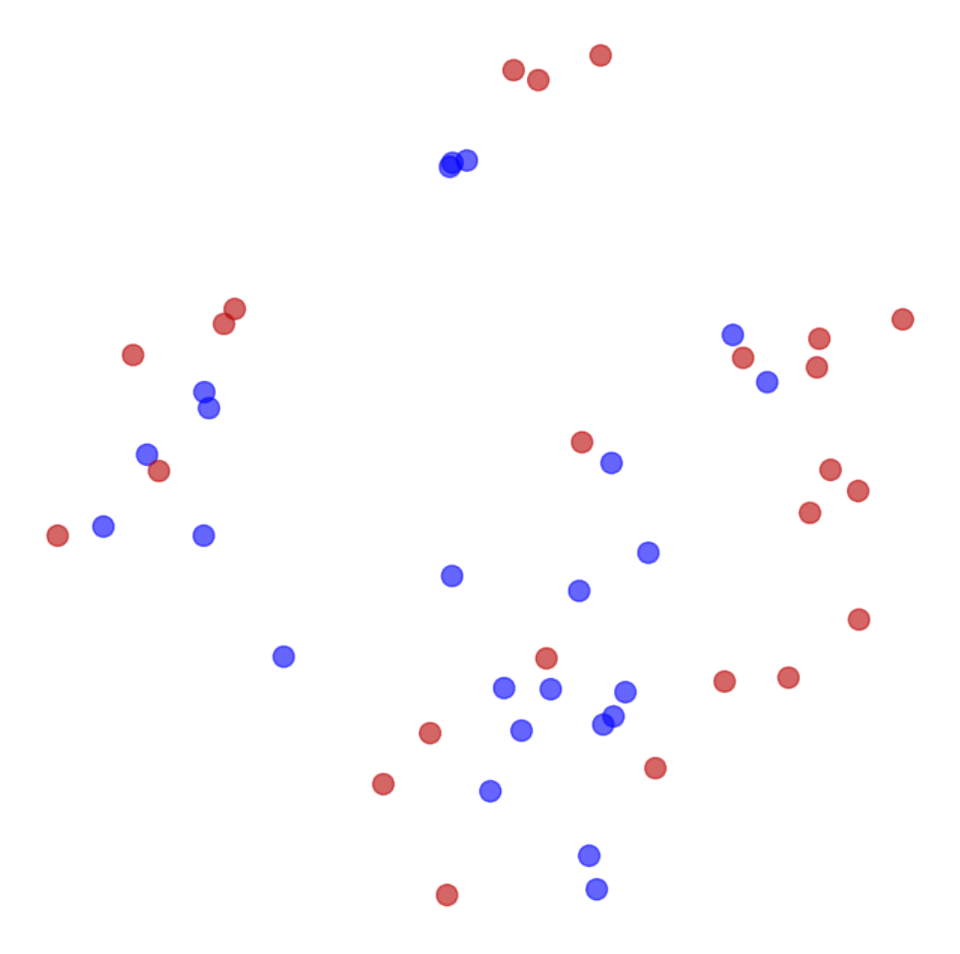}}
  \caption{Linear SVM}
  \label{fig:wtsneliverlinearsvm}
\end{subfigure}
\begin{subfigure}[t]{.23\textwidth}
  \centering
  \fbox{\includegraphics[width=.9\linewidth, height=.9\linewidth, alt={Weighted t-SNE scatterplot of the liver cancer dataset using the neural network (relevance aggregation) feature scores, with samples colored by class}]{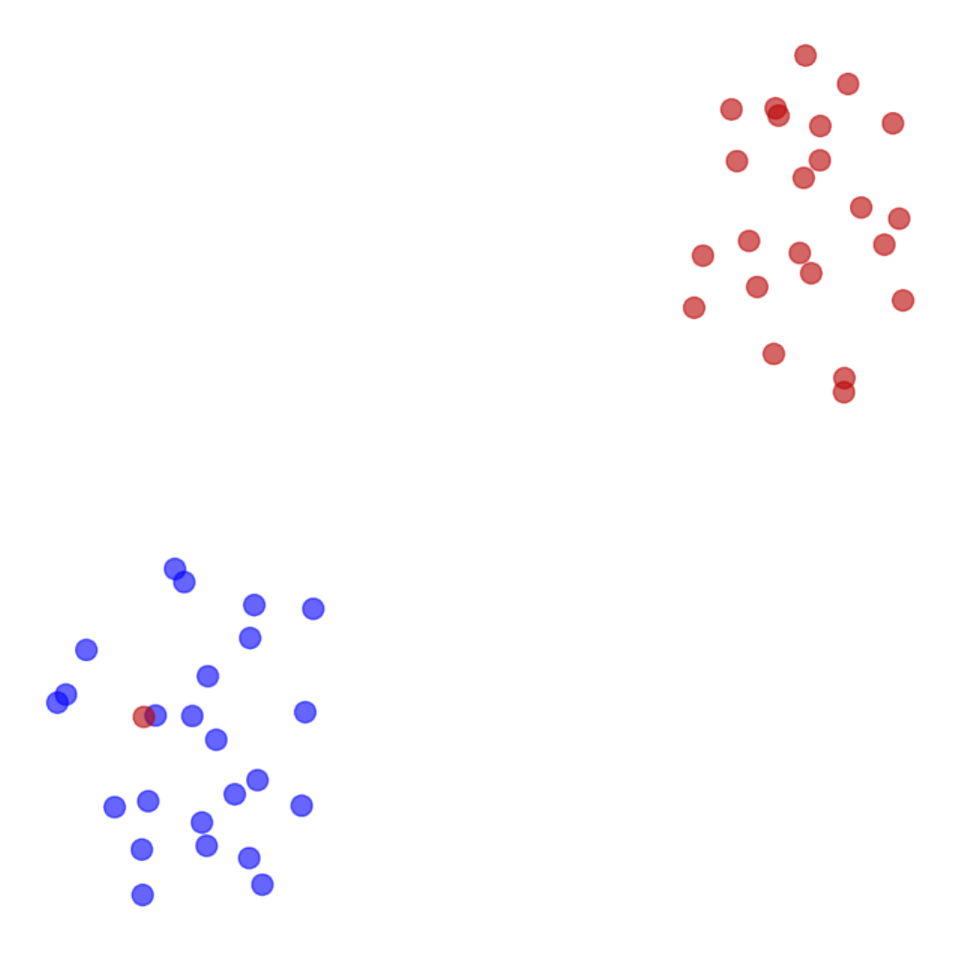}}
  \caption{Neural Network}
  \label{fig:wtsneliverrelagg}
\end{subfigure}
\caption{Weighted t-SNE visualization for the liver cancer dataset. Each red or blue point represents one sample from one of the two classes. The color legend is shown in Fig.~\ref{fig:wtsnelivernoselection} and omitted in the other figures.}
\label{fig:wtsneliver}
\end{figure}

\begin{figure}[tbp]
\centering
\begin{subfigure}[t]{.23\textwidth}
  \centering
  \fbox{\includegraphics[width=.9\linewidth, height=.9\linewidth, alt={t-SNE scatterplot of the prostate cancer dataset without feature scoring, with samples colored by class}]{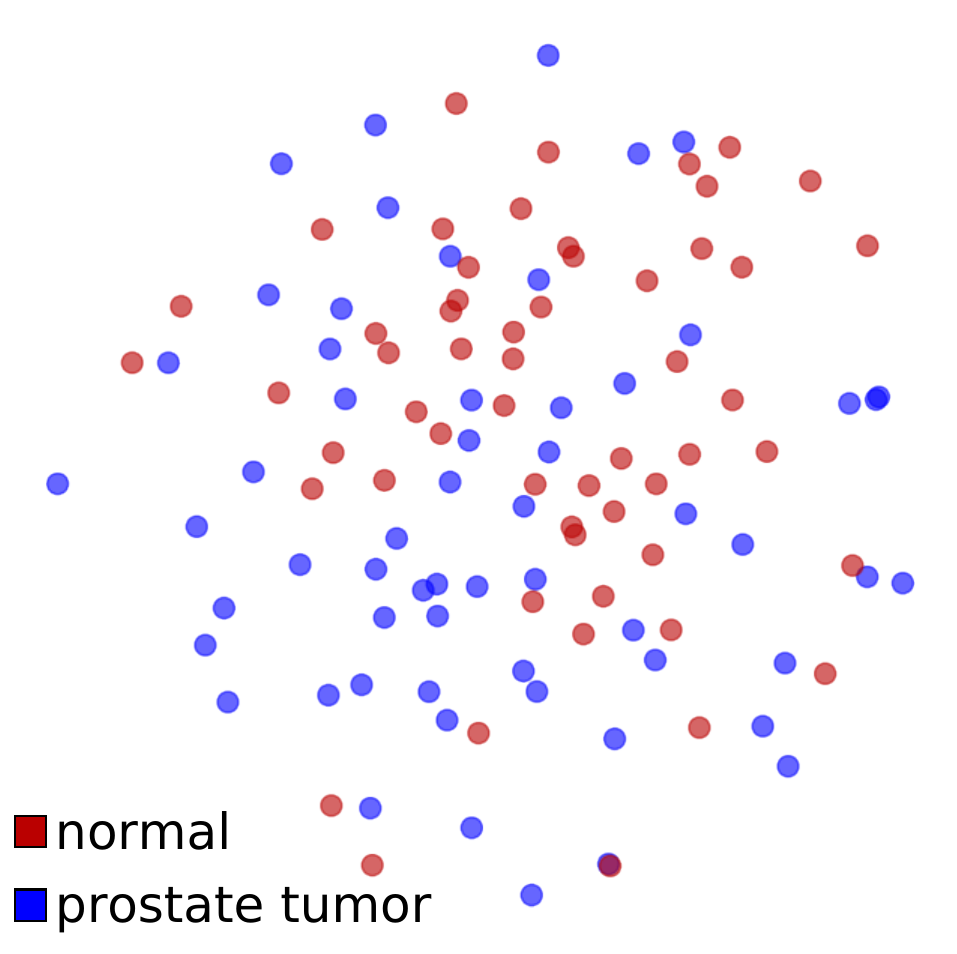}}
  \caption{No scoring}
  \label{fig:wtsneprostateoselection}
\end{subfigure}
\begin{subfigure}[t]{.23\textwidth}
  \centering
  \fbox{\includegraphics[width=.9\linewidth, height=.9\linewidth, alt={Weighted t-SNE scatterplot of the prostate cancer dataset using the Kruskal-Wallis feature scores, with samples colored by class}]{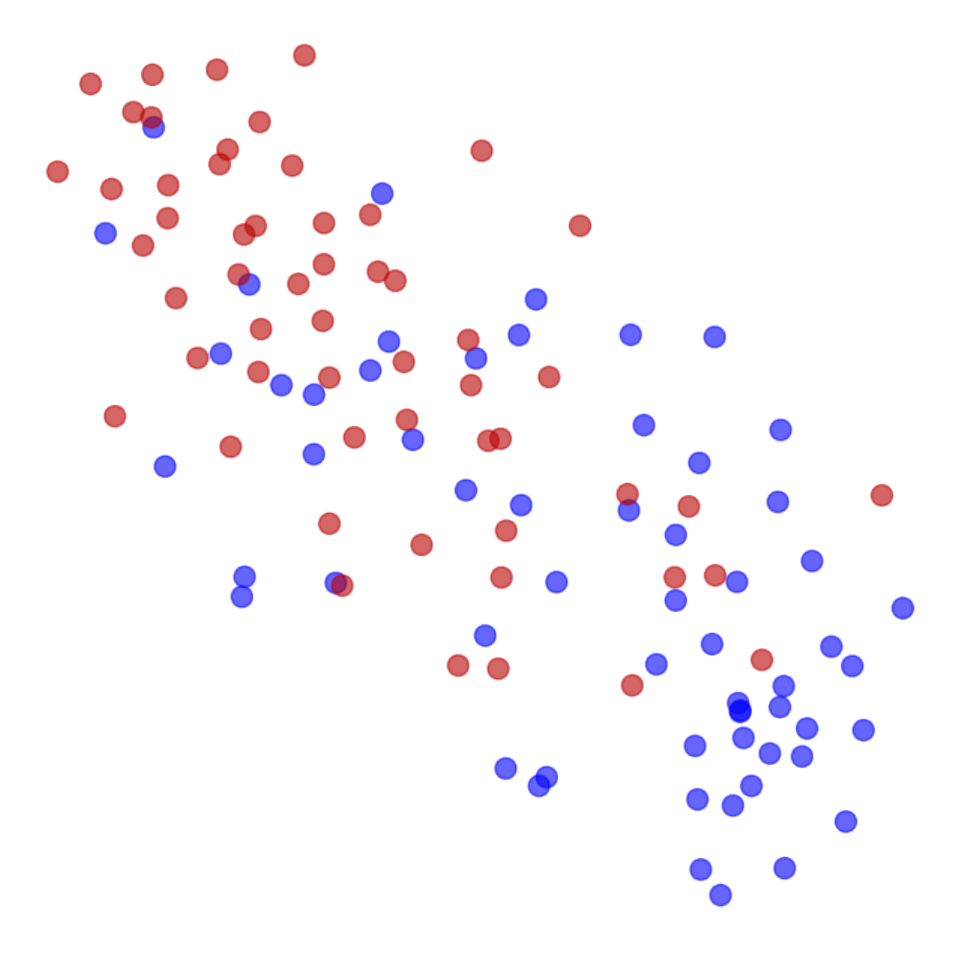}}
  \caption{Kruskal--Wallis}
  \label{fig:wtsneprostatekw}
\end{subfigure}
\begin{subfigure}[t]{.23\textwidth}
  \centering
  \fbox{\includegraphics[width=.9\linewidth, height=.9\linewidth, alt={Weighted t-SNE scatterplot of the prostate cancer dataset using the Mutual Information feature scores, with samples colored by class}]{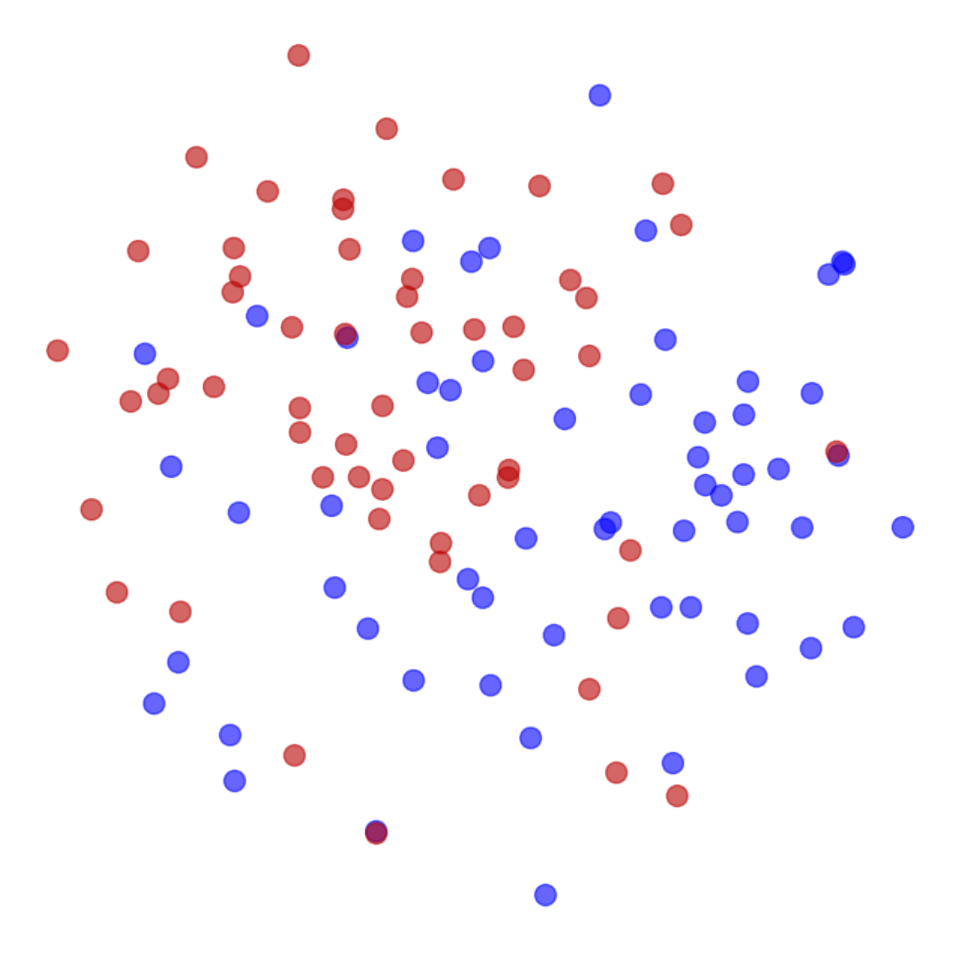}}
  \caption{Mutual Information}
  \label{fig:wtsneprostatemi}
\end{subfigure}
\begin{subfigure}[t]{.23\textwidth}
  \centering
  \fbox{\includegraphics[width=.9\linewidth, height=.9\linewidth, alt={Weighted t-SNE scatterplot of the prostate cancer dataset using the mRMR feature scores, with samples colored by class}]{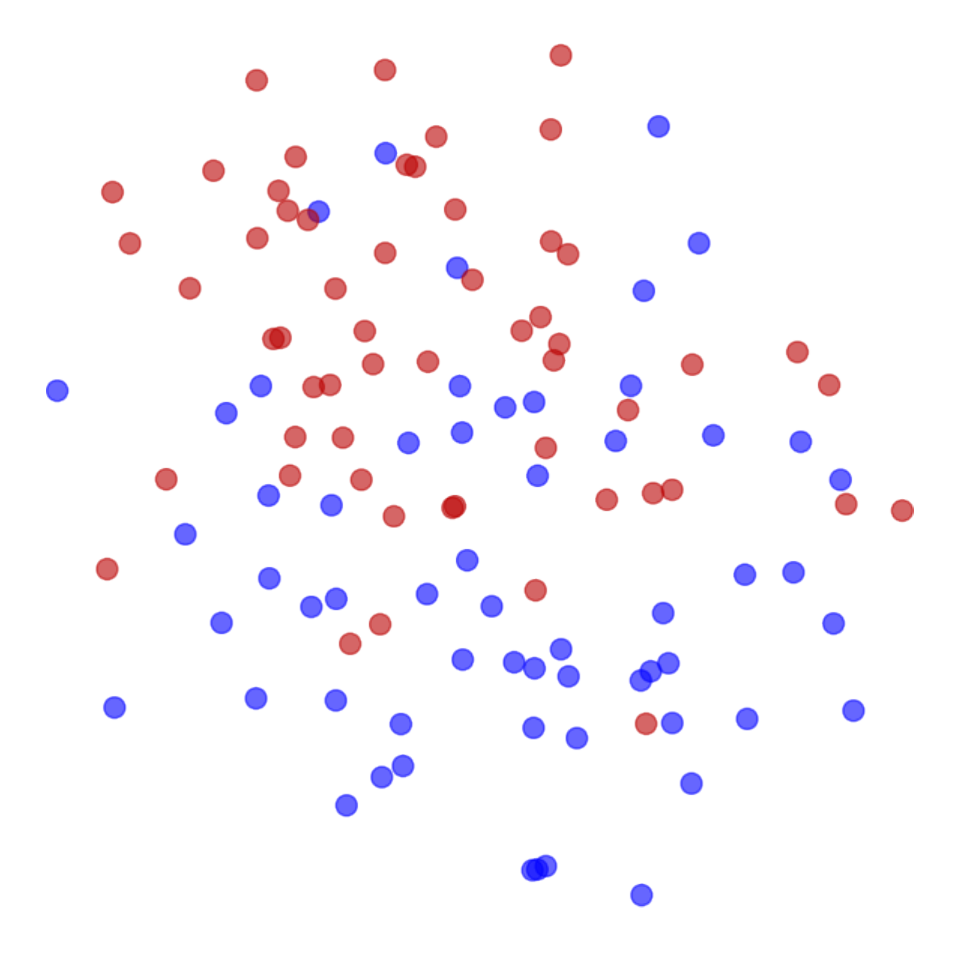}}
  \caption{mRMR}
  \label{fig:wtsneprostatemrmr}
\end{subfigure}

\begin{subfigure}[t]{.23\textwidth}
  \centering
  \fbox{\includegraphics[width=.9\linewidth, height=.9\linewidth, alt={Weighted t-SNE scatterplot of the prostate cancer dataset using the ReliefF feature scores, with samples colored by class}]{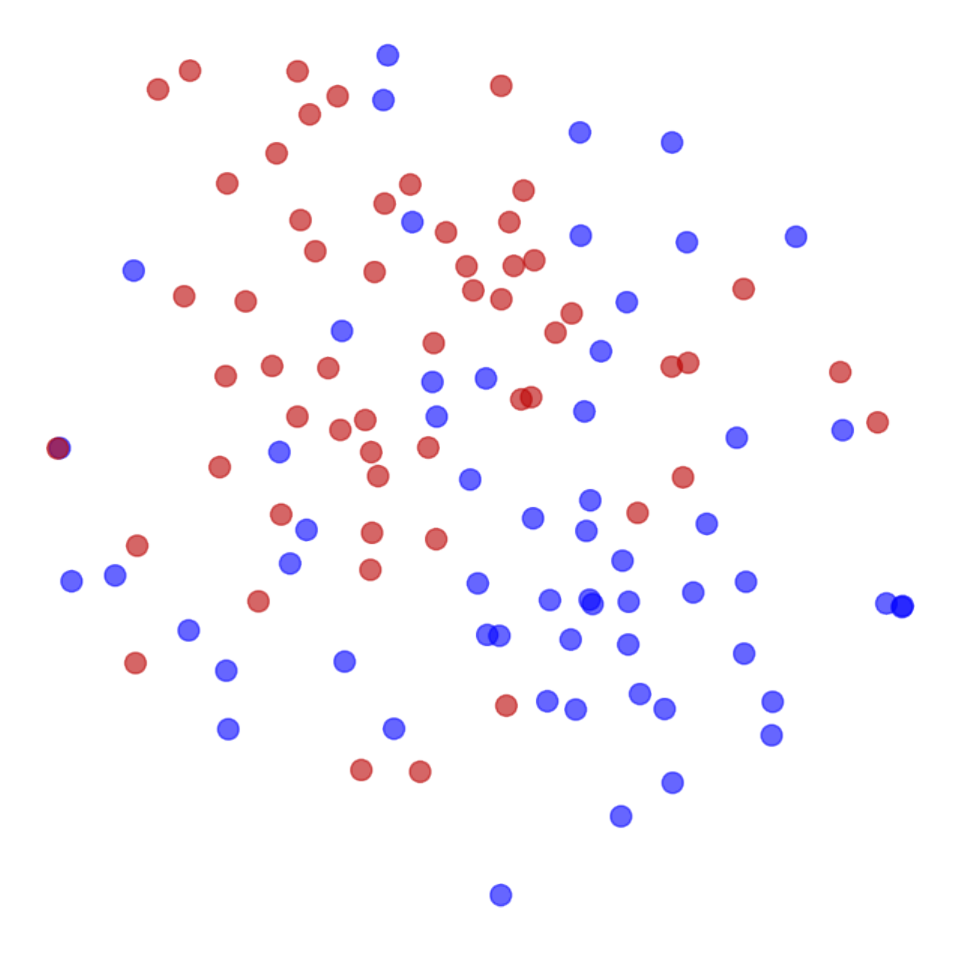}}
  \caption{ReliefF}
  \label{fig:wtsneprostaterelieff}
\end{subfigure}
\begin{subfigure}[t]{.23\textwidth}
  \centering
  \fbox{\includegraphics[width=.9\linewidth, height=.9\linewidth, alt={Weighted t-SNE scatterplot of the prostate cancer dataset using the Lasso feature scores, with samples colored by class}]{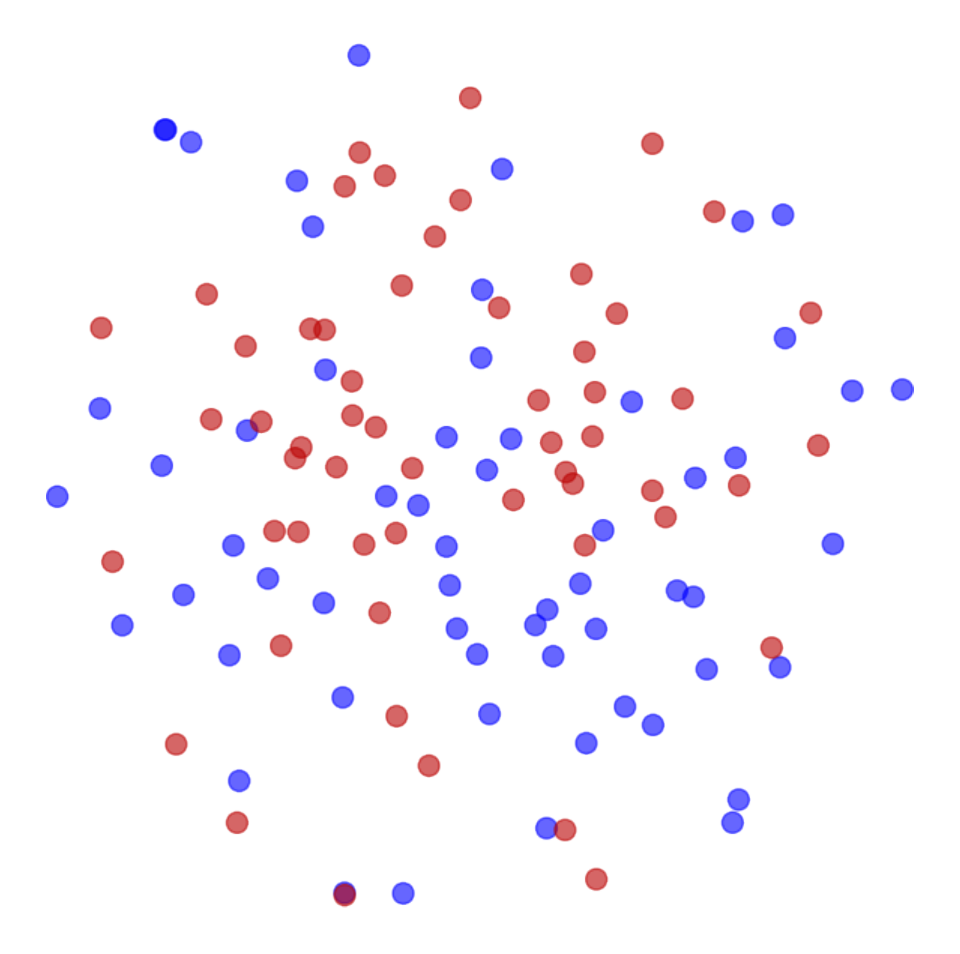}}
  \caption{Lasso}
  \label{fig:wtsneprostatelasso}
\end{subfigure}
\begin{subfigure}[t]{.23\textwidth}
  \centering
  \fbox{\includegraphics[width=.9\linewidth, height=.9\linewidth, alt={Weighted t-SNE scatterplot of the prostate cancer dataset using the Decision Tree feature scores, with samples colored by class}]{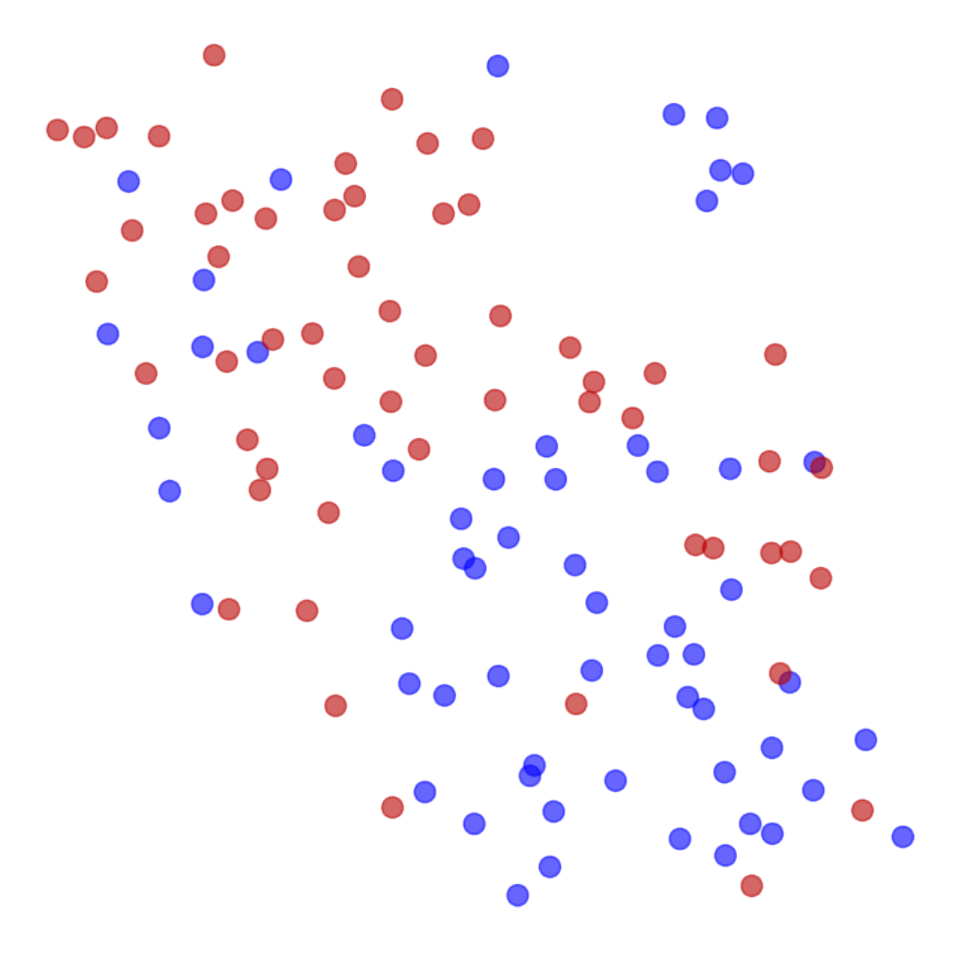}}
  \caption{Decision Tree}
  \label{fig:wtsneprostatedt}
\end{subfigure}
\begin{subfigure}[t]{.23\textwidth}
  \centering
  \fbox{\includegraphics[width=.9\linewidth, height=.9\linewidth, alt={Weighted t-SNE scatterplot of the prostate cancer dataset using the Random Forest feature scores, with samples colored by class}]{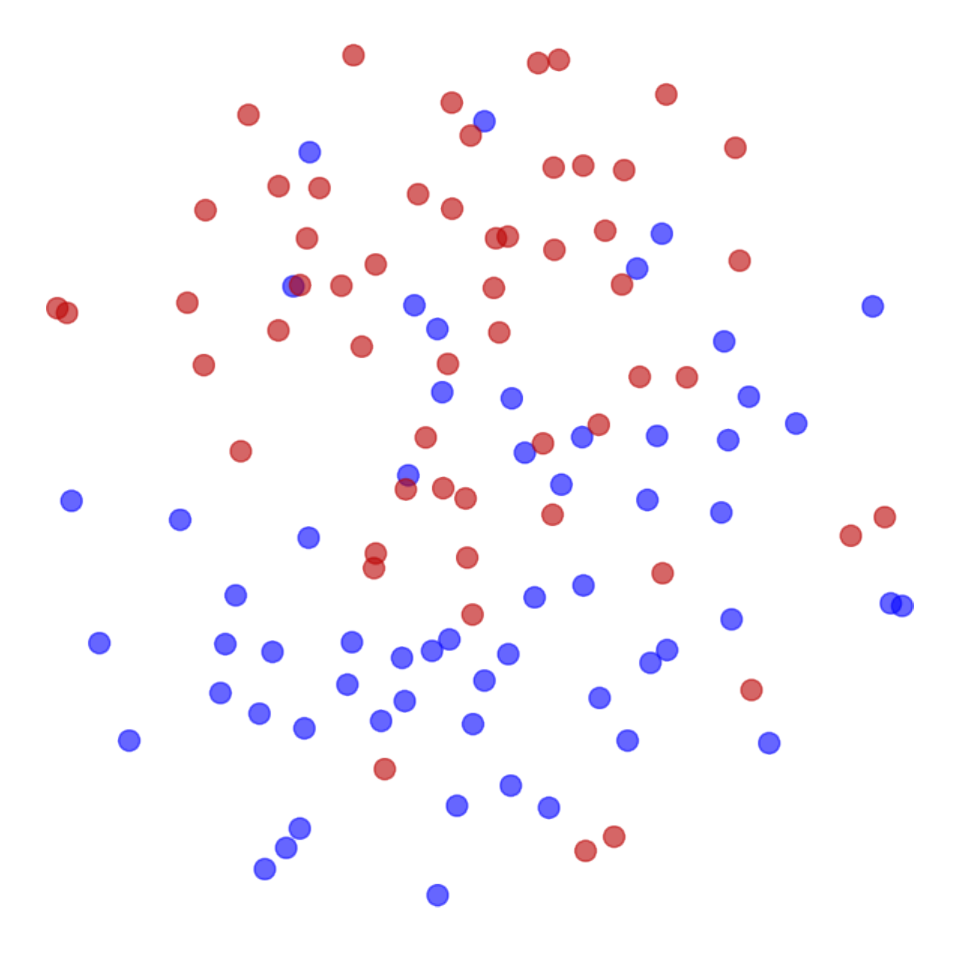}}
  \caption{Random Forest}
  \label{fig:wtsneprostaterf}
\end{subfigure}

\begin{subfigure}[t]{.23\textwidth}
  \centering
  \fbox{\includegraphics[width=.9\linewidth, height=.9\linewidth, alt={Weighted t-SNE scatterplot of the prostate cancer dataset using the Linear SVM feature scores, with samples colored by class}]{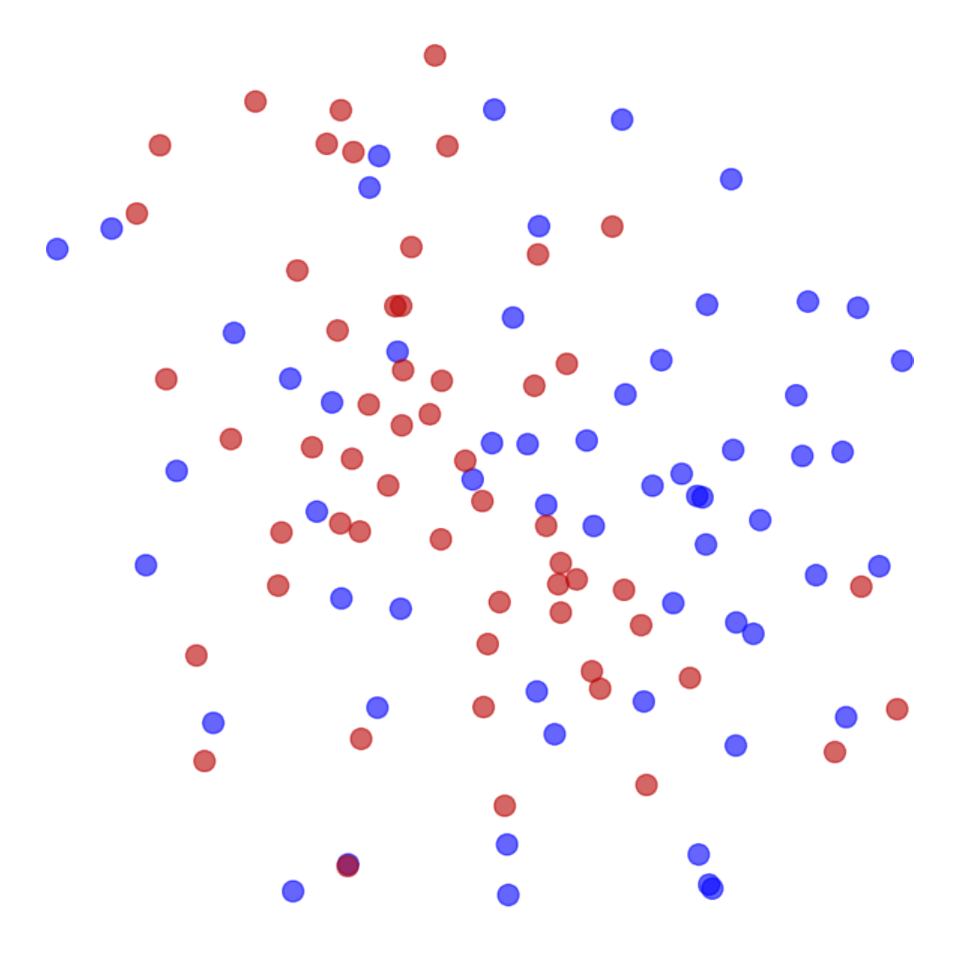}}
  \caption{Linear SVM}
  \label{fig:wtsneprostatelinearsvm}
\end{subfigure}
\begin{subfigure}[t]{.23\textwidth}
  \centering
  \fbox{\includegraphics[width=.9\linewidth, height=.9\linewidth, alt={Weighted t-SNE scatterplot of the prostate cancer dataset using the neural network (relevance aggregation) feature scores, with samples colored by class}]{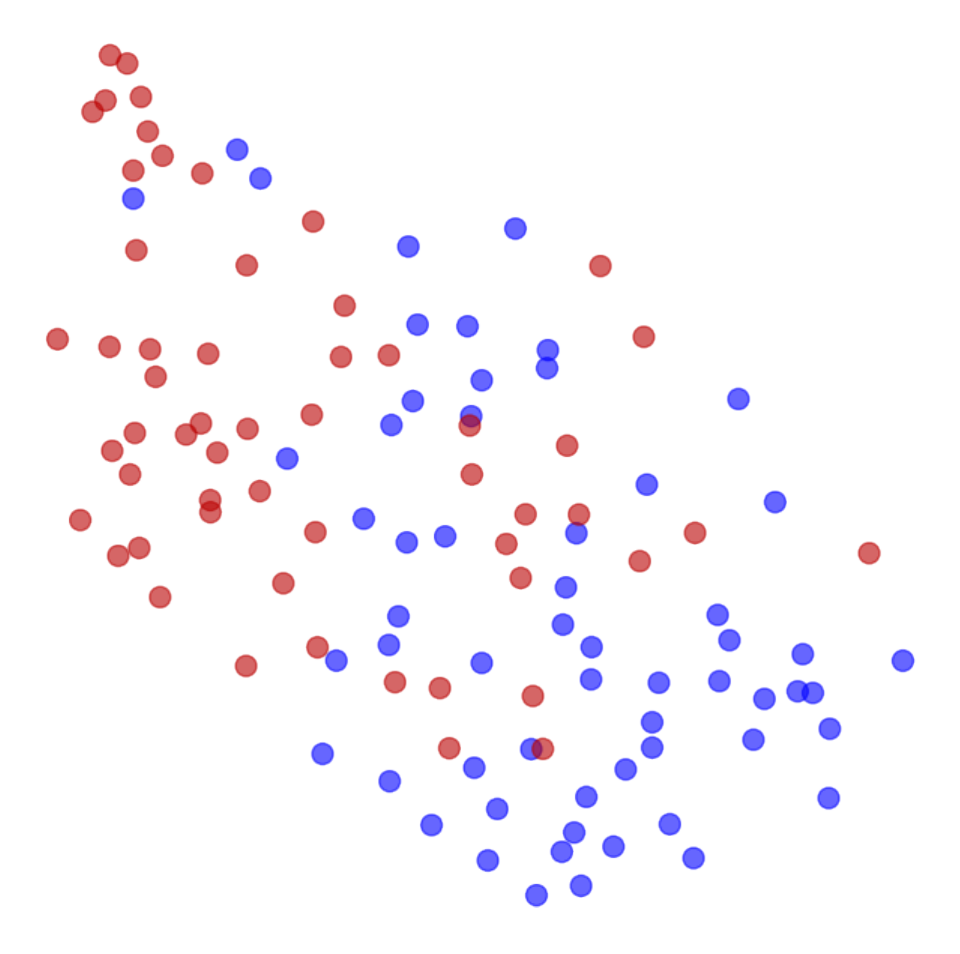}}
  \caption{Neural Network}
  \label{fig:wtsneprostaterelagg}
\end{subfigure}
\caption{Weighted t-SNE visualization for the prostate cancer dataset. Each red or blue point represents one sample from one of the two classes. The color legend is shown in Fig.~\ref{fig:wtsneprostateoselection} and omitted in the other figures.}
\label{fig:wtsneprostate}
\end{figure}

The results for each binary classification dataset are shown in Figs.~\ref{fig:wtsnexor}, \ref{fig:wtsnesynth}, \ref{fig:wtsneliver}, and \ref{fig:wtsneprostate}.
As an example of the utility of the weighted t-SNE, the user can compare the visualizations of each feature scorer to the original data and each other and inspect which ones have better results.
For instance, it is clear from Fig.~\ref{fig:wtsnexornoselection} that the XOR dataset with all the original features does not have distinguishable classes (as expected because most of the features are random).
Considering the data modified to reflect the weights of the feature scorers, some structure starts to appear in the visualizations.
However, for this dataset only ReliefF (Fig.~\ref{fig:wtsnexorrelieff}), Random Forest (Fig.~\ref{fig:wtsnexorrf}), and Neural Network (Fig.~\ref{fig:wtsnexorrelagg}) were able to capture the correct structure of XOR.
The two classes in these figures appear separated into four clusters because each cluster contains samples of one of the four possible combinations of the two relevant features.
For instance, the class $0$ (in red) is split between two clusters ($0 \oplus 0$ and $1 \oplus 1$).
The challenge of this dataset is to identify the non-linear relationship between the variables among very noisy data.
The three scorers with the best visualization were the ones that gave larger importance scores to the only two relevant features.
For instance, the random forest scored the two relevant features with a $0.99$ and a $1.0$, and all other features were scored with less than $0.2$.

\begin{figure}[tbp]
\centering
\begin{subfigure}[t]{.32\linewidth}
  \centering
  \fbox{\includegraphics[width=.9\linewidth, height=.9\linewidth, alt={Weighted t-SNE of the Synth dataset with Kruskal-Wallis scores}]{figures/wtsne/synth/KruskallWallisFilter_Synth_A_tsne2d.pdf}}
  \caption{Kruskal--Wallis}
  \label{fig:perfectwtsnesynthkw}
\end{subfigure}
\begin{subfigure}[t]{.32\linewidth}
  \centering
  \fbox{\includegraphics[width=.9\linewidth, height=.9\linewidth, alt={Weighted t-SNE of the Synth dataset with a perfect selection of the fifty relevant features}]{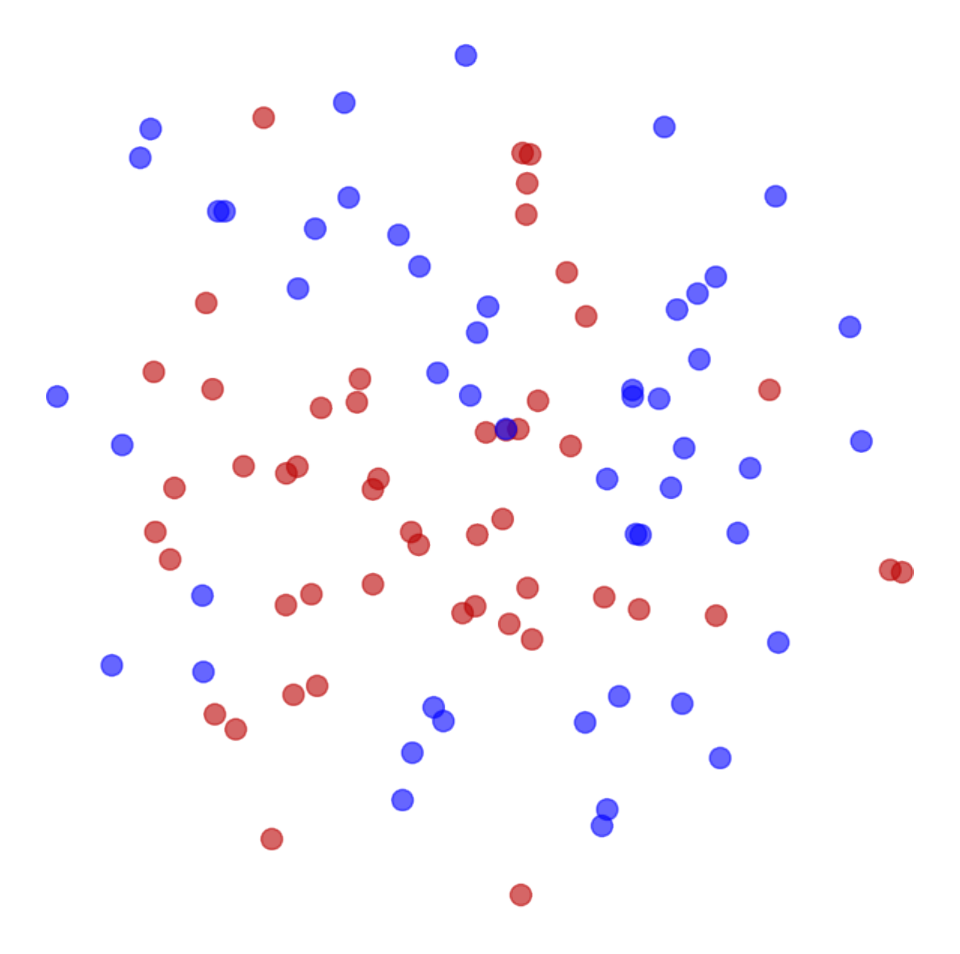}}
  \caption{``Perfect'' selection}
  \label{fig:perfectwtsnesynthperfect}
\end{subfigure}
\caption{Comparison between the Kruskal--Wallis Filter and a ``perfect'' selection. Each figure is the visualization of the 2D embedding with weighted t-SNE for the results of Kruskal--Wallis Filter and a ``perfect'' selection. This visualization refers to the synthetic dataset with $5000$ features, in which only $50$ are relevant. The ``perfect'' solution weights all relevant features with a score of one and all noisy features with a score of zero. Details as in Fig.~\ref{fig:wtsnesynth}.}
\label{fig:wtsnesynthrelieffperfect}
\end{figure}

The same happens for the Synth dataset in Fig.~\ref{fig:wtsnesynth}, which like the XOR dataset, is composed mainly of random features.
However, this time the two algorithms with the apparent best results are the Kruskal--Wallis Filter (Fig.~\ref{fig:wtsnesynthkw}) and the ReliefF (Fig.~\ref{fig:wtsnesynthrelieff}).
The algorithm used to create the Synth dataset \cite{guyon2003design, scikit-learn} builds the classes by initially creating clusters of points normally distributed, which may explain why a simple statistical method like the Kruskal--Wallis Filter was able to obtain a good result.
The size of this dataset also presents a great challenge for models based on classifiers, as it is hard to learn the correct relationships with only 100 data points in the space of five thousand features, of which only 50 are relevant.
Interestingly, Kruskal--Wallis Filter and ReliefF were able to ``outperform'' a visualization with ``perfect'' selection (Fig.~\ref{fig:perfectwtsnesynthkw}).
In this case, a ``perfect'' selection was created by weighing all fifty relevant features with a score of one and all remaining features with zero.
However, this scenario did not create a clear class separation in the final visualization (Fig.~\ref{fig:perfectwtsnesynthperfect}).
This example also illustrates the point made at the end of Section~\ref{subsec:tsne}.
In this case, the ``perfect'' selection is equivalent to the visualization of a feature selection applied to the Synth dataset, in which fifty features were selected.
In feature selection, the importance score of each feature is disregarded, generating different results from the weighted t-SNE approach.
This result shows that incorporating importance scores can reveal a ``hidden'' structure in the visualization.
While not equivalent, both strategies can be used to make complementary visualizations of feature scoring and selection.

The most relevant features (genes) are unknown for the two cancer datasets in Figs.~\ref{fig:wtsneliver} and \ref{fig:wtsneprostate}.
Moreover, the dimensionality is one order of magnitude larger than the synthetic datasets with fewer samples.
These characteristics make for a more challenging analysis, as is often the case with real-world data.
Even so, by inspecting the visualizations of the liver cancer dataset, it is possible to distinguish between scorers that are splitting the two groups (for instance, the mRMR in Fig.~\ref{fig:wtsnelivermrmr} and Neural Network in Fig.~\ref{fig:wtsneliverrelagg}) and those that are not (Figs.~\ref{fig:wtsneliverlasso}, \ref{fig:wtsneliverlinearsvm}).
For the prostate cancer data in Fig.~\ref{fig:wtsneprostate}, most visualizations do not split the two classes.
The use of weighted t-SNE shows that the scorers are not able to score features in a way that visually improved data classification.
In this case, the visualization helps inform the user that further experimentation may be needed.
This outcome is actually expected because this dataset is described as a hard classification and clustering task in the database by \citet{feltes2019cumida}.

Another interesting application of weighted t-SNE is in outlier detection.
In the regular visualization of the liver cancer dataset (Fig.~\ref{fig:wtsnelivernoselection}), it is impossible to identify any outliers because the two classes are entangled.
However, an outlier can be spotted once the features are weighted using their importance scores and the classes appear separated in the visualization.
In Figs.~\ref{fig:wtsnelivermrmr}, \ref{fig:wtsneliverrf}, and \ref{fig:wtsneliverrelagg}, it is possible to see a single hepatocarcinoma sample (red point) clustered together with the normal (healthy) samples (blue points).
This sample that appears out of place has the accession code GSM557108 and is identified as ``liver tissue of subject 15, tumor''.\footnote{\url{https://www.ncbi.nlm.nih.gov/geo/query/acc.cgi?acc=GSM557108}.}
The healthy samples in this dataset were obtained from the adjacent tissue of primary hepatocarcinoma samples,\footnote{\url{https://www.ncbi.nlm.nih.gov/geo/query/acc.cgi?acc=GSE22405}.} so it is possible that this outlier was actually a healthy sample that was mislabeled, or perhaps a tumor sample with contamination from neighboring healthy tissue.

As mentioned in Section~\ref{subsec:silhouette}, the silhouette coefficient can be used to summarize the results of the visualization of feature importance \cite{artur2019novel} (Figs.~\ref{fig:wtsnexor}, \ref{fig:wtsnesynth}, \ref{fig:wtsneliver}, and \ref{fig:wtsneprostate}).
The silhouette coefficient for each scorer and dataset is shown in Table~\ref{tab:silhouette}.
The scorers with the best visual split between classes have the larger silhouette coefficients.
Table~\ref{tab:silhouette} also shows the Kullback--Leibler divergence of each t-SNE optimization in the KL columns.
The complementary Table~\ref{tab:trustworthiness} presents the trustworthiness of each 2D visualization presented in this work, as discussed in Section~\ref{subsec:trustworthiness}.
It was computed using the \emph{scikit-learn} implementation.\footnote{\url{https://scikit-learn.org/stable/modules/generated/sklearn.manifold.trustworthiness.html}.}

\begin{table}[tbp]
\centering
\caption{Silhouette coefficient, KL divergence, and F1-score of a kNN trained on the data for four distinct datasets and nine feature scorers. The HD columns show the silhouette coefficient for the original high-dimensional space, the 2D columns show the silhouette coefficient for the 2D embedding representation learned by t-SNE, and the KL columns show the Kullback--Leibler divergence from the t-SNE optimization. For each scorer, the silhouette coefficient is computed considering the scalarization of the dimensions by their importance, except for ``no scoring''. The kNN column shows the F1-score on a test set (33\% of samples) of a kNN trained on the datasets using the importance scores.}
\label{tab:silhouette}
\resizebox{\textwidth}{!}{%
\begin{tabular}{rlllr|lllr|lllr|lllr}
\hline
\multicolumn{1}{l}{} & \multicolumn{4}{c|}{\textbf{XOR}} & \multicolumn{4}{c|}{\textbf{Synth}} & \multicolumn{4}{c|}{\textbf{Liver}} & \multicolumn{4}{c}{\textbf{Prostate}} \\ \cline{2-17}
\multicolumn{1}{l}{} & \multicolumn{1}{c}{\textbf{HD}} & \multicolumn{1}{c}{\textbf{2D}} & \multicolumn{1}{c}{\textbf{KL}} & \multicolumn{1}{l|}{\textbf{kNN}} & \multicolumn{1}{c}{\textbf{HD}} & \multicolumn{1}{c}{\textbf{2D}} & \multicolumn{1}{c}{\textbf{KL}} & \multicolumn{1}{l|}{\textbf{kNN}} & \multicolumn{1}{c}{\textbf{HD}} & \multicolumn{1}{c}{\textbf{2D}} & \multicolumn{1}{c}{\textbf{KL}} & \multicolumn{1}{l|}{\textbf{kNN}} & \multicolumn{1}{c}{\textbf{HD}} & \multicolumn{1}{c}{\textbf{2D}} & \multicolumn{1}{c}{\textbf{KL}} & \multicolumn{1}{l}{\textbf{kNN}} \\ \hline
\textbf{No scoring}         & 0.000    & 0.003    & 2.12 & 0.58 & 0.001 & $-$0.009 & 0.16 & 0.63 & 0.038 & 0.038 & 0.14 & 0.68 & 0.008 & 0.046 & 0.51 & 0.58 \\
\textbf{Kruskal--Wallis}    & 0.004    & 0.010    & 0.48 & 0.51 & 0.041 & 0.609    & 0.56 & 1.00 & 0.183 & 0.449 & 0.15 & 0.94 & 0.072 & 0.196 & 0.35 & 0.71 \\
\textbf{Mutual Information} & 0.000    & $-$0.001 & 1.19 & 0.54 & 0.006 & 0.180    & 0.69 & 0.66 & 0.133 & 0.234 & 0.15 & 0.94 & 0.025 & 0.121 & 0.49 & 0.68 \\
\textbf{mRMR}               & 0.000    & $-$0.002 & 0.74 & 0.54 & 0.008 & 0.030    & 0.77 & 0.56 & 0.228 & 0.560 & 0.18 & 0.94 & 0.031 & 0.156 & 0.54 & 0.66 \\
\textbf{ReliefF}            & 0.104    & 0.204    & 0.80 & 1.00 & 0.005 & 0.433    & 0.33 & 0.97 & 0.128 & 0.333 & 0.18 & 0.94 & 0.018 & 0.126 & 0.48 & 0.68 \\
\textbf{Lasso}              & 0.000    & 0.005    & 1.54 & 0.55 & 0.001 & 0.052    & 0.16 & 0.60 & 0.039 & 0.033 & 0.14 & 0.68 & 0.008 & 0.027 & 0.51 & 0.55 \\
\textbf{Decision Tree}      & $-$0.001 & $-$0.002 & 0.65 & 0.58 & 0.050 & 0.058    & 0.48 & 0.61 & 0.358 & 0.422 & 0.07 & 0.81 & 0.078 & 0.122 & 0.42 & 0.71 \\
\textbf{Random Forest}      & 0.089    & 0.204    & 0.91 & 1.00 & 0.014 & 0.045    & 0.74 & 0.67 & 0.235 & 0.490 & 0.21 & 0.94 & 0.052 & 0.162 & 0.51 & 0.73 \\
\textbf{Linear SVM}         & 0.000    & $-$0.001 & 1.59 & 0.56 & 0.002 & 0.180    & 0.19 & 0.64 & 0.040 & 0.037 & 0.16 & 0.81 & 0.008 & 0.049 & 0.55 & 0.60 \\
\textbf{Neural Network}     & 0.148    & 0.192    & 0.49 & 1.00 & 0.014 & 0.200    & 0.65 & 0.85 & 0.387 & 0.782 & 0.17 & 0.94 & 0.098 & 0.200 & 0.40 & 0.73 \\ \hline
\end{tabular}%
}
\end{table}

\begin{table}[tbp]
\centering
\caption{Trustworthiness $T(7)$ of the 2D t-SNE and weighted t-SNE projections for all datasets and feature scorers.}
\label{tab:trustworthiness}
\small
\begin{tabular}{lrrrrrr}
\hline
                                & \textbf{XOR} & \textbf{Synth} & \textbf{Liver} & \textbf{Prostate} & \textbf{Regression} & \textbf{Mouse cortex} \\ \hline
\textbf{No scoring}             & 0.814        & 0.653          & 0.939          & 0.809             & 0.714               & 0.548 \\
\textbf{Kruskal--Wallis}        & 0.997        & 0.863          & 0.949          & 0.896             & --                  & -- \\
\textbf{Mutual Information}     & 0.978        & 0.728          & 0.949          & 0.843             & --                  & -- \\
\textbf{mRMR}                   & 0.990        & 0.792          & 0.937          & 0.821             & --                  & -- \\
\textbf{ReliefF}                & 0.978        & 0.701          & 0.954          & 0.826             & --                  & -- \\
\textbf{Lasso}                  & 0.934        & 0.648          & 0.943          & 0.815             & --                  & 0.995 \\
\textbf{Decision Tree}          & 0.987        & 0.921          & 0.984          & 0.930             & --                  & -- \\
\textbf{Random Forest}          & 0.974        & 0.806          & 0.937          & 0.871             & --                  & 0.998 \\
\textbf{Linear SVM}             & 0.930        & 0.647          & 0.937          & 0.797             & --                  & -- \\
\textbf{Neural Network}         & 0.993        & 0.834          & 0.956          & 0.921             & 0.681               & 0.979 \\ \hline
\end{tabular}
\end{table}

The first analysis to be made is to compare the silhouette coefficients of each scoring to the coefficient obtained without scores as a baseline to be beaten.
Table~\ref{tab:silhouette} also allows comparing the original high-dimensional space and the 2D embedded space from t-SNE.
Because of the behavior of distance metrics in high dimensions \cite{aggarwal2001surprising}, the silhouettes in the original space tend to be closer to zero than in the embedding space.
The expected behavior is for the patterns of scorers with larger or smaller coefficients to be preserved in the high-dimensional and embedded spaces.

A user interested in incorporating a feature scorer in a larger analysis pipeline can then use weighted t-SNE to decide which of the available options better suits their specific dataset or task.
The user would select a few scorers they wish to inspect, get the importance of the features from each of them, and then generate the weighted visualizations for each scorer being tested and a regular t-SNE plot for the dataset.
Then, they can use the visualization and the corresponding silhouette coefficients to select the scorer that produced the best or expected results.
The following experiment illustrates and validates this idea.
A simple k-nearest neighbors (kNN) \cite{altman1992introduction} classifier ($k = 3$) was trained on each of the datasets using the different scorers (including the baseline case with no scores).
The kNN was chosen to avoid biases in the results from other classifiers used as scorers, such as SVM or decision tree.
The results in Table~\ref{tab:silhouette} are the F1-scores \cite{tharwat2020classification} of the kNN on the test sets (a 33\% split from the complete dataset).
As can be seen, the feature scorers with better silhouette coefficients achieved higher classification performance.
Thus, together with other metrics and analyses, the weighted t-SNE visualization can be used to select the best feature scorers for a classification pipeline.

\subsection{Machine learning interpretability}\label{subsec:machinelearning}

Even though machine learning has become widespread in the past years and many models achieved state-of-the-art results in challenging tasks, the more complex models, especially deep learning or large ensemble models such as random forests, are still widely regarded as ``black-boxes'' \cite{molnar2022interpretable, grisci2021relevance}.
Their intrinsically complex structures make explaining or predicting their behaviors hard.
The learned features are only implicitly described by many internal model parameters \cite{garcia2018task}.
The lack of explainable decisions can lead to under-performance, distrust, or machine bias \cite{lamy2019explainable, molnar2022interpretable, prates2019assessing}.
The techniques that help humans to understand the cause of a decision made by a machine learning model are known as interpretable or explainable machine learning.

The experiments above show that weighted t-SNE can be used to visualize the feature importance learned by a few popular machine learning models, such as decision trees, random forest, Lasso, SVM, and neural networks.
For the first two, the feature importance measures the total reduction of the impurity brought by that feature (Gini importance), or in other words, how well that feature is splitting the classes inside the model \cite{hastie2009elements, molnar2022interpretable}.
The reduction of variance or Gini index in relation to the parent node should be summed for all splits using the feature and scaled to 100 to compute the overall importance of a feature in a decision tree \cite{molnar2022interpretable}.
For the linear SVM and Lasso, both being linear models, the coefficients assigned to the features during training time can be used to measure their importance to the model.
For neural networks, the feature importance was computed using relevance aggregation \cite{grisci2021relevance}, an algorithm suited for tabular data that uses Layer-wise Relevance Propagation (LRP) \cite{bach2015pixel} to measure the impact each feature has in the output of the network.
The calculation of feature importance for each model is described in more detail in the \hyperref[app:supp]{Supplementary material}.

Weighted t-SNE can help interpret these methods by visually displaying the impact of the importance of the learned features.
In the example shown in Fig.~\ref{fig:teaser}, it is visible that the neural network is giving more importance to features that better separate the samples across their target values.
The synthetic regression dataset has only four truly relevant features out of $100$, and the neural network in Fig.~\ref{fig:teaser} gave the largest importance scores to three of them ($0.44$, $0.40$, $0.25$, respectively, while the next largest score of $0.20$ was given to an irrelevant feature and the fourth truly relevant feature received a score of $0.11$).
In the example of Fig.~\ref{fig:wtsnexor}, one can visually intuit that the neural network and the random forest learned a correct representation of the problem and that the decision tree and the linear SVM did not before even seeing the accuracy of these models.
It is important to highlight that the linear SVM was already expected not to be able to perform well on the XOR dataset, a non-linear problem, as it, by definition, cannot find non-linear relations between features.
However, as the goal of the experiments is to analyze the visualizations and not the scorers, the weighted t-SNE visualization of the linear SVM results is accomplishing its goal of showing that this is not a suitable scorer for the task.

\begin{figure}[tbp]
\centering
\begin{subfigure}[t]{.45\linewidth}
  \fbox{\includegraphics[width=.9\linewidth, height=.9\linewidth, alt={t-SNE scatterplot of the mouse cortex dataset without feature scoring, with cells colored by cell type}]{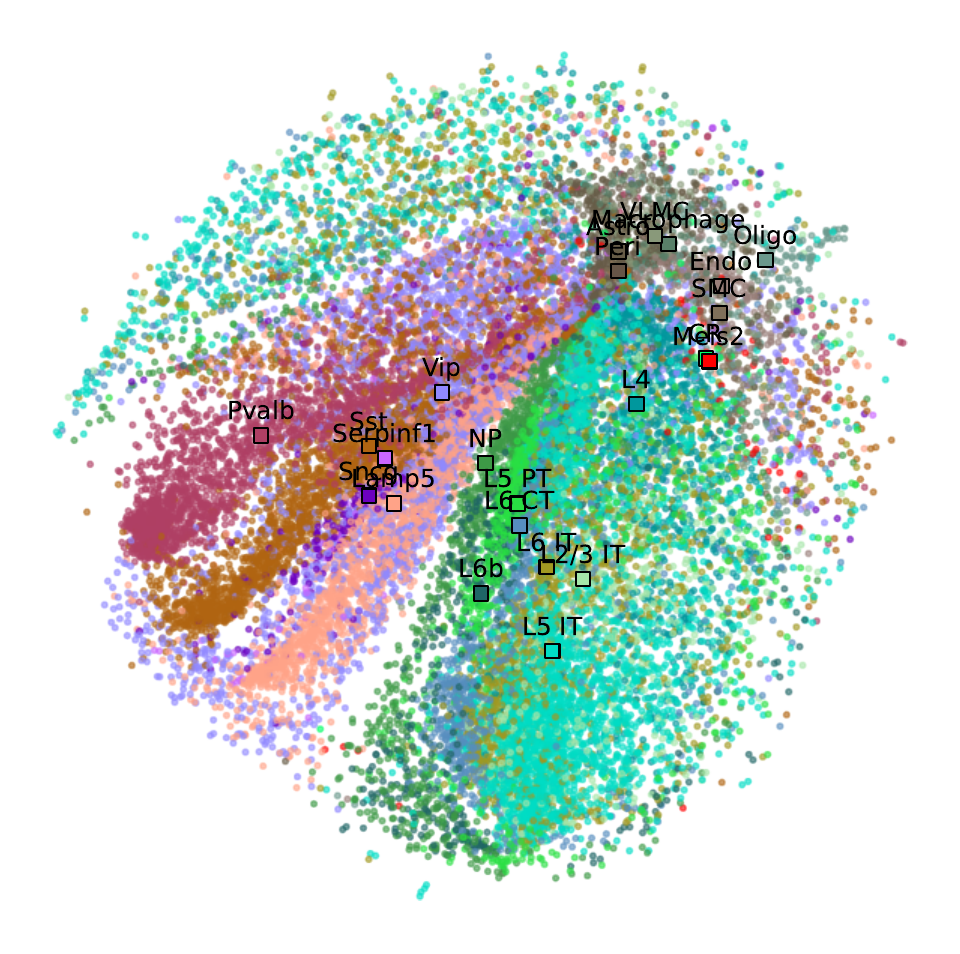}}
  \caption{No scoring \\ F1-score: - $\lvert$ SC: $-$0.190}
  \label{fig:mousetsne}
\end{subfigure}
\begin{subfigure}[t]{.45\linewidth}
  \fbox{\includegraphics[width=.9\linewidth, height=.9\linewidth, alt={Weighted t-SNE scatterplot of the mouse cortex dataset using Lasso feature scores, with cells colored by cell type}]{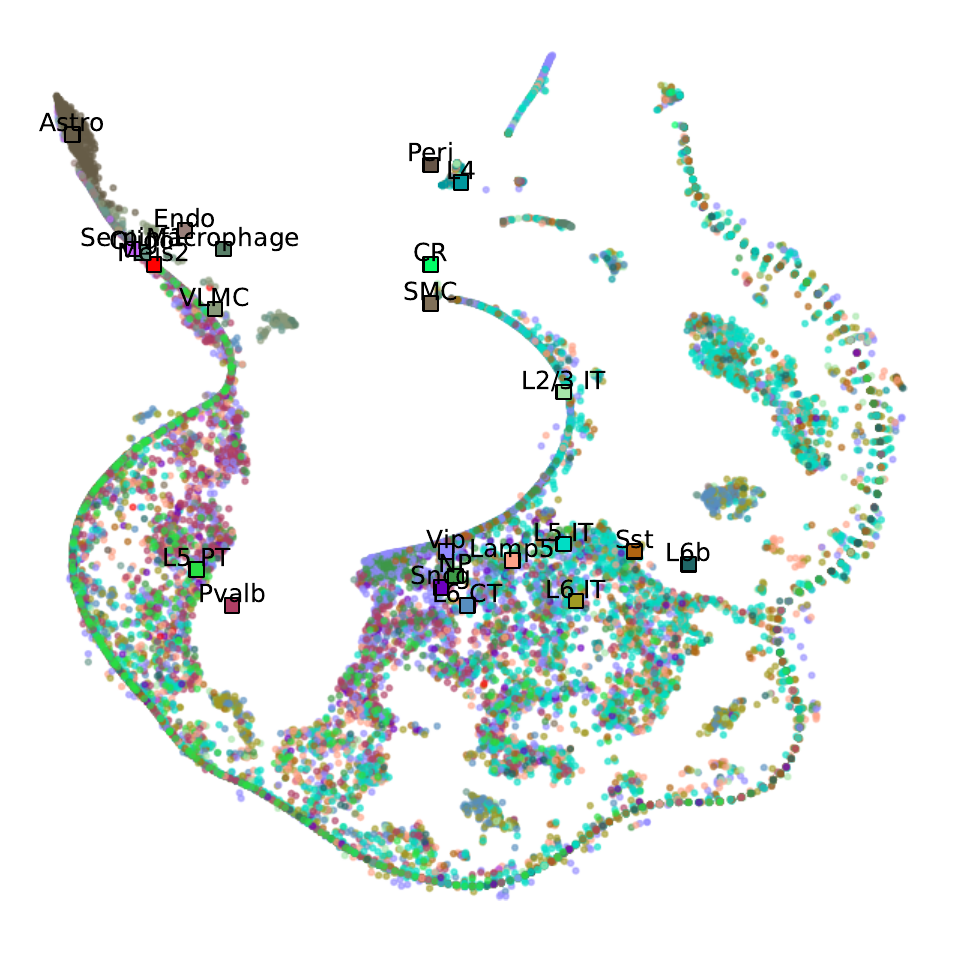}}
  \caption{Lasso \\ F1-score: 0.43 $\lvert$ SC: $-$0.226}
  \label{fig:mouselasso}
\end{subfigure}
\\
\begin{subfigure}[t]{.45\linewidth}
  \fbox{\includegraphics[width=.9\linewidth, height=.9\linewidth, alt={Weighted t-SNE scatterplot of the mouse cortex dataset using random forest feature scores, with cells colored by cell type}]{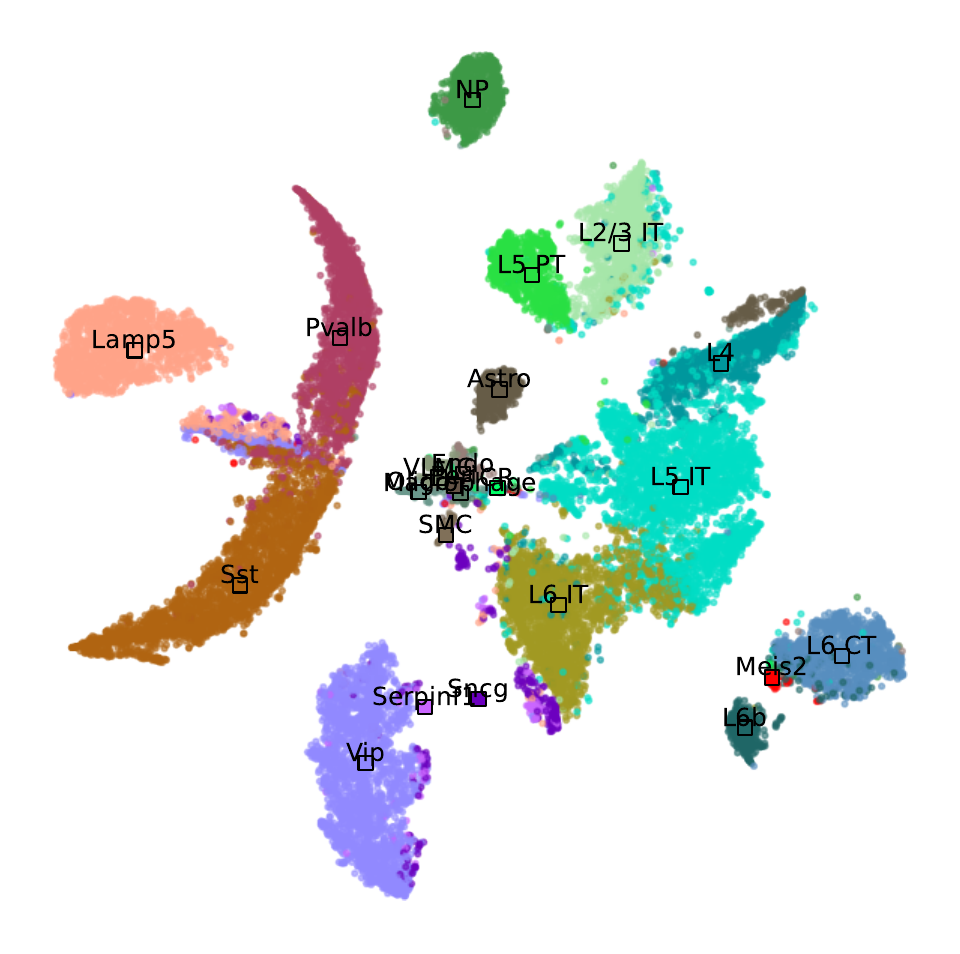}}
  \caption{Random Forest \\ F1-score: 0.72 $\lvert$ SC: 0.346}
  \label{fig:mouserf}
\end{subfigure}
\begin{subfigure}[t]{.45\linewidth}
  \fbox{\includegraphics[width=.9\linewidth, height=.9\linewidth, alt={Weighted t-SNE scatterplot of the mouse cortex dataset using neural network feature scores, with cells colored by cell type}]{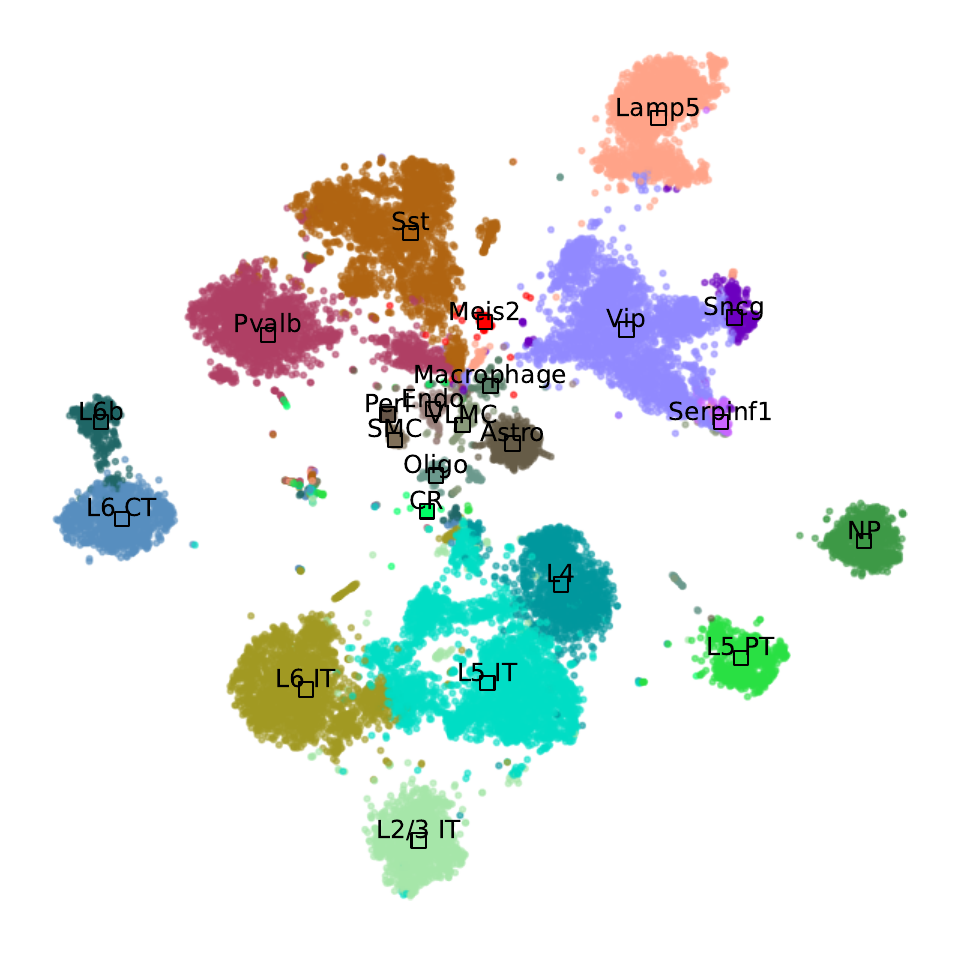}}
  \caption{Neural Network \\ F1-score: 0.98 $\lvert$ SC: 0.385}
  \label{fig:mousenn}
\end{subfigure}
\caption{Weighted t-SNE visualization for the mouse cortex dataset. Each figure is the visualization of the 2D embedding with weighted t-SNE for three machine learning models, plus the original data in ``no scoring''. The captions describe the F1-score of each model and the silhouette coefficient (SC) of each 2D embedding. Each color corresponds to a different class. Following the scheme from \citet{kobak2019art}, warm colors correspond to inhibitory neurons, cold colors correspond to excitatory neurons, and brown or gray colors correspond to non-neural cells.}
\label{fig:wtsnemouse}
\end{figure}

Fig.~\ref{fig:wtsnemouse} shows the results for larger real-world datasets.
The mouse cortex dataset has over 20 thousand samples, over 40 thousand features, and 23 classes.
Visualizing the scores learned by three distinct machine learning models, namely Lasso (Fig.~\ref{fig:mouselasso}), random forest (Fig.~\ref{fig:mouserf}), and neural network (Fig.~\ref{fig:mousenn}) using weighted t-SNE reveals that the models with better performance (larger F1-score) also produced a better split of samples in the 2D embedding (larger silhouette coefficient).
The regular t-SNE visualization of the dataset (Fig.~\ref{fig:mousetsne}) may appear different from other publications, for instance, from its depiction by \citet{kobak2019art}, because we purposely did not perform feature selection and other filtering protocols running the t-SNE algorithm.
This is necessary because we wish to compare feature scorers without biases introduced by other algorithms.

\begin{figure}[tbp]
\centering
\begin{subfigure}[t]{.32\linewidth}
  \centering
  \fbox{\includegraphics[width=.9\linewidth, height=.9\linewidth, alt={Weighted t-SNE scatterplot of the XOR dataset using neural network feature scores, showing four clusters}]{figures/wtsne/xor/xor150.pdf}}
  \caption{Weighted t-SNE}
  \label{fig:twtsnexorNN}
\end{subfigure}
\begin{subfigure}[t]{.32\linewidth}
  \centering
  \fbox{\includegraphics[width=.9\linewidth, height=.9\linewidth, alt={Scatterplot of the activations of the two last-layer neurons of the neural network for the XOR dataset, showing two classes}]{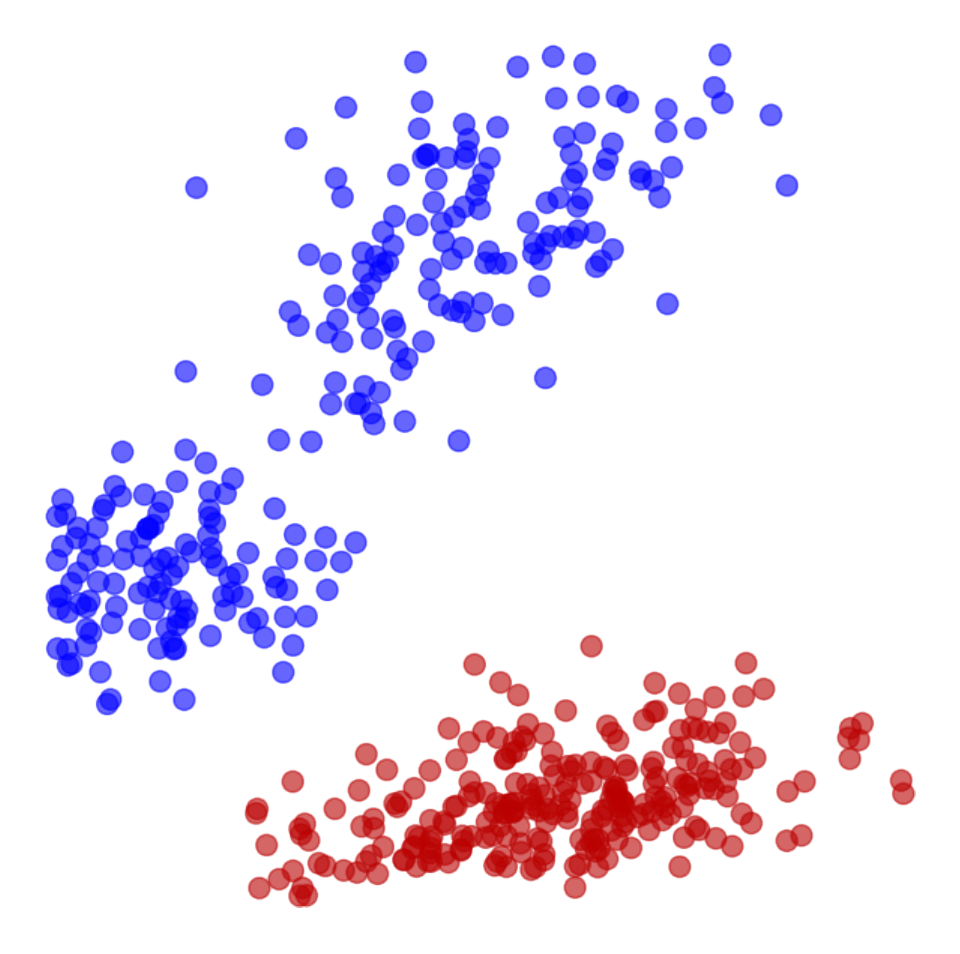}}
  \caption{Activation of last layer neurons}
  \label{fig:actxorNN}
\end{subfigure}
\caption{Comparison between the weighted t-SNE and the activation of the neurons in the last layer of a neural network. On the left is the same weighted t-SNE presented in Fig.~\ref{fig:wtsnexorrelagg}, and on the right is the neurons' activation of the last layer (before the softmax function) of the same neural network. Because this layer only has two neurons, the activation of the first neuron is on the $x$-axis and the activation of the second neuron is on the $y$-axis. Visualization details of the XOR dataset as in Fig.~\ref{fig:wtsnexor}.}
\label{fig:wtsneactxor}
\end{figure}

Regarding the visualization of the importance scores obtained with neural networks, especially in the case of weighted t-SNE (Fig.~\ref{fig:twtsnexorNN}), it is essential to differentiate it from the visualization made by projecting the layers activation (Fig.~\ref{fig:actxorNN}) (the internal state of the network for a given input) \cite{cantareira2020exploring}.
In the case discussed here, what is visualized is the feature importance of the original data dimensions as learned by the neural networks.
So it is a direct representation of the input features.
In the second case, the visualization is of the neural network's internal data representation and does not directly map to the input features.
One option does not invalidate the other, and both can be used together to increase the understanding of a model, as presented in Fig.~\ref{fig:wtsneactxor}.
While both visualizations show that the neural network learned to split the two classes of the XOR dataset, weighted t-SNE presents the samples regarding the input features of the dataset so that the final plot resembles the XOR function.

\begin{figure}[tbp]
\centering
\begin{subfigure}[t]{.32\linewidth}
  \fbox{\includegraphics[width=.9\linewidth, height=.9\linewidth, alt={Weighted t-SNE of the XOR dataset at training epoch 45, with the classes still mixed}]{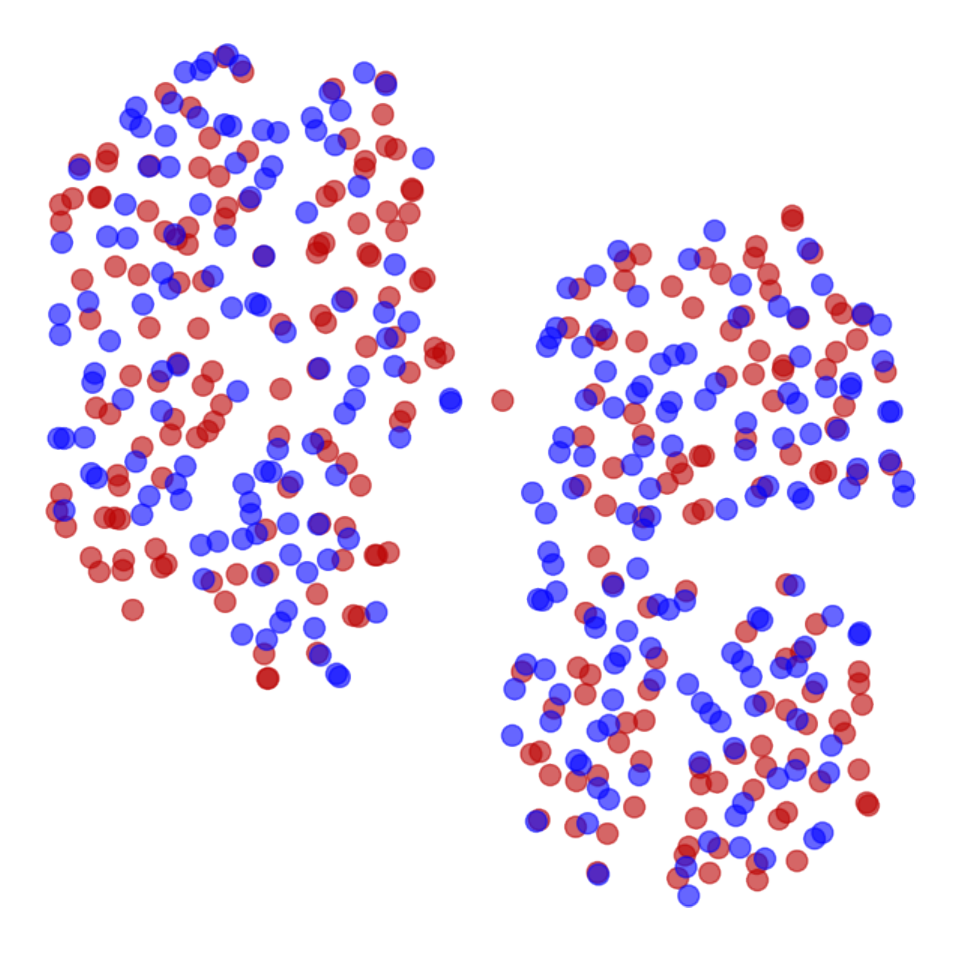}}
  \caption{Epoch: 45 \\ Loss: 0.844 \\ SC: $-$0.0008}
  \label{fig:xor045}
\end{subfigure}
\begin{subfigure}[t]{.32\linewidth}
  \fbox{\includegraphics[width=.9\linewidth, height=.9\linewidth, alt={Weighted t-SNE of the XOR dataset at training epoch 90, with the classes starting to separate}]{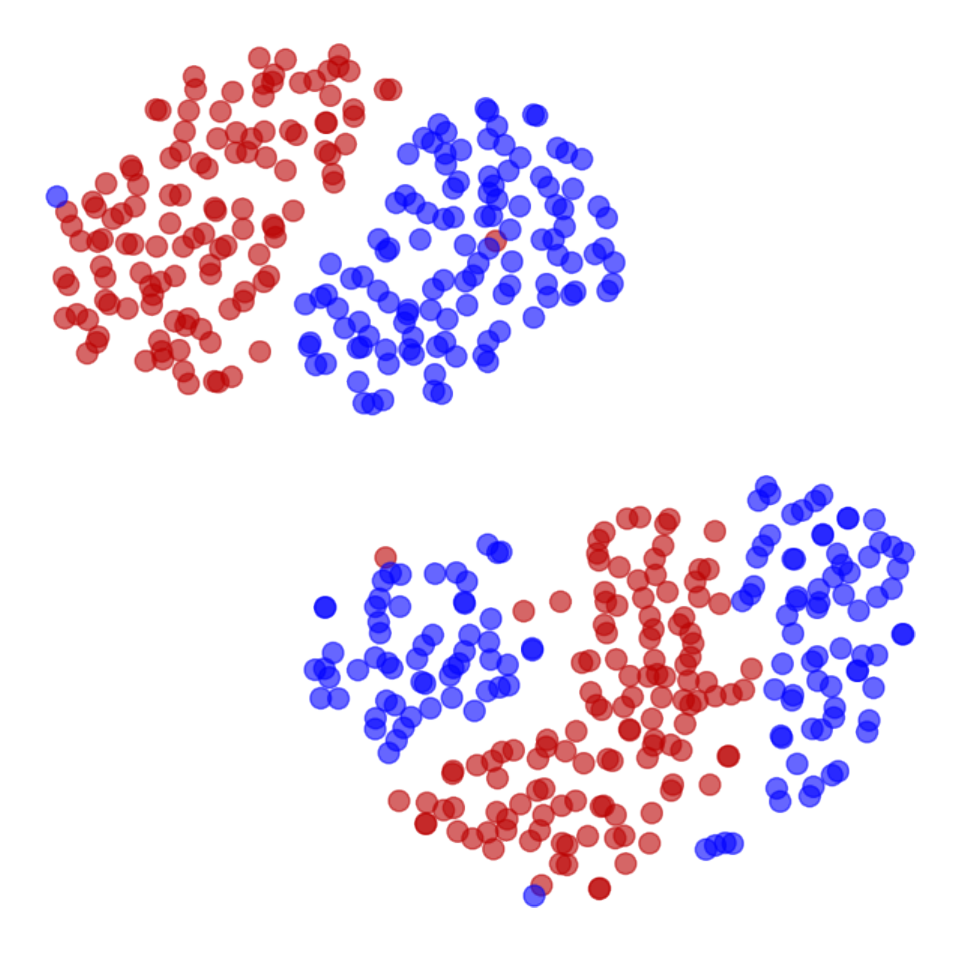}}
  \caption{Epoch: 90 \\ Loss: 0.726 \\ SC: 0.0875}
  \label{fig:xor090}
\end{subfigure}
\begin{subfigure}[t]{.32\linewidth}
  \fbox{\includegraphics[width=.9\linewidth, height=.9\linewidth, alt={Weighted t-SNE of the XOR dataset at training epoch 135, with four well separated clusters}]{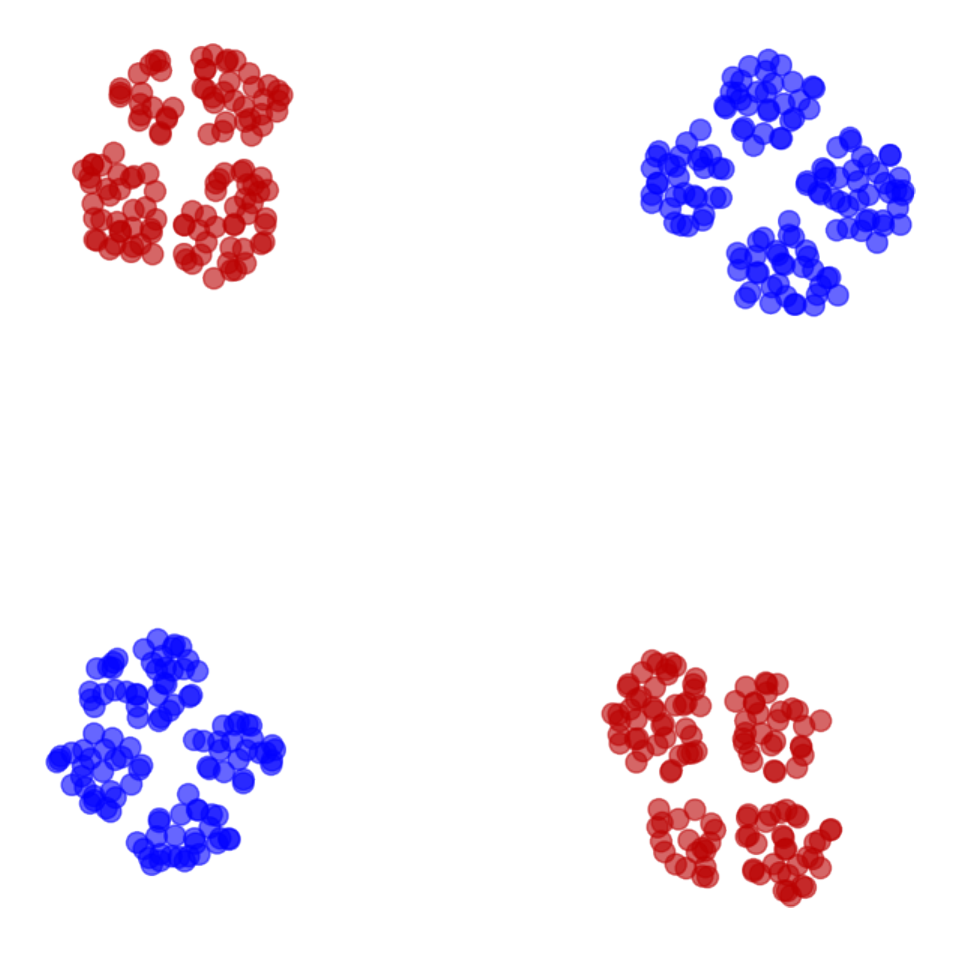}}
  \caption{Epoch: 135 \\ Loss: 0.366 \\ SC: 0.1992}
  \label{fig:xor135}
\end{subfigure}
\caption{Visualization of the training of a neural network with the XOR dataset. Each figure shows the weighted t-SNE for the feature scores of a specific training epoch of a neural network. Each caption describes the training epoch, the training loss, and the silhouette coefficient (SC) of the 2D embedding. Visualization details of the XOR dataset as in Fig.~\ref{fig:wtsnexor}. An animation of this plot is available at \url{https://sbcblab.github.io/wtsne/}.}
\label{fig:xormovie}
\end{figure}

\begin{figure}[tbp]
\centering
\begin{subfigure}[t]{.49\linewidth}
  \centering
  \fbox{\includegraphics[width=.9\linewidth, height=.9\linewidth, alt={Weighted t-SNE of the mouse cortex dataset at training epoch 2}]{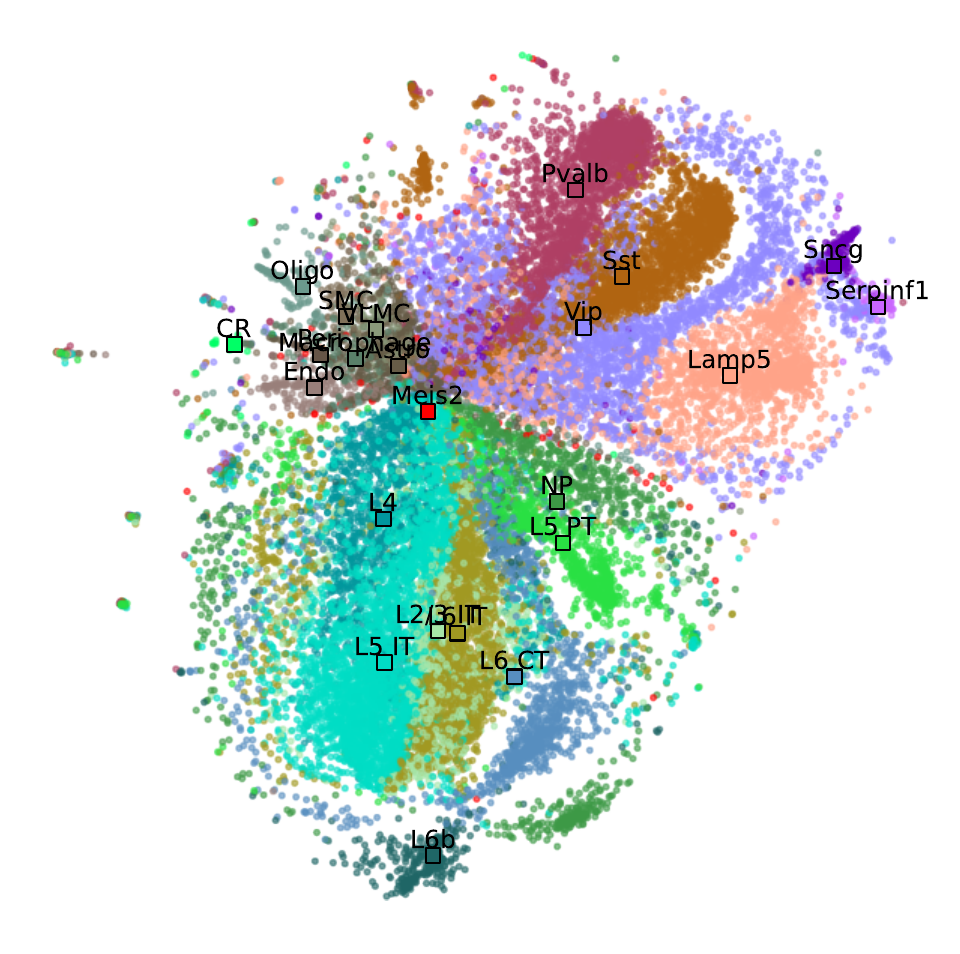}}
  \caption{Epoch: 2 \\ Loss: 19.65 $\lvert$ SC: $-$0.049}
  \label{fig:mouse02}
\end{subfigure}
\begin{subfigure}[t]{.49\linewidth}
  \centering
  \fbox{\includegraphics[width=.9\linewidth, height=.9\linewidth, alt={Weighted t-SNE of the mouse cortex dataset at training epoch 4}]{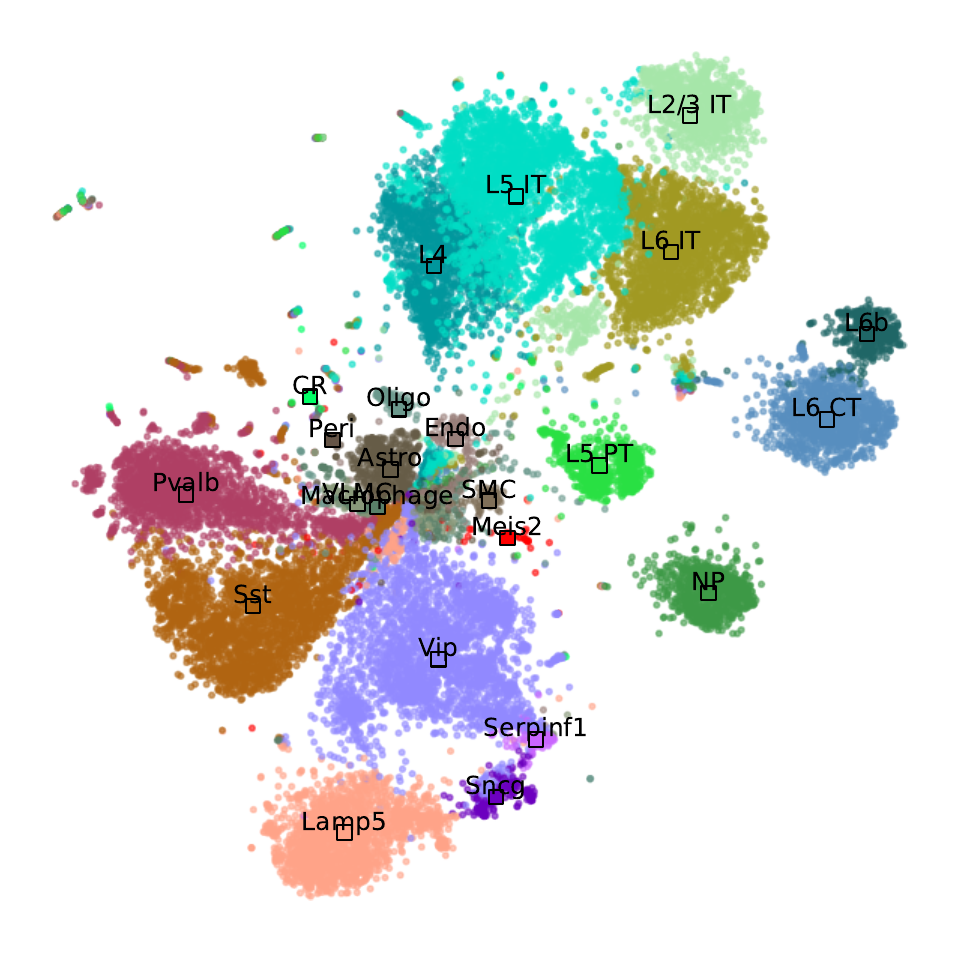}}
  \caption{Epoch: 4 \\ Loss: 9.07 $\lvert$ SC: 0.246}
  \label{fig:mouse04}
\end{subfigure}

\begin{subfigure}[t]{.49\linewidth}
  \centering
  \fbox{\includegraphics[width=.9\linewidth, height=.9\linewidth, alt={Weighted t-SNE of the mouse cortex dataset at training epoch 6}]{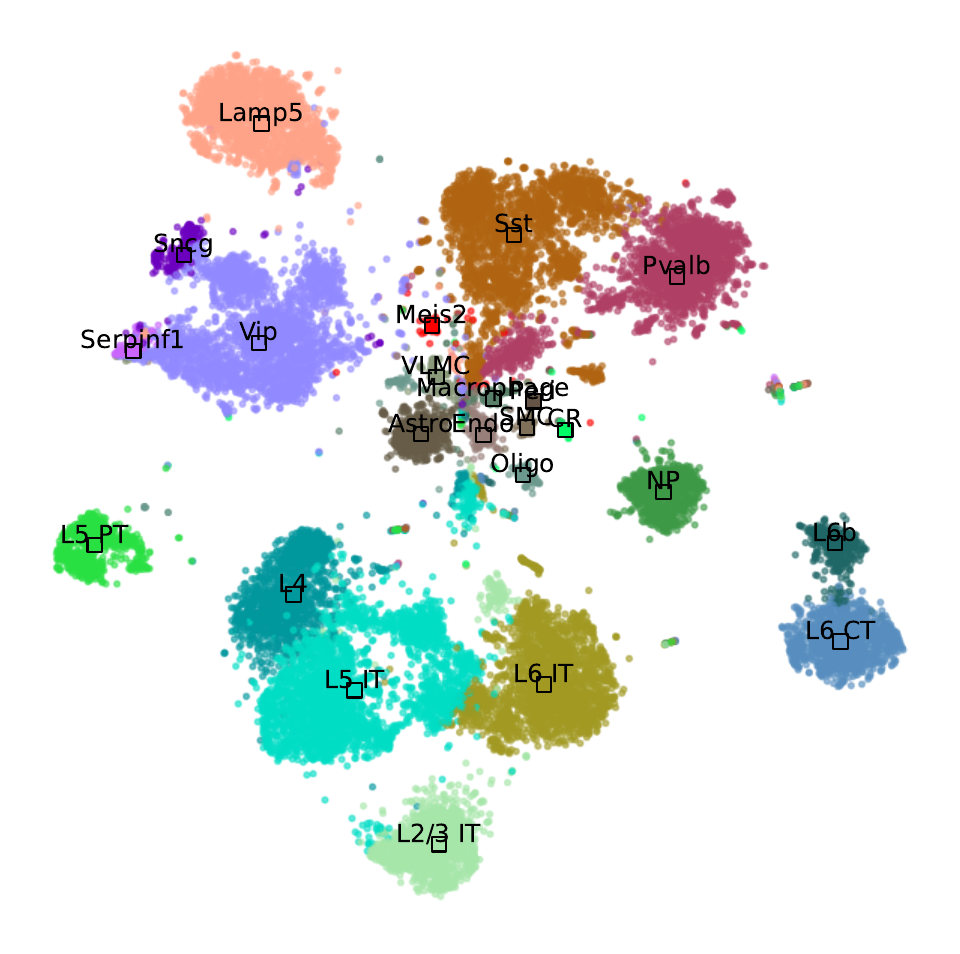}}
  \caption{Epoch: 6 \\ Loss: 4.56 $\lvert$ SC: 0.344}
  \label{fig:mouse06}
\end{subfigure}
\begin{subfigure}[t]{.49\linewidth}
  \centering
  \fbox{\includegraphics[width=.9\linewidth, height=.9\linewidth, alt={Weighted t-SNE of the mouse cortex dataset at training epoch 8}]{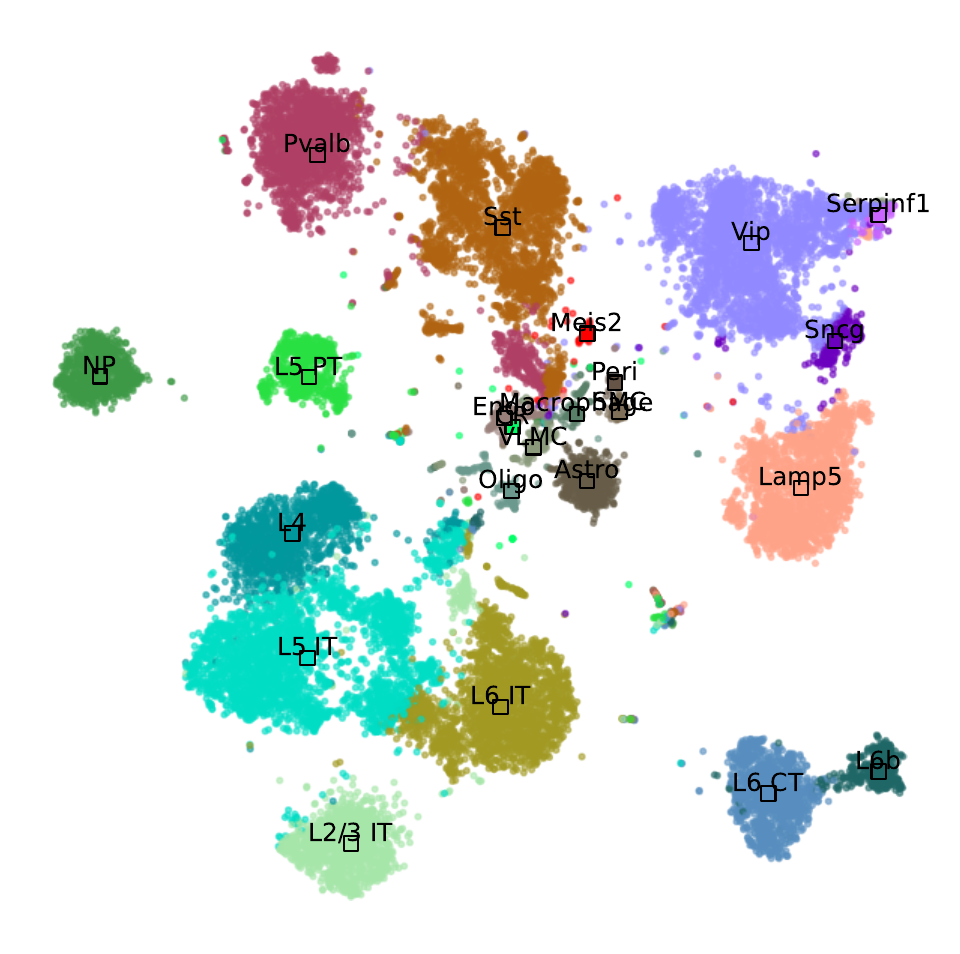}}
  \caption{Epoch: 8 \\ Loss: 3.86 $\lvert$ SC: 0.367}
  \label{fig:mouse08}
\end{subfigure}
\caption{Visualization of the training of a neural network with the mouse cortex dataset. Each figure shows the weighted t-SNE for the feature scores of a specific training epoch of a neural network. Each caption describes the training epoch, the training loss, and the silhouette coefficient (SC) of the 2D embedding. Visualization details of the mouse cortex dataset as in Fig.~\ref{fig:wtsnemouse}. An animation of this plot is available at \url{https://sbcblab.github.io/wtsne/}.}
\label{fig:mousemovie}
\end{figure}

Weighted t-SNE can also be applied at different moments during the training of the neural networks, allowing the visualization of the learning process according to the current feature importance scores.
Figs.~\ref{fig:xormovie} and \ref{fig:mousemovie} show how the visualization changes at different training epochs and how the networks ``learn'' to split the samples into their clusters as the training progress.
Table~\ref{tab:correlation} shows that the training loss of the neural networks is inversely correlated to the corresponding silhouette coefficient, indicating that improving the prediction also improves the feature scoring, thus resulting in better clustering in the 2D embedding.

\begin{table}[tbp]
\centering
\caption{Pearson correlation between training loss of neural network and the corresponding silhouette coefficient of the weighted t-SNE 2D embedding.}
\label{tab:correlation}
\begin{tabular}{lr}
\hline
\textbf{Dataset} & \textbf{Correlation} \\ \hline
XOR              & $-$0.85 \\
Synth            & $-$0.37 \\
Liver            & $-$0.81 \\
Prostate         & $-$0.67 \\
Regression       & $-$0.74 \\
Mouse cortex     & $-$0.98 \\ \hline
\end{tabular}
\end{table}

So far, our experiments have focused on the relationship between cluster visualization in weighted t-SNE, silhouette coefficient, and prediction performance.
The key takeaway was that feature scorers producing clearer clusters in weighted t-SNE tend to correlate with higher classification performance, as evidenced by the kNN classifier results in Table~\ref{tab:silhouette}, the alignment of silhouette coefficients with F1-scores in Fig.~\ref{fig:wtsnemouse}, and the correlation between neural network loss and silhouette scores in Fig.~\ref{fig:mousemovie} and Table~\ref{tab:correlation}.
These findings suggest that weighted t-SNE can be a useful tool for selecting feature scorers that improve downstream prediction in supervised learning pipelines.
This focus was due to the relevance of feature scoring in supervised learning and the need to somehow quantify an intrinsically visual and interpretative result such as weighted t-SNE visualizations.
Because labeled datasets were used in the experiments, it is also fair to assume that the data is clusterable, as the samples are already sorted into predefined classes.
However, this strong connection between cluster clarity and classification accuracy raises an important question: if accuracy correlates with better feature scoring, is visualization even necessary, or would prediction metrics alone suffice?

\begin{figure}[tbp]
\centering
\begin{subfigure}[t]{.40\linewidth}
  \centering
  \fbox{\includegraphics[width=.9\linewidth, height=.9\linewidth, alt={Weighted t-SNE scatterplot for Random Forest A, showing a red cluster separated from a single cluster mixing blue and cyan samples}]{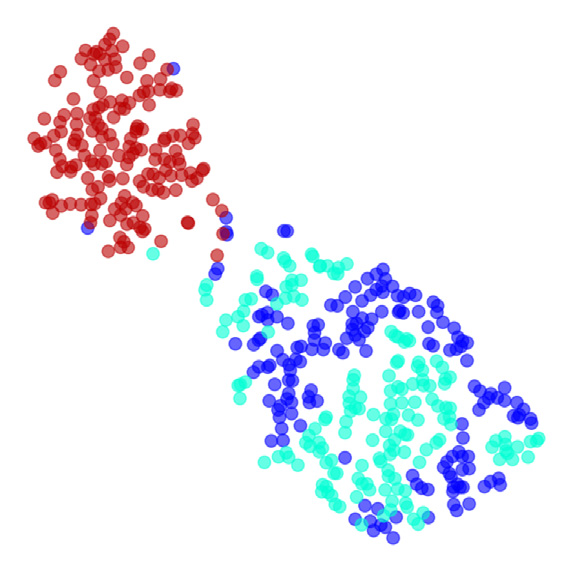}}
  \caption{Random Forest A}
  \label{fig:rfA}
\end{subfigure}
\begin{subfigure}[t]{.40\linewidth}
  \centering
  \fbox{\includegraphics[width=.9\linewidth, height=.9\linewidth, alt={Weighted t-SNE scatterplot for Random Forest B, showing three separated clusters of red, blue, and cyan samples}]{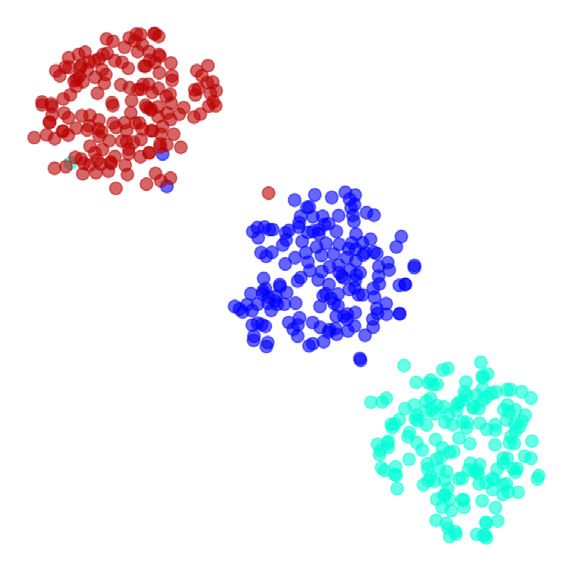}}
  \caption{Random Forest B}
  \label{fig:rfB}
\end{subfigure}
\caption{Comparison using weighted t-SNE between two random forests with the same hyperparameters. Both of them achieved the same training F1-score of 0.99 and test F1-score of 0.93. Even so, they scored features differently, resulting in distinct visualizations. The two random forests were trained to label the red samples from the blue and cyan samples in a binary classification.}
\label{fig:rfcompare}
\end{figure}

To show that weighted t-SNE reveals insights beyond what prediction metrics can capture, we devised the following experiment.
We created a synthetic dataset with three classes, 500 samples, and 150 features, where each class is defined by five relevant features (15 relevant features in total), while the remaining features contain only noise.
The original dataset contained three distinct classes: red, blue, and cyan, as shown in Fig.~\ref{fig:rfcompare}.
However, before training, we manually merged the blue and cyan samples into a single class, reducing the task to a binary classification problem between red and blue.
We then trained two random forest models (A and B) on this binary-labeled dataset using the same hyperparameters.
Both models achieved identical classification performance: F1-score of 0.99 on training and 0.93 on test.
However, because of the random initialization of the algorithm, each random forest assigned different feature importance scores, despite reaching the same F1-scores.
The weighted t-SNE projections for both models are shown in Fig.~\ref{fig:rfcompare}.
Random Forest A (Fig.~\ref{fig:rfA}) groups the samples into two clusters, red and blue, strictly following the binary classification labels, ignoring internal structure within the blue class.
Random Forest B (Fig.~\ref{fig:rfB}) instead produces three clusters, preserving the original three-class structure, even though it was trained on a binary task.
These differences arise because each model relied on different features, as seen in Table~\ref{tab:rfimportance}, which lists the top ten most important features for both models.
Even though they achieved the same F1-scores, they assigned importance to different features, ultimately influencing how the samples were structured in the visualization.

\begin{table}[tbp]
\centering
\caption{The feature importance of the ten top ranking features of the two random forests from Fig.~\ref{fig:rfcompare}.}
\label{tab:rfimportance}
\begin{tabular}{lrlr}
\hline
\multicolumn{2}{c}{\textbf{Random Forest A}} & \multicolumn{2}{c}{\textbf{Random Forest B}} \\ \hline
\textbf{Feature} & \textbf{Importance} & \textbf{Feature} & \textbf{Importance} \\ \hline
rel14 & 0.150922 & rel13 & 0.126705 \\
rel11 & 0.144392 & rel15 & 0.112081 \\
rel15 & 0.116410 & rel11 & 0.111083 \\
rel12 & 0.100376 & rel14 & 0.075461 \\
rel7  & 0.046047 & rel9  & 0.068682 \\
rel13 & 0.036134 & rel6  & 0.046066 \\
rel10 & 0.033920 & rel12 & 0.043192 \\
rel6  & 0.027337 & rel8  & 0.040205 \\
rel3  & 0.021719 & rel10 & 0.037155 \\
rel4  & 0.019296 & rel4  & 0.026757 \\ \hline
\end{tabular}
\end{table}

Distinguishing between these two models based solely on F1-scores would be impossible.
The models are equally predictive, yet one preserves meaningful structure in the data while the other does not.
This highlights a limitation of using prediction metrics alone to evaluate feature scorers: they do not capture how scorers organize the data or whether they reveal hidden structure.
This raises the interesting question of which model is better.
We argue that there is no single correct answer.
If the decision to merge blue and cyan was intentional, then Random Forest A is preferable because it enforces this labeling choice.
If the user was unaware that blue contained two distinct subgroups, then Random Forest B provides a valuable insight, revealing that the merged class actually contains two coherent clusters.
In either case, weighted t-SNE provides an intuitive way for users and domain experts to inspect feature scoring, offering insights that would be difficult to extract from raw feature importance values alone, especially in high-dimensional datasets.

While this is a synthetic example, similar situations frequently occur in real-world applications where interpretability is critical.
In fraud detection, one model may prioritize transaction history while another focuses on user behavior.
Both may achieve the same fraud classification accuracy, but a domain expert might prefer one feature set over the other depending on interpretability needs.

Weighted t-SNE can be especially useful for the interpretability of neural networks applied to tabular datasets such as the examples in Table~\ref{tab:tsnedata}.
Empirical evidence has shown that providing good explanations for networks with many fully connected layers, commonly used in tabular datasets, is hindered by a lack of selectivity \cite{montavon2018methods}.
In works based on transformers, the attention layer is restricted to categorical features and is not applied to continuous features \cite{huang2020tabtransformer}, which is a large drawback for interpreting results for data comprising mainly of this kind of features, such as gene expression data.
Meanwhile, surrogate models such as Local Interpretable Model-Agnostic Explanations (LIME) \cite{ribeiro2016should} rely on defining a neighborhood, which is not well defined for tabular data \cite{molnar2022interpretable}, even though newer methods are exploring solutions for this issue \cite{bjorklund2019sparse, bjorklund2022slicemap}.
In this context, the use of weighted t-SNE in conjunction with methods like relevance aggregation can be a new and helpful tool for inspecting neural networks trained on tabular datasets \cite{grisci2021relevance}.
Together with other metrics, such as the training loss, it can help researchers and practitioners decide if the model is fulfilling their expectations.
While outside the scope of the current work, the visualization of deep features using weighted t-SNE is also a possible future direction discussed in Section~\ref{sec:conclusion}.
It would include adapting weighted t-SNE for unstructured data types, such as images and text, by integrating gradient-based feature attributions or deep learning interpretability techniques.

\subsection{Limitations}\label{subsec:limitations}

While weighted t-SNE provides a novel approach to visualizing feature scorer results, it has certain limitations that should be acknowledged and contextualized.
Unlike traditional t-SNE, which works directly on raw data, our method requires a precomputed set of scores from a feature scorer.
If these scores are unreliable or poorly computed, the visualization may misrepresent the structure of the data.
However, this is an inherent challenge in feature scoring itself rather than a flaw in weighted t-SNE.
Moreover, this limitation is mitigated by the very purpose of weighted t-SNE: it allows users to compare different feature scorers visually, helping identify cases where a scorer produces misleading importance rankings.
Unlike standard t-SNE, which is typically used in unsupervised exploratory analysis, weighted t-SNE depends on labeled data to obtain feature importance scores.
While this may contradict the usual unsupervised nature of dimensionality reduction, this is a necessary property of most feature scorers, as feature importance usually requires a target variable to be defined.
Nonetheless, future work could investigate unsupervised feature scoring techniques to expand the applicability of weighted t-SNE to unlabeled datasets.
Like traditional t-SNE, weighted t-SNE inherits some sensitivity to hyperparameters such as perplexity and learning rate.
However, weighted t-SNE itself does not significantly increase computational cost beyond standard t-SNE, since the feature weighting step is a simple transformation.

Feature selection methods are often evaluated by classification or regression accuracy, and weighted t-SNE does not directly optimize for these metrics.
This is by design, as our goal is not to replace accuracy-based evaluation but to complement it with interpretability and visualization.
Feature importance scores affect model behavior in ways that accuracy alone cannot reveal: for example, two models with similar predictive performance may use completely different feature sets, which would be difficult to analyze numerically but can be visually assessed with weighted t-SNE.
In future work, a hybrid evaluation strategy could be developed that incorporates both accuracy-based metrics and visualization to provide a more holistic assessment of feature scorers.

Not all high-dimensional datasets exhibit clear cluster structures, making it difficult to interpret whether a visualization is ``better'' based on apparent separability.
However, weighted t-SNE does not assume that data must be clusterable, rather, it shows how feature scorers influence the structure of the data visualization.
In cases where no meaningful separation appears, this is itself useful information: it suggests that the scorer may not be effectively capturing feature relevance.
Future research could explore alternative evaluation metrics beyond the silhouette coefficient to assess the quality of weighted t-SNE projections in non-clusterable datasets.
We used the silhouette score to provide a numerical approximation of the separability observed in the visualizations.
However, the silhouette score is not without limitations.
It is usually higher for convex clusters than for other types of clusters.
Other clustering validity indices, such as the Davies--Bouldin index or Dunn index, could be explored in future work to complement or refine this approach.
Since weighted t-SNE is fundamentally a visualization tool, numerical cluster metrics should be interpreted with caution, and user discretion is always recommended.

Weighted t-SNE is, by design, a comparative and contrastive tool.
The visualizations are only meaningful when interpreted in relation to a baseline, typically, a regular t-SNE projection of the original dataset without feature scores.
A single weighted t-SNE visualization from one scorer, without comparison to a baseline, provides limited insight into how the feature importance scores affect the data structure.
Users should always compare weighted t-SNE outputs to the original unweighted t-SNE, as only then can the changes introduced by feature scores be properly assessed.

Our experiments were conducted exclusively on tabular datasets, and the adaptation of weighted t-SNE to other data types, such as text or image data, remains an open challenge.
In structured tabular data, feature importance is typically well-defined, making weighted t-SNE straightforward to apply.
However, feature importance in unstructured data, such as deep learning feature maps for images or attention weights for text, follows different paradigms.

\section{Conclusion}\label{sec:conclusion}

This work introduced weighted t-SNE, a novel visualization technique that enhances the inspection of feature scoring algorithms.
The method allows users to visually assess feature scorers, providing an intuitive alternative to numerical importance values and predictive metrics.
A series of experiments demonstrated the effectiveness of weighted t-SNE across nine feature scorers, two synthetic datasets, and two cancer microarray datasets.
The key findings were: Feature scorers with higher classification performance tended to produce clearer clusters in weighted t-SNE visualizations, as confirmed by silhouette coefficient analysis (Table~\ref{tab:silhouette}, Fig.~\ref{fig:wtsnemouse}, Table~\ref{tab:correlation}).
The visualization of feature scores using weighted t-SNE is different from the visualization of feature selection (Fig.~\ref{fig:wtsnesynthrelieffperfect}) or internal model's parameters (Fig.~\ref{fig:wtsneactxor}).
Weighted t-SNE enabled direct visual comparison between scorers, revealing differences in how they structured the data, which would be difficult to discern from raw importance values (Figs.~\ref{fig:wtsnexor}, \ref{fig:wtsnesynth}, \ref{fig:wtsneliver}, \ref{fig:wtsneprostate}).
The visualizations generated with weighted t-SNE are trustworthy in regard to the high-dimensional data (Table~\ref{tab:trustworthiness}).
Outlier detection was improved, as weighted t-SNE made it easier to visually identify samples that were otherwise obfuscated by irrelevant features in high-dimensional space (Fig.~\ref{fig:wtsneliver}).
In regression tasks, weighted t-SNE successfully reflected the importance of continuous target variables, extending its applicability beyond classification (Fig.~\ref{fig:teaser}).
Tracking model learning over time was made possible, as weighted t-SNE was used to visualize changes in feature representations across neural network epochs, helping assess how models evolve during training (Figs.~\ref{fig:xormovie}, \ref{fig:mousemovie}).
Weighted t-SNE successfully guided feature scorer selection, allowing us to identify feature scorers that improved the F1-score of an independently trained kNN classifier (Table~\ref{tab:silhouette}).
An experiment with two random forests showed that even models with identical predictive performance can assign importance to different features, leading to distinct visualizations, an insight that accuracy-based evaluation alone fails to provide (Fig.~\ref{fig:rfcompare}, Table~\ref{tab:rfimportance}).
These results establish weighted t-SNE as a valuable tool for feature importance analysis in machine learning.
It enhances feature importance analysis by providing a contrastive, visual tool that reveals differences between feature scorers, aids in outlier detection, tracks model learning, and helps select feature scorers that improve downstream tasks.
Unlike accuracy-based evaluation or raw importance values, it uncovers hidden structure, preserves meaningful feature relationships, and integrates seamlessly into machine learning workflows.

More ideas based on weighted t-SNE can be developed in the future.
Techniques such as joint t-SNE \cite{wang2021joint} could be used to improve the comparability between the 2D embeddings of several scorers, and alternatives exploring other algorithms like the UMAP instead of the t-SNE can be tested.
One possible future research is regarding the use of weighted t-SNE to visualize the importance of deep features in modern deep learning models.
Here we define deep features as abstract features extracted from deep learning models, such as convolutional neural networks or transformers.
However, this would require modifications to account for the unique characteristics of deep features, such as their abstract nature and context-dependence (e.g., image, text, or time-series data).
The method would need to select a specific deep learning model and specialized feature scorers (e.g., gradient-based methods like Grad-CAM \cite{selvaraju2017grad} or attention mechanisms \cite{vaswani2017attention}), and find the interpretation of high-dimensional embeddings, maybe through sparse autoencoders \cite{oneill2024disentangling}.
Another research direction is investigating the use of weighted t-SNE in unsupervised learning contexts by incorporating unsupervised feature importance metrics or exploring interactive visual analytics tools that allow dynamic comparison of multiple feature scorers on the same dataset.

Feature importance is not just about improving accuracy.
It is about understanding what a model has learned.
Weighted t-SNE provides a powerful interpretability tool by making feature importance visually accessible, helping users detect outliers, uncover hidden structure, track model learning, and align models with domain knowledge.
As deep learning and explainable AI research continue to focus on image and text data \cite{molnar2022interpretable}, many real-world datasets, particularly in science and business, remain tabular \cite{borisov2022deep}.
Weighted t-SNE is well-suited for high-dimensional tabular datasets, such as gene expression data, where model interpretability is essential for drawing meaningful scientific conclusions \cite{feltes2019cumida, dorn2021comparison, grisci2019neuroevolution}.
By integrating weighted t-SNE into machine learning pipelines, researchers and practitioners can gain deeper insights into feature importance, ensuring that models not only perform well but also make decisions that align with human understanding.

\section*{CRediT authorship contribution statement}
\addcontentsline{toc}{section}{CRediT authorship contribution statement}

\textbf{Bruno Iochins Grisci:} Writing -- review \& editing, Writing -- original draft, Visualization, Validation, Software, Methodology, Investigation, Formal analysis, Data curation, Conceptualization.
\textbf{Mario Inostroza-Ponta:} Writing -- review \& editing, Methodology, Formal analysis.
\textbf{Márcio Dorn:} Writing -- original draft, Visualization, Supervision, Methodology, Investigation, Funding acquisition, Formal analysis, Data curation.

\section*{Data and code availability}\label{sec:dataavail}
\addcontentsline{toc}{section}{Data and code availability}

Interactive versions of all the figures and animations of Figs.~\ref{fig:xormovie} and \ref{fig:mousemovie} can be accessed in: \url{https://sbcblab.github.io/wtsne/}.
The necessary source code, hyperparameters, and datasets used in the experiments and results can be accessed on GitHub: \url{https://github.com/sbcblab/weighted_tSNE}.
The microarray data used in this work is available in the CuMiDa repository: \url{https://sbcb.inf.ufrgs.br/cumida}.

\section*{Declaration of competing interest}
\addcontentsline{toc}{section}{Declaration of competing interest}

The authors declare that they have no known competing financial interests or personal relationships that could have appeared to influence the work reported in this paper.

\section*{Acknowledgments}
\addcontentsline{toc}{section}{Acknowledgments}

This work was supported by grants from the Fundação de Amparo à Pesquisa do Estado do Rio Grande do Sul (FAPERGS), Brazil [24/2551-0001392-0; 23/2551-0001894-2; 19/2551-0001906-8; 24/2551-0000725-3], Conselho Nacional de Desenvolvimento Científico e Tecnológico (CNPq), Brazil [314082/2021-2; 440279/2022-4; 408154/2022-5 and 404319/2024-6], Fondecyt Regular 1231505, ANID/Chile, and the Emerging Leaders in the Americas Program Scholarship with the support of the Government of Canada.
This study was financed in part by the Coordenação de Aperfeiçoamento de Pessoal de Nível Superior - Brasil (CAPES) - Finance Code 001.

\FloatBarrier

\bibliographystyle{model1-num-names}
\bibliography{refs}

\bigskip

\noindent \textbf{Bruno Iochins Grisci} is a Professor in the Department of Theoretical Informatics at the Federal University of Rio Grande do Sul (UFRGS). He holds a Ph.D., M.Sc., and B.Sc.\ in Computer Science from UFRGS. His research focuses on explainable AI, machine learning, metaheuristics, bioinformatics, and data visualization. He has conducted research at Unisinos (with Dell Inc.) and completed exchange programs at the University of Birmingham, the University of Santiago de Chile, Karlsruhe Institute of Technology, and Dalhousie University. He is an alumnus of the Heidelberg Laureate Forum and the Emerging Leaders in the Americas Program.

\bigskip

\noindent \textbf{Mario Inostroza-Ponta} received his Computer Engineering degree from the Universidad de Santiago de Chile, Santiago, Chile, in 2001, and his Ph.D.\ in Computer Science from the University of Newcastle, Callaghan, NSW, Australia, in 2008. He is currently an Associate Professor with the Departamento de Ingeniería Informática, Universidad de Santiago de Chile. He spent a year as a Post-Doctoral Researcher at the Institute of Informatics, Federal University of Rio Grande do Sul, Porto Alegre, Brazil. His current research interests include the development of ad-hoc computational techniques for bioinformatics and other areas, using data mining, metaheuristics, and multiobjective optimization.

\bigskip

\noindent \textbf{Márcio Dorn} received his M.Sc.\ degree in Computer Science from the Pontifical Catholic University of Rio Grande do Sul, Porto Alegre, Brazil, in 2008, and his Ph.D.\ degree in Computer Science from the Federal University of Rio Grande do Sul (UFRGS), Porto Alegre, Brazil, in 2012. He is a Professor at the Institute of Informatics, UFRGS, where he also leads the Structural Bioinformatics and Computational Biology Laboratory. He was a Research Associate at the Karlsruhe Institute of Technology, Karlsruhe, Germany, in 2008, 2009, and 2017. His current research interests include bioinformatics, structural bioinformatics, machine learning, metaheuristics, artificial intelligence, and high-performance computing. Prof.~Dorn is a CNPq (Brazilian National Research Council) Advanced Fellow and an Alexander von Humboldt Research Fellow.

\clearpage
\appendix
\makeatletter
\newcommand{\section@cntformat}{Appendix~\thesection.\quad}
\makeatother
\setcounter{equation}{0}
\setcounter{table}{0}
\renewcommand{\theequation}{A.\arabic{equation}}
\renewcommand{\theHequation}{A.\arabic{equation}}
\renewcommand{\thetable}{A.\arabic{table}}
\renewcommand{\theHtable}{A.\arabic{table}}
\renewcommand{\thealgocf}{A.\arabic{algocf}}

\section{Supplementary data}\label{app:supp}
\suppressfloats[t]

Supplementary material related to this article can be found online at \url{https://doi.org/10.1016/j.neucom.2025.130561}.

\subsection{Feature scorers}\label{app:featurescorers}

In the sections below we give further detail on how each of the feature scorers used in the experiments presented in the main text compute feature importance.
A reproduction of the summary of the feature scorers from the main text is reproduced in Table~\ref{tab:supp_fsprop}.
The definitions and equations listed below come mainly from the work of \citet{barbieri2024analysis}, \citet{grisci2021relevance}, \citet{montavon2018methods, montavon2019layer}, \citet{molnar2022interpretable}.

\subsubsection{Kruskal Wallis H test}

The one-way analysis of variance (ANOVA) is a statistical technique that evaluates variation within and across groups (i.e., within a random variable and between different variables, respectively) to determine whether a set of random variables originates from the same distribution.
In this study, we employ the Kruskal-Wallis H test, a non-parametric alternative to ANOVA.
Unlike ANOVA, this test does not require the assumption of normally distributed data.
Its mathematical definition is outlined in Equation~\ref{eq:supp_kw}.

\begin{equation}
H = \left( \frac{12}{n(n+1)} \sum_{j=1}^{k} \frac{R_i^2}{n_i} \right) - 3(n+1)
\label{eq:supp_kw}
\end{equation}

\begin{table}[tbp]
\centering
\caption{Properties of the feature scorers used in the experiments. The type refers to how the scorer works and the correlation to how interactions between features are treated.}
\label{tab:supp_fsprop}
\begin{tabular}{llll}
\hline
\textbf{Scorer}    & \textbf{Type} & \textbf{Correlation} & \textbf{Reference} \\ \hline
Kruskal-Wallis     & Filter        & Univariate           & \cite{kruskal1952use} \\
Mutual Information & Filter        & Univariate           & \cite{vergara2014review} \\
mRMR               & Filter        & Multivariate         & \cite{peng2005feature} \\
ReliefF            & Filter        & Multivariate         & \cite{kononenko1997overcoming} \\
Lasso              & Embedded      & Multivariate         & \cite{fonti2017feature} \\
Decision Tree      & Embedded      & Multivariate         & \cite{stone1984classification} \\
Random Forest      & Ensemble      & Multivariate         & \cite{breiman2001random} \\
Linear SVM         & Embedded      & Multivariate         & \cite{cortes1995support} \\
Neural Network     & Embedded      & Multivariate         & \cite{grisci2021relevance} \\ \hline
\end{tabular}
\end{table}

Where $k$ is the number of compared groups, $n$ is the total number of samples, $n_i$ is the number of samples in the $i$-th group, and $R_i^2$ is the sum of the values in the $i$-th group.

In feature scoring, the goal is to give a numerical importance value to features that most effectively distinguish the target class.
To achieve this, we compute the Kruskal-Wallis H test score for each feature by grouping its values according to the target class.
Features with higher test scores are then selected \cite{grisci2019neuroevolution}.
Since this method evaluates each attribute independently in relation to the target class, it is considered univariate and does not account for potential relationships or dependencies between features.

\subsubsection{Mutual Information}

A core concept in information theory is entropy, which quantifies the uncertainty associated with a random variable in predicting the likelihood of an event \cite{vergara2014review}.
The entropy $H(x)$ of a random variable is mathematically expressed in Equation~\ref{eq:supp_entropy}.

\begin{equation}
H(x) = - \sum_{i=1}^{n} P(x_i) \log_2(P(x_i))
\label{eq:supp_entropy}
\end{equation}

Where $n$ is the total number of instances, $x$ is a random variable, $x_i$ is the i-th value in the variable $x$, and $P(x_i)$ is the probability of $x_i$ occurring.

Building on the idea of entropy, Mutual Information (MI) measures the amount of information one random variable provides about another \cite{vergara2014review}.
Its formal definition is provided in Equation~\ref{eq:supp_mi}.

\begin{equation}
I(x;y) = \sum_{i=1}^{n} \sum_{j=1}^{n} P(x_i, y_j) \cdot \log \left( \frac{P(x_i, y_j)}{P(x_i) \cdot p(y_j)} \right)
\label{eq:supp_mi}
\end{equation}

Where $n$ is the total number of instances, $x$ and $y$ are random variables, $x_i$ and $y_i$ are the i-th value in the variable $x$ and $y$, respectively, $P(x_i)$ and $P(y_i)$ are the probabilities of $x_i$ and $y_i$ occurring, respectively, and $P(x_i, y_i)$ is the joint probability of $x_i$ and $y_i$ occurring.

Using mutual information, we can assess a feature's relevance by determining how much information it shares with the target class.
This allows us to rank features based on their computed relevance.
Like the Kruskal-Wallis filter, this method evaluates features individually and does not account for dependencies between them, making it a univariate approach.

\subsubsection{mRMR}

In certain scenarios, we aim to evaluate the contribution of a new variable when added to an existing set of variables.
This can be achieved by assessing both the relevance and non-redundancy of the new variable.
The Minimum Redundancy Maximum Relevance (mRMR) method, introduced by \citet{peng2005feature}, is an iterative feature scoring algorithm.
It prioritizes features that maximize mutual information with the target class while minimizing redundancy with respect to the features already included in the selected subset.
At each iteration, the algorithm seeks to identify the feature $x_j$ that optimizes the criterion defined in Equation~\ref{eq:supp_mrmr}.

\begin{equation}
\max_{x_j \in X - S} \left[ I \left( x_j ; c \right) - R(x_j, S) \right]
\label{eq:supp_mrmr}
\end{equation}

Where $X$ is the set of all attributes, $S$ is the set of already selected features, $I$ is the mutual information function, and $R$ is the redundancy function.

\subsubsection{ReliefF}

ReliefF is a widely used algorithm for feature scoring or selection, renowned for its intuitive and efficient approach.
Its key strength lies in its core concept: for each instance in the dataset, it identifies the $k$-nearest neighbors from each class.
It then evaluates how much each attribute varies between these instances, making it a lightweight yet effective method that accounts for feature correlations \cite{robnik2003theoretical}.

The algorithm begins by initializing a weight vector that represents the relevance of each feature.
It then selects an instance and searches for its $k$-nearest neighbors within each class.
Instances belonging to the same class as the target instance are referred to as hits, while those from different classes are called misses.
The underlying logic is as follows: if the value of a given attribute $A$ differs between the target instance and a miss, $A$ is considered a useful feature for distinguishing between the classes, and its weight is increased.
Conversely, if the value of attribute $A$ differs between the target instance and a hit, $A$ is less effective at separating classes (since they belong to the same class), and its weight is decreased.
To ensure balance, misses are weighted by the probability $P(c)$ of their class occurring in the dataset, allowing misses and hits to contribute equally.
The algorithm ultimately returns a weight vector that reflects the relevance of each feature.

\subsubsection{Lasso}

The Least Absolute Shrinkage and Selection Operator, commonly referred to as Lasso, is a linear regression technique that incorporates $l1$-regularization to enhance prediction accuracy and mitigate overfitting.
Its mathematical formulation involves solving the optimization problem presented in Equation~\ref{eq:supp_lasso}.

\begin{equation}
\min_{\beta_0, \beta} \left\{ \frac{1}{N} \left\| y - \beta_0 - X\beta \right\|_2^2 + \lambda \| \beta \|_1 \right\}
\label{eq:supp_lasso}
\end{equation}

Where $y$ is the outcome, $X$ is the independent variables, $N$ is the number of instances in the data, $\beta$ is the unknown parameters to be calculated, and $\lambda$ is the regularization term.
The importance of each feature is simply the coefficient of that feature in the final regressor.

\subsubsection{Decision Tree}

This model represents a decision tree, where each node, structured in a top-down fashion, corresponds to a decision path determined by specific conditions (typically based on the values of a sample's attributes during prediction).
Each leaf node, in turn, represents a final label or outcome associated with that particular path.
As described by \citet{loh2011classification}, tree generation algorithms construct the tree recursively from the top down.
At each node, splits (decisions) are created by evaluating attributes using an impurity function (such as the Gini index), which assigns scores to attributes deemed discriminant.
This process continues until the tree is fully formed, with each branch representing a sequence of decisions leading to a final classification or outcome.
Feature importance is calculated as the (normalized) total reduction in the impurity criterion achieved by that feature, in what is commonly referred to as the Gini importance.

\subsubsection{Random Forest}

Random Forest is an ensemble learning algorithm that harnesses the principle of the ``wisdom of the crowd'' to reduce the variance in outcomes generated by individual classifiers.
The concept of Random Forest was first introduced by \citet{breiman2001random}.
It consists of a collection of decision trees, each trained on different subsets of instances created through bootstrapping (random sampling with replacement).
For a given instance, predictions are made by every tree in the forest, and the final predicted label is determined through a majority voting mechanism.
This approach involves aggregating the feature weights computed by each tree in the Random Forest into a single averaged feature weight vector, enabling the ranking of features based on their importance.
The feature importance is computed using the Gini importance as for the decision trees.

\subsubsection{Support Vector Machine}

Support Vector Machines (SVM) are a robust and efficient class of algorithms used for both regression and classification tasks.
The method works by constructing an $n$-dimensional hyper-plane or a set of hyper-planes that optimally separate instances by class, maximizing the margin---the distance between the hyper-planes and the nearest instances from each class \cite{cortes1995support}.

To handle cases where the data is not perfectly linearly separable, SVM employs the soft-margin approach, utilizing the hinge loss function.
Additionally, it incorporates Tikhonov regularization, also known as $l2$-regularization, to improve generalization.
The process of computing the classifier is equivalent to solving the optimization problem defined in Equation~\ref{eq:supp_svm}.

\begin{equation}
\left[ \frac{1}{n} \sum_{i=1}^{N} \max \left( 0, 1 - y_i \left( \vec{w} \cdot \vec{x}_i - b \right) \right) \right] + \lambda \| \vec{w} \|^2
\label{eq:supp_svm}
\end{equation}

Where $\vec{w}$ is the normal vector to the hyper-plane, $\frac{b}{\vec{w}}$ is the plane offset from the origin along $\vec{w}$, $n$ is the number of instances in the data, $x_i$ is i-th instance in the data, and $y_i$ is every $x_i$ class, which can be either 1 or -1, each representing one class.
For the task of feature scoring, and using a linear kernel for the SVM, the importance of each feature is the coefficient of that feature in the final model.

\subsubsection{Neural Networks}

An Artificial Neural Network (ANN) is a computational model inspired by the structure and function of the biological brain.
It consists of interconnected units called neurons, organized in layers, that work together to process and learn from data.
Each neuron receives input, performs a simple computation, and passes the result to the next layer.
Through training, the network adjusts its connections (weights) to recognize patterns, make predictions, or classify data.
Retrieving feature importance from ANNs is not a trivial task, given their ``black-box'' nature.
However, there are many methods that try to solve this problem, such as the Layer-Wise Relevance Propagation (LRP) \cite{bach2015pixel} and its adaptation for tabular data, the relevance aggregation \cite{grisci2021relevance}.

LRP works computing a backward pass that sends the output of the network back through its structure, using several possible rules that take into account the input domain and layer type.
The two LRP rules used in this work were LRP-$\alpha\beta$ \cite{bach2015pixel} in Equation~\ref{eq:supp_lrpab} and $w^2$-rule \cite{montavon2017explaining} in Equation~\ref{eq:supp_w2}.

\begin{equation}
R_j = \sum_k \left( \alpha \frac{a_j w_{jk}^{+}}{\sum_j a_j w_{jk}^{+}} - \beta \frac{a_j w_{jk}^{-}}{\sum_j a_j w_{jk}^{-}} \right) R_k
\label{eq:supp_lrpab}
\end{equation}

\begin{equation}
R_j = \sum_k \frac{w_{jk}^2}{\sum_j w_{jk}^2} R_k
\label{eq:supp_w2}
\end{equation}

For both equations, $k$ and $j$ are the $k$-th and $j$-th layers, $a$ is the output of a neuron, $w^+$ and $w^-$ are positive and negative weights, $\alpha$ and $\beta$ are constants, and $R$ is the relevance signal.
The constants follow the property that $\alpha - \beta = 1$ and $\beta \geq 0$ \cite{bach2015pixel}.
The $w^2$-rule is a special case for the input layer when any real-valued input is admissible.

For tabular data, it is adequate to use relevance aggregation.
The core concept involves training a neural network on the target dataset (line~\ref{alg:train}) and then calculating the relevance of each input for every sample (line~\ref{alg:relevance}).
The algorithm employs absolute values to ensure equal consideration of both relevance (positive values) and counter-relevance (negative values) (line~\ref{alg:abs}).
The computed values are rescaled to ensure that all samples contribute equally during aggregation (line~\ref{alg:abs}).
This step is especially crucial for regression tasks, where differences in target values could otherwise skew the relevance toward samples with higher target magnitudes.

\begin{algorithm}[tbp]
\caption{Relevance aggregation}\label{alg:relagg}
\KwData{$D_{n \times m}$: data, $c$: classes, $network$: neural network}
\KwResult{Ordered relevance scores}
\Begin{
  $R, S, score \gets [\,]$ \;
  train $network$ on $D_{n \times m}$\;\label{alg:train}
  \For{$sample_{1 \times m}$ in $D_{n \times m}$}{
    $out \gets$ predict($network$, $sample_{1 \times m}$)\;
    $rel_{1 \times m} \gets$ compute\_relevance($network$, $sample_{1 \times m}$, $out$)\;\label{alg:relevance}
    $rel_{1 \times m} \gets$ abs($rel_{1 \times m}$) / max(abs($rel_{1 \times m}$))\;\label{alg:abs}
    $R_{sample} \gets rel_{1 \times m}$\;
  }
  \For{$feat_{n \times 1}$ in $R_{n \times m}$}{
    \For{$class$ in $c$}{
      $S_{feat,class} \gets$ average($feat_{n \in class}$)\;\label{alg:classavg}
    }
  }
  \For{$feat_{1 \times c}$ in $S_{m \times c}$}{
    $score_{feat} \gets$ average($feat_{1 \times c}$)\;\label{alg:globalavg}
  }
  \Return{sort($score_{m \times 1}$)\;}\label{alg:sort}
}
\end{algorithm}

Aggregation is performed at two levels.
The first level operates by class, computing the average of the rescaled relevance for each input, but only for samples that belong to the same class (line~\ref{alg:classavg}).
For regression tasks, the default approach is to treat all samples as part of a single class.
This intermediate step of calculating aggregation scores for different classes enables the distinction of relevance scores between sample groups, recognizing that the neural network may rely on different sets of features to identify each group.
The second level of aggregation involves averaging the relevance scores across all classes (line~\ref{alg:globalavg}), producing a global relevance score that can be sorted (line~\ref{alg:sort}).
The final score is derived by averaging the class scores, ensuring that each class contributes equally to the overall result.
The algorithm~\ref{alg:relagg} is reproduced from \citet{grisci2021relevance}.

\FloatBarrier

\end{document}